\documentclass[manuscript]{acmart}
\usepackage{amsmath, amsthm, amsfonts}
\newcommand{\PA}[1]{\text{PA}(r(#1))}
\newcommand{\X} {[r(1), \dots, r(N)]}
\newcommand{\PX} {P(r(i) \mid \PA{i})}
\newcommand{\TPX} {\tilde P(r(i) \mid \PA{i})}
\usepackage{graphicx}
\usepackage{subcaption}
\usepackage{tikz}
\usetikzlibrary{shapes, arrows, positioning}
\usepackage{multirow}
\usepackage{algorithm}
\usepackage{enumitem}
\usepackage{algorithmic}
\newtheorem{d1}{Definition}
\graphicspath{{./figures/}}

\usepackage{tikz}
\usepackage{listings}
\usetikzlibrary{positioning, shapes.geometric, arrows.meta}
\AtBeginDocument{%
  }

\acmJournal{TSAS}

\begin{document}
\title{Root Cause Analysis of Hydrogen Bond Separation in Spatio-Temporal Molecular Dynamics using Causal Models}

\author{Rahmat~Kemisola~Adesunkanmi}
\email{rahma@iastate.edu}
\orcid{0000-0002-5483-5076}
\affiliation{%
  \institution{ Iowa State University}
  \city{Ames}
  \state{Iowa}
  \country{USA}}

\author{Ashfaq~Khokhar}
\email{ashfaq.khokhar@gmail.com}
\orcid{0000-0002-6504-8502}
\affiliation{%
  \institution{ Iowa State University}
  \city{Ames}
  \state{Iowa}
  \country{USA}}


 \author{Goce Trajcevski}
\email{gocet25@iastate.edu}
\affiliation{%
 \institution{Iowa State University}
 \city{Ames}
 \state{Iowa}
 \country{USA}}
 
 \author{Sohail Murad}
 \email{murad@iit.edu}
 \orcid{0000-0002-1486-0680}
 \affiliation{%
  \institution{Illinois Institute of Technology}
  \city{Chicago}
  \state{Ilinnois}
  \country{USA}}
  
\renewcommand{\shortauthors}{Adesunkanmi et al.}
\begin{abstract}
Molecular dynamics simulations (MDS) face challenges, including resource-heavy computations and the need to manually scan outputs to detect "interesting events," such as the formation and persistence of hydrogen bonds between atoms of different molecules. A critical research gap lies in identifying the underlying causes of hydrogen bond formation and separation —understanding which interactions or prior events contribute to their emergence over time. With this challenge in mind, we propose leveraging spatio-temporal data analytics and machine learning models to enhance the detection of these phenomena. In this paper, our approach is inspired by causal modeling and aims to identify the root cause variables of hydrogen bond formation and separation events. Specifically, we treat the separation of hydrogen bonds as an "intervention" occurring and represent the causal structure of the bonding and separation events in the MDS as graphical causal models. These causal models are built using a variational autoencoder-inspired architecture that enables us to infer causal relationships across samples with diverse underlying causal graphs while leveraging shared dynamic information. We further include a step to infer the root causes of changes in the joint distribution of the causal models. By constructing causal models that capture shifts in the conditional distributions of molecular interactions during bond formation or separation, this framework provides a novel perspective on root cause analysis in molecular dynamic systems. We validate the efficacy of our model empirically on the atomic trajectories that used MDS for chiral separation, demonstrating that we can predict many steps in the future and also find the variables driving the observed changes in the system.
\end{abstract}

\ccsdesc[500]{Computing methodologies~Machine learning}
\ccsdesc[500]{Computing methodologies~Bayesian network models}

\ccsdesc[500]{Computing methodologies~Neural networks}
\keywords{Time Series, Root Cause Analysis, Causal Model, Variational Autoencoder }

\maketitle
\section{Introduction}\label{sec:introRCA}
 Managing large-scale spatio-temporal data—such as the motion of atoms involved in chemical interactions—may yield significant contributions to applications like drug discovery and new materials development. Spatio-temporal data management has made significant strides, addressing both foundational research problems and domain-specific applications. Early efforts in the field focused on formalizing data types and operators to query constructs~\cite{Erwig_1999} and processing algorithms along with indexing structures~\cite{Theoderidis_1996}—all of which set the stage for today's advanced capabilities. These developments have had a significant practical impact across high-stakes applications, such as urban computing and smart cities~\cite{Catlett_2019}, ecology~\cite{Chen_2011}, wildlife~\cite{Urbano_2010}, and disease mapping~\cite{López‐Quílez_2009}. With the rise of smart sensors, IoT, and Big Data, spatio-temporal data analytics now face novel challenges and opportunities, pushing the boundaries of the field and reinforcing its significance in both research and practical applications. More recently, significant efforts have resulted in machine learning (ML) and deep learning (DL) techniques like the implementation of Graph neural networks (GNNs)~\cite{scarselli2008graph}, multilayer perceptrons~\cite{murtagh1991multilayer}, and recurrent neural networks (RNNs)~\cite{Jain_2016} to address multiple problems related to spatio-temporal data, enabling effective models for various classification~\cite{Mao_2019}, data mining~\cite{Wang_2020} and prediction tasks~\cite{ Zhang_2023, Shi_2018}, offering novel challenges from data science perspective and even yielding a recent methodologies, including Spatio-temporal graph neural networks (STGNNs)~\cite{Sahili_2023, Jin_2023}. Specifically, in MDS, researchers have introduced these learning techniques like the neural networks~\cite{Stocker_2022} and its graph variants~\cite{Li_2022, Hofstetter_2022}. GNN allows for the extension of the existing neural network methods for processing the data represented in graph domains. The neural relational inference (NRI) model~\cite{kipf2018neural}, an extension of GNN and the variational autoencoder (VAE)~\cite{doersch2016tutorial, kingma2013auto}, is unsupervised and learns both interactions and dynamics directly from observational data. In this model, a VAE encodes the underlying interaction graph, which is reconstructed using GNNs. NRI has been extended to capture causality and causal models~\cite{liu2022structural,lowe2022amortized}.

\begin{figure}[!ht]
\centering
{\includegraphics[width=0.8\textwidth]{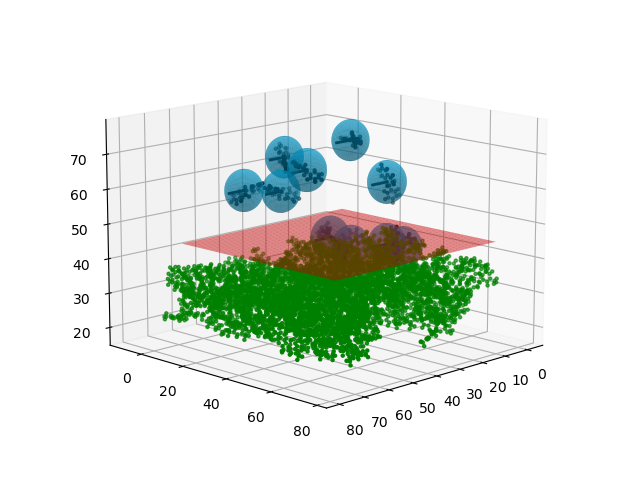}}
\caption{Molecular Dynamics Simulation (generalized version)~\cite{Anowar_InfoSys2024}.}
\label{fig:MDS}
\Description{Molecular structure, Relative positioning of rings, Bond forming}
\end{figure}

In this study, we address challenges in the domain of chemistry, specifically within the context of \textbf{Molecular Dynamics Simulations (MDS)}. Due to the extreme costs of physical experiments, domain researchers often resort to MDS to study chemical interactions. MDS has become a popular tool used in drug discovery and protein interaction binding due to its ability to circumvent the substantial costs and risks associated with physical experimentation, particularly in the initial exploratory stages. MDS generates a vast amount of spatio-temporal data in the form of atomic motion. However, atomic trajectories in standard MDS environments present unique analytical challenges~\cite{ADBIS2022}. First, individual atoms move within a 3D space and exhibit interdependent motion due to intra-molecular forces that bind multiple atoms into structured groups. Additionally, inter-molecular interactions can lead, in some instances, to bond formation between atoms from separate molecules. Consequently, the evolving relationships among these atoms are fascinating to domain scientists. However, this study of the interesting events, i.e., chemical bonds, is often performed by scanning the entire output dataset and based on algorithms for visualization based on domain experts. An illustration of a single timestamped view of MDS is shown in Figure~\ref{fig:MDS}. In this figure, the drug molecules are represented by spheres (sky blue) - with the radius of each sphere shown in black lines. The atoms from the polymer molecules (green) are at the bottom, with a surface plane (light red) in the middle.

Often, the domain scientists are interested in specific hard-to-analyze correlations that evolve, for example, the formation and persistence of Hydrogen Bonds (HB) among atoms. Beyond identifying these bonds, a more profound challenge is understanding the preceding events or "surrounding" factors that contribute to their separation. We posit that spatio-temporal analytics can significantly enhance the detection of these critical events in MDS data, providing chemists and other domain experts with valuable insights. In recent years, ML-based approaches have been developed to model molecular interactions~\cite{unke2021machine}, often using GNNs to capture atomic relationships~\cite{chmiela2017machine, unke2019physnet}. Although these approaches show promising results, they usually overlook causal interactions in molecular data. Addressing these challenges, we aim to enhance MDS accuracy and computational efficiency through causal discovery and root cause analysis. Causal discovery and learning~\cite{scholkopf2021toward, adesunkanmi2024neuro, peters2017elements} aim to develop data-driven models without relying on manually designed physical parameters. Techniques like the structural causal model (SCM) and potential outcome model (POM)~\cite{rubin2005causal} help identify causal relations from observational or interventional data~\cite{pearl2009causality}. For example, Structural Causal Molecular Dynamics (SCMD)~\cite{liu2022structural} uses SCMs~\cite{pearl2009causality} to model the causal structure of molecular interactions, often represented as dynamic Bayesian networks (DBNs). In particular, \cite{zhu2022neural} also utilizes the NRI framework to infer the latent interactions of protein allosteric communications. In DBNs, nodes represent atoms, while edges represent causal dependencies between atoms, which can be parameterized using GNNs to model the spatio-temporal dynamics. Since different molecules share underlying physical laws, an autoencoder can capture generic dynamics across molecules, enabling zero-shot generalized capability on unseen molecules. We can infer the causal graphs via encoders and predict the future forces or positions of all atoms using the decoder. Root cause analysis (RCA) is essential for system resilience, as it helps identify the underlying causes of faults or deviation in normal system behavior, ensuring faster recovery and reducing damage~\cite{wang2023incremental}. RCA spans various domains, from telecommunications~\cite{zhang2020influence} to medicine~\cite{peerally2017problem}, where modern systems are often networked with complex interdependencies~\cite{ikram2022root}. Recent work in RCA uses causal structure discovery to identify root causes of system faults~\cite{qiu2020causality, wang2018cloudranger}. The goal is to construct a graph with nodes as metrics and a directed edge between two nodes showing the direction and magnitude of the causal effect. 
In our study, we aim to learn the graphical causal models in the form of structural causal models (SCM) and probabilistic causal models (PCM) from MDS data to respectively predict and identify causal relationships among atoms and perform root cause analysis based on observational data. We propose a causal model-based root cause analysis (RCA) in MDS, leveraging causal modeling to uncover causal relationships in spatio-temporal MDS. Our approach identifies a minimal set of variables whose distributional changes explain the HB break/separation observed in the data. In essence, we summarize the key contributions of our paper below:

\begin{itemize}
\item We develop a graphical causal model that infers causal relationships across different samples of spatio-temporal MDS with varying causal graphs while using a shared dynamic model to represent the behavior of molecular dynamics simulations (MDS). This approach is inspired by prior works~\cite{kipf2018neural, lowe2022amortized, liu2022structural} and employs variational autoencoders (VAE). The encoder produces two outputs:
\begin{itemize}
    \item The skeleton graph (SCM), which describes the data generation process of the observed trajectories, and the corresponding causal relationships can be represented via the dynamic Bayesian networks (DBNs).
    \item The PCM, which describes the probability of the edge type (HB vs Separation) occurring.
\end{itemize}
\item The PCM is used to address a root cause problem, answering questions like: "What atomic structures influenced the separation of an HB over time?" through key assumptions:
\begin{itemize}

\item The network structure and dependencies are consistent over time, assuming the stationarity of the system.
\item Some atomic arrangements and structures affect bonding/separation between atoms.
\item The observed changes in marginal or conditional distributions of the atoms forming an HB indicate when an HB forms and persists or separates.
\end{itemize}
\item We quantify the probability of a variable's state given its parent variables in the DAG~\cite{janzing2019causal, spirtes2001causation}.
\item We demonstrate that our model effectively identifies the causal factors driving changes in the MDS.
\item Our method is generic and has been validated using MDS under two different phenomena. The results show that our approach can predict the dynamics and identify the root causes of changes in the data.

\end{itemize}




\section{Preliminaries}
A graph consists of vertices (nodes) and edges that connect pairs of vertices. In this work, the vertices represent atoms, and the edges will denote a causal relationship between these atoms. A directed acyclic graph (DAG) will be our main focus here because it contains no directed cycles. In a DAG: a {\it root} node has no parents $\text{PA}$, and a {\it sink} node has no children $\text{ch}$; every DAG includes at least one root and one sink; a connected DAG where each node has at most one parent is called a tree, and if each node in the tree has only one child, it is called a chain. These structures help us understand and visualize causation in complex systems, as discussed by Pearl~\cite{pearl2009causality}.

\subsection{Causal Models and Mechanism Changes}\label{sec:pcm}
A graphical causal model (GCM), which aims at representing the true data generation process, consists of a causal DAG, $\mathcal{G}$, encoding direct causal dependencies among variables of a data $R = \X$ and defines a causal mechanism for each variable. The causal mechanism $\PX$ describes the conditional distribution of each variable given its parents in the DAG. Specific cases of a GCM are the SCM and the PCM. A PCM only requires generating samples conditioned on a node's parents, and an SCM model mechanisms as a deterministic functional causal model (FCM) of parents and unobserved noise. The causal mechanism of the SCM consists of sets of endogenous $(R)$ and exogenous $(U)$ variables connected by certain functions $(f)$ that determine the values of the variables in $R$ based on the values of the variables in its parents $PA(R)$. When the variable in the causal graph is a root node, it simply defines its distribution. For a non-root node, a node is modeled as $r(i):=f_i(\PA{i}, u(i))$, where $\PA{i}= \{r(j)\vert \forall r(j) = \PA{i}\in \mathcal{ G}\}$, where $u(i))\in U$ is unobserved noise that is independent of $\PA{i}$. The PCM and SCM are capable of modeling the second rung of causality described by Pearl~\cite{dowhy_gcm,dowhy,pearl2009causality}.

\subsection{Intervention Rung of Causality}
Intervention on a variable is the process of changing the mechanism of that variable. Intervention methods like the randomized controlled trials (RCTs) are captured by the \textbf{do-operator} and destruct the previous mechanism; for instance, \textbf{hard interventions} like $do(r = \hat{r})$ fix $r$ to a specific value $\hat{r}$, removing its dependence on other factors. Alternatively, \textbf{soft interventions} modify, rather than sever entirely, the causal mechanism, allowing the structure to remain intact while changing the mechanism and distribution and, for example, replacing $f_r(\text{PA}(r), u(r))$ with an alternative $f^\prime_r (\text{PA}(r), u(r))$, where $f \neq f^\prime$, leads to a modified causal influence without altering the causal graph itself but changes the mechanism and the conditional distribution, thus aligning with the faithfulness assumption~\cite{ikram2022root, dowhy_gcm, pearl2009causality}.

\section{Root Cause Analysis Framework}
Given a spatio-temporal trajectory data describing molecular dynamics, denoted by a set of variables $R= \X \in \mathbb{R}^{N, T, D}$, where at time $t \in T$, the observed trajectory $\mathbf{r}_t = (r_t(1),\ldots,r_t(N)$ is generated by the following dynamics, modeled by an SCM:
\begin{align} 
r_{t+1}(i) = f_i(\text{PA}(r_t(i)) + u_t(i),
\end{align}
where $T$ is the total time of the trajectory, $N$ is the number of variables (atoms) and $D$ is the number of dimensions in space $(x,y,z)$. We assume that the trajectories of the variables $t + 1$ are only affected by those at time step $t$. Our main aim is to find atomic structures responsible for HB separation in the MDS. This framework is divided into two problems. The first is obtaining the causal model, and the second is finding the root cause variables of HB separation using the estimated causal model. 

\subsection{Deriving Causal Models from Spatio-temporal data}

The Bayesian network (BN)~\cite{dean1989model} uses probability to represent the relationships between variables and their conditional dependencies. It is a DAG and is a popular tool for building an SCM:
\begin{align}
    \mathcal{C}_R = (\mathcal{G}, P_R),
\end{align}
where the general form of a BN's joint distribution is:
\begin{align}\label{eq:jd}
 P_R = \prod_{i=1}^N \PX.
\end{align}
The dynamic Bayesian network (DBN) extends the BN to model temporal dependencies or changes in a system over time. It is essentially a sequence of Bayesian Networks where:
\begin{itemize}
    \item Each time slice in the DBN represents a BN.
    \item The structure of the DBN encodes dependencies both within each time slice and across time slices (temporal dependencies).
\end{itemize}

The joint distribution in a DBN is a product of local conditional probabilities across time slices. 


The DBN is utilized to model the SCM in a probabilistic way~\cite{liu2022structural, lowe2022amortized}, which can then enable us can infer the underlying causal mechanisms that govern the system. The DAG property of the BN and its variants ensures that there are no feedback loops or circular causality, which makes it possible to understand how changes propagate through the system. When performing an intervention, the absence of cycles ensures that the effects of the intervention can be traced in a clear, directional manner without ambiguity. The DBN is modeled with a variational autoencoder (VAE)~\cite{doersch2016tutorial, liu2022structural, lowe2022amortized}, enabling end-to-end learning of both the graph structure and the temporal causal dependencies. Specifically, vertices representations are encoded using GNN layers, providing the input to the network. The encoder within this VAE infers the distribution of causal graphs based on observed data, while the decoder generates predictions of future states by simulating temporal interactions among atoms. We are particularly interested in the outputs of the encoder for the RCA.

\subsubsection{Encoder}
The encoder $q_\phi (a\mid r)$ learns the posterior distribution of the adjacency matrix conditioned on the observed trajectories. It applies a GNN $f_{\text{Enc},\phi}$~\cite{yujia2016gated, gilmer2017neural} to the input, which propagates information across a fully connected graph $\mathcal{G}$. This encoder is specifically designed to capture the causal graph structure and classify edge types between vertices using a softmax layer, allowing it to distinguish interactions. Given input trajectories $R= \X$, the encoder computes the following message-passing operations:
\begin{itemize}
\item Node Embedding:
    \begin{align} \label{eq:enc1}
       m^1(j) = f_{\text{emb}}(r(j))
    \end{align}
\item Edge Embedding (Node-to-Edge):
    \begin{align}
       v \rightarrow e: \quad m^1(i,j) = f_e^1\bigg([m^1(i), m^1(j)]\bigg)
    \end{align}
\item Aggregate Message Passing (Edge-to-Node):
    \begin{align}
       e \rightarrow v: \quad m^2(j) = f_v^1\bigg(\sum_{i \neq j} m^1(i,j)\bigg)
    \end{align}
\item Final Edge Embedding (Node-to-Edge):
    \begin{align}
       v \rightarrow e: \quad m^2(i,j) = f_e^2\bigg([m^2(i), m^2(j)]\bigg)
    \end{align}
\end{itemize}

The edge type posterior is finally modeled as
\begin{align}\label{eq:encl}
q_\phi(a(i,j)\mid r) = \text{softmax}(f_{\text{out}}(m^2(i,j))).
\end{align}

$f_{\text{emb}}$, $f_e^1$, $f_v^1$, $f_e^2$ are multilayer perceptions (MLPs), where every message passing operation is performed by a $2$-layer perceptron. $\phi$ is the set of the parameters of the neural networks in the encoder, and $f_{\text{out}}$ is a dense layer that classifies the edge type. $q_\phi(a(i,j)\mid r)$ generates the probability distribution of each type edge based on the embedding. To backpropagate through the samples of the discrete distribution $q_\phi (a(i,j)\mid r)$, we relax it by adding Gumbel distributed noise~\cite{maddison2016concrete}:
\begin{align}
a(i,j)\sim \text{softmax}(\frac{f_{\text{out}}(m^2(i,j))+g}{\tau}),
\end{align}
where $g \in \mathbb{R}^{N_e}$ is an independent and uniformly distributed vector from the Gumbel distribution $(0, 1)$, and $\tau$ (softmax temperature) is the smoothness of sampling. We consider the possibility that there are $N_e$ different edge types expressing causal relationships.

The NRI encoder's probabilistic outputs can be seen as soft edges in a causal graph: The encoder predicts directed relationships between entities. These represent the strength or confidence in causal relationships. $P(a(i,j) = 1 | R)$ indicates the probability that node $i$ causally influences node $j$ with a certain probability. With the learned binary causal structure $E$, we have the skeleton of the DBN, and we are able to estimate the edge strength of each path in the DBN.    We have three edge types, indicating different dependencies below:
\begin{itemize}
    \item Type 1: Non-causal edges $e_0$;
\item Type 2: Causal edges with HB $e_1$;
\item Type 3: Causal edges indicating hydrogen bond separation $e_2$.
\end{itemize}

\subsubsection{Decoder:}
The decoder, $p_{\theta}(\mathbf{r}_{t+1}\mid \mathbf{r}_1,\ldots \mathbf{r}_t, a)$, learns the future states in the system based on both the current state and the inferred causal structure. This dynamics is Markovian for causal structures, such that $p_{\theta}(\mathbf{r}_{t+1}\mid \mathbf{r}_1,\ldots \mathbf{r}_t, a) =  p_{\theta}(\mathbf{r}_{t+1}\mid \mathbf{r}_t, a)$. The decoder models the temporal dynamics of the system, using GNNs to parameterize this distribution. By taking the predicted edge types $a(i, j)$ and the feature vectors of the time series at time $t$, $\mathbf{r}_t$, as input, the decoder is able to first propagate information along the predicted edges determined by the inferred causal graph: 
\begin{itemize}
\item Edge Embedding (Node-to-Edge):
\begin{align} \label{eq:dec1}  
v \rightarrow e: \quad \hat{m}_t(i,j)=\sum_{e}a(i,j,e) f_d^e\bigg([r_t(i), r_t(j)]\bigg). \end{align}

The neural network $f_d^0$ is hard-coded as zero as it is not considered relevant in the prediction process. The decoder then accumulates the incoming messages to each node and applies a neural network $f^d_v$ to predict the next time-step:

\item Message Passing (Edge-to-Node):
\begin{align}
  e \rightarrow v: \quad & \mu_{t+1}(j)   = r_t(j)+ f^d_v \bigg( \sum_{i\neq j}\hat{m}_t(i,j)\bigg) \label{eq:fdec}\\
  & = f_{\text{Dec}}(r_t(j)) 
\end{align}

$\mu_{t+1}(j)$ is the predicted state for node $j$ at the next time-step $t+1$, which incorporates both the cumulative messages from other nodes (weighted by their causal connection types) and the current state $r_t(j)$. This process enables the model to forecast the system's evolution in a way that respects the inferred causal structure. $f_d^e$ and $f^d_v$ are also each represented by a $2$-layer perceptrons. 
To refine the modeling of temporal dynamics, the decoder employs two separate GNNs  to handle different types of causal interactions:
  \begin{itemize}
     \item $ f^{2}_d$ models interactions for edge type $2$.
     \item $ f^{3}_d$ models interactions for edge type $3$.
 \end{itemize}

 The VAE framework is shown in Figure~\ref{fig:RCA2}.

\begin{figure*}[!ht]
\centering
{\includegraphics[width=\textwidth]{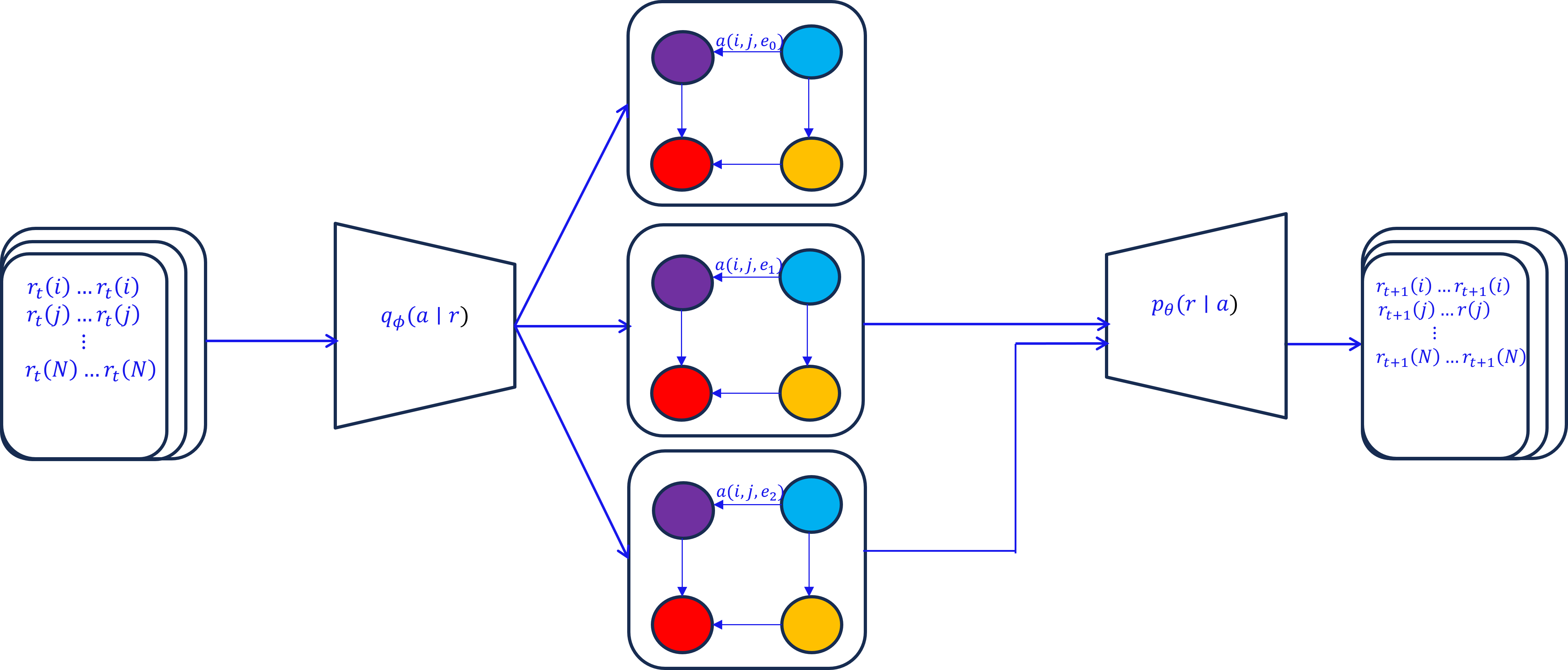}}
\caption{Variational autoencoder}
\label{fig:RCA2}
\Description{RCA}
\end{figure*} 
To predict multiple steps into the future, in the multi-step prediction, $\mathbf{r}_t$ is replaced with the predicted mean $\mu_t$ for $k$ amount of steps; then the actual previous step is reintroduced. Formally from Eq.~(\ref{eq:fdec}):

\begin{align} \label{eq:decl} 
\mu_2(j) = f_{\text{Dec}}(r_1(j)) \\
\mu_{t+1}(j) = f_{\text{Dec}}(\mu_t(j)), \quad t=2, \ldots, k
\end{align}
\end{itemize}

 The errors are compounded over $k$ steps, making a degenerate decoder significantly optimal. This design ensures that the decoder remains sensitive to the learned latent edges.

\subsubsection{Loss functions and regularization}
In the encoder, the KL-divergence~\cite{rezende2014stochastic} between the posterior and the prior is included in the loss function as 

\begin{align} \mathcal{L}_{\text{KL}} &= \mathcal{D}(q_{\phi}(a\mid r) \parallel p_\theta(a))\\
&= \sum_{i\neq j} H(q_\phi (a(i,j)\mid r)), \end{align}
where $\mathcal{D}$ is the KL-divergence and $H$ is the entropy.
In the decoder, the loss for the temporal dynamic prediction is
described by
\begin{align} \mathcal{L}_{\text{Dec}} &= \text{MSE} (\mathbf{r}_{t+1}, p_\theta (\mathbf{r}_{t+1}\mid \mathbf{r}_t, E))\\
&\frac{1}{T-1} \parallel \mathbf{r}_{t+1} -\mathbf{\hat{r}}_{t+1}\parallel^2_2,\end{align}
where $\mathbf{\hat{r}}_{t+1}$ is the predict value of $\mathbf{r}_{t+1}$.
Regularization is added to enforce sparsity in the causal mechanisms~\cite{scholkopf2021toward}. we used the $L_1$ sparsity constraint
\begin{align}\label{eq:reg}
     \mathcal{L}_{\text{PCM}} = \|  a(e_1) + a(e_2) \| _1
     \end{align}
for prediction and the group lasso
\begin{align}\label{eq:reg2}
\mathcal{L}_{\text{PCM}} =\sum_{i,j} \|  a(i,j,e_1) + a(i,j,e_2) \| _2
\end{align}
for root cause analysis. 

The total loss function will consist of three parts:
\begin{align}\label{eq:lossRCA} \mathcal{L} &= \lambda_1\mathcal{L}_{\text{KL}}+  \lambda_2 \mathcal{L}_{\text{Dec}}+  \lambda_3\mathcal{L}_{\text{PCM}},  \end{align}
where  $[\lambda_1, \lambda_2,  \lambda_3]$ are tunable hyprerparameters greater than $0$.

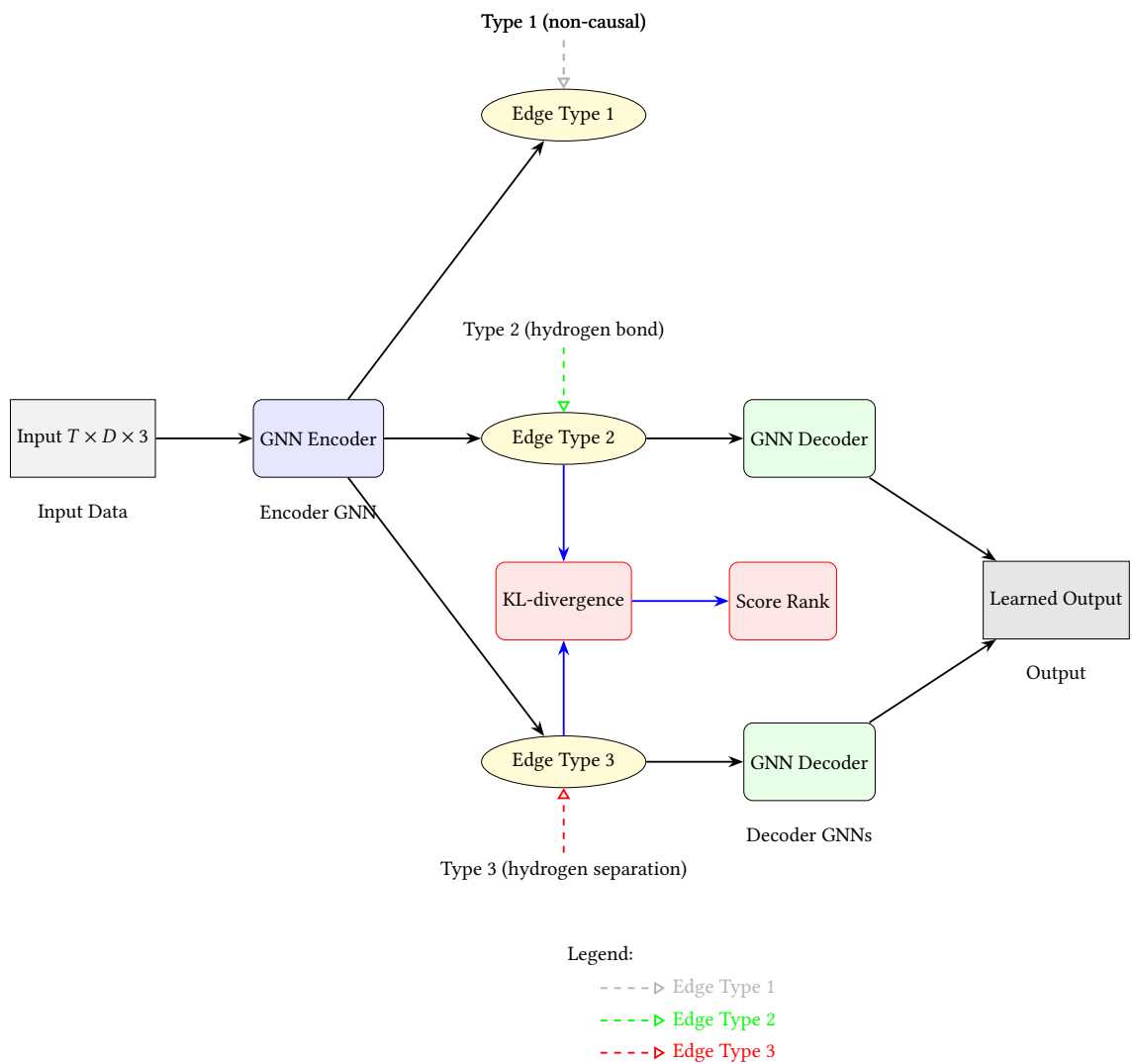
\begin{figure}[!ht]
\centering
 \resizebox{\columnwidth}{!}{
\begin{tikzpicture}[
    input_node/.style={rectangle, draw=black, fill=gray!10, minimum width=1.5cm, minimum height=1.2cm},
    gnn_block/.style={rectangle, draw=black, fill=blue!10, rounded corners, minimum width=1.2cm, minimum height=1.2cm},
    latent_block/.style={ellipse, draw=black, fill=yellow!20, minimum width=1.8cm, minimum height=0.8cm},
    decoder_block/.style={rectangle, draw=black, fill=green!10, rounded corners, minimum width=1.2cm, minimum height=1.2cm},
    output_node/.style={rectangle, draw=black, fill=gray!20, minimum width=1.5cm, minimum height=1.2cm},
    edge_type/.style={thick, dashed, -{Triangle[open]}},
    arrow/.style={-Stealth, thick},
    kl_block/.style={rectangle, draw=red, fill=red!10, rounded corners, minimum width=1.2cm, minimum height=1.2cm},
]

\node[input_node] (input) at (0,0) {Input $T \times D \times 3$};
\node[gnn_block, right=1.5cm of input] (encoder1) {GNN Encoder};
\node[latent_block, right=1.5cm of encoder1] (latent2) { Edge Type 2};
\node[latent_block, right=1.5cm of encoder1, yshift=5cm] (latent1) { Edge Type 1}; 
\node[latent_block, right=1.5cm of encoder1, yshift=-5cm] (latent3) { Edge Type 3}; 
\node[decoder_block, right=1.5cm of latent2 ] (decoder1) {GNN Decoder};

\node[kl_block, below=1.5cm of latent2 ] (kl) {KL-divergence};
\node[kl_block, right=1.5cm of kl ] (klr) {Score Rank};
\node[decoder_block,right=1.5cm of latent3] (decoder2) {GNN Decoder};
\node[output_node, right=5.2cm of latent2, yshift=-2.5cm] (output) {Learned Output};

\draw[arrow] (input) -- (encoder1);
\draw[arrow] (encoder1) -- (latent1);
\draw[arrow] (encoder1) -- (latent2);
\draw[arrow] (encoder1) -- (latent3);
\draw[arrow] (latent2) -- (decoder1);
\draw[arrow] (latent3) -- (decoder2);
\draw[arrow] (decoder1) -- (output);
\draw[arrow] (decoder2) -- (output);
\draw[arrow, color=blue] (latent3) -- (kl);
\draw[arrow, color=blue] (latent2) -- (kl);
\draw[arrow, color=blue] (kl) -- (klr);

\node[above=0.75cm of latent1] (type1) {Type 1 (non-causal)};
\draw[edge_type, color=gray!60] (type1) -- (latent1);

\node[above=1cm of latent2] (type2) {Type 2 (hydrogen bond)};
\draw[edge_type, color=green] (type2) -- (latent2);

\node[below=1cm of latent3] (type3) {Type 3 (hydrogen separation)};
\draw[edge_type, color=red, dashed] (type3) -- (latent3);

\node[above=0.75cm of latent1] (type1) {Type 1 (non-causal)};
\draw[edge_type, color=gray!60] (type1) -- (latent1);

\node[below=0.3cm of input] {Input Data};
\node[below=0.3cm of encoder1] {Encoder GNN};
\node[below=0.3cm of decoder2] {Decoder GNNs};
\node[below=0.3cm of output] {Output};

\begin{scope}[shift={(8, -8)}]
    \node (legend) at (0,0) {Legend:};
    \draw[edge_type, color=gray!60] (0, -0.5) -- (1, -0.5) node[right] {Edge Type 1 };
    \draw[edge_type, color=green] (0, -1) -- (1, -1) node[right] {Edge Type 2 };
    \draw[edge_type, color=red, dashed] (0, -1.5) -- (1, -1.5) node[right] {Edge Type 3 };
\end{scope}

\end{tikzpicture}}
\caption{Root Cause Analysis Framework}
\label{fig:RCA1}
\Description{Root Cause Analysis framework}
\end{figure}

\subsection{Root Cause Analysis}
We can infer the causal graphs via encoders and predict the future trajectories of all atoms using the decoder as described in subsection~\ref{sec:pcm} and shown in Figure~\ref{fig:RCA1}. The encoder outputs the PCM of edge types $2$ and $3$, which we are interested in extracting to answer a critical question  If an HB separation has occurred, which atomic structures changed for this bond to happen? by making these critical assumptions:
\begin{itemize}
    \item  The network structure and conditional dependence are the same as in other time steps, which assume the stationary property of the system. 
\item It is suspected that the arrangement/structure of other atoms affects the bonding of the two atoms. 
\end{itemize}
We consider the formation/separation of HBs as an intervention on the root cause nodes. Doing so allows us to use distributional invariances within the observational and also across observational and interventional data for learning. In this work, we aim to find the variables responsible for any causal mechanism change in the PCMs. When an HB persistence to separation occurs at any time $t$, we conclude that at those times $t$, an intervention has been performed on $R$. The idea behind this method is to systematically learn the causal mechanisms $P_R = a(i,j,e_1)$ and $\tilde{P}_R = a(i,j,e_2)$ using the model on the dataset since this dataset cannot be separable. This model learns both PCMs at a go. Our goal is to identify the mechanisms that have changed, which would lead to a different marginal/conditional distribution of the target.  In contrast, unchanged mechanisms would result in the same distribution. Note that a change in the mechanism could be due to a functional change in the underlying model or a change in the (unobserved) noise distribution. However, both changes would lead to a change in the mechanism\cite{dowhy_gcm}.

Given a probabilistic causal model, the causal Markov assumption allows us to factorize the joint distribution $P_R$ into causal mechanisms $\PX$ at each node $r(i)$. Each causal mechanism remains invariant to interventions in other variables. With this, we can formally define what causal mechanism changes entail.

\begin{d1}{Mechanism Changes}
Mechanism changes to a causal model $\mathcal{C}:= (\mathcal{G},P_R)$ on a subset of atoms $R$ of an MDS when HB persists indexed by a change set $S\subseteq [1, \ldots, p]$, transform $\mathcal{C}$ into $\tilde { \mathcal{C}}:= (\mathcal{G},\tilde{P}_R)$ when this bond has separated,  where
\begin{align}
  \tilde{P}_R  = \prod_{i\in S} \TPX\prod_{i\notin S}  P(r(i) \mid \PA{i})
\end{align}
 is a new joint distribution obtained by replacing the edge type $a(i,j,e_1)$ causal mechanism at each node $r(i)$, where $i \in S$, with the edge $a(i,j,e_2)$ causal mechanism~\cite{dowhy_gcm}.
\end{d1}

The $\tilde{P}_R$ can also be seen as the post-intervention joint distribution $P_R^{\text{do}(S)}$, where $\text{do}(S)$ represents mechanism changes through stochastic intervention at each node $r(i)$ indexed by $S$~\cite{correa2020calculus}. To assess the contribution  of each variable $r(i)$ on the change from $P_R$ to $\tilde P_R$ describe by the KL-divergence, $\mathcal{D}$
        \begin{align}\label{eq:rca}
        \mathcal{D}(\tilde{P}_R \parallel P_R) &= \sum_{i=1}^N \mathcal{D}\bigg( \tilde P(r(i) \mid \PA{i}) \parallel  P(r(i)\mid \PA{i}) \bigg)     \\
   &= \sum_{r\in R} \tilde{P}_R(r)\log \bigg(\frac{\tilde{P}_R(r)}{P_R(r)}\bigg),
        \end{align}
we find the KL-divergence from its causal mechanism $ P(r(i)\mid \PA{i}) $  to  $\tilde P(r(i)\mid \PA{i})$.

\begin{d1}{Root Cause Analysis Problem}\label{def:2}
Suppose that the causal mechanism of atom $r(i)$ changes from when HB persists $P(r(i)\mid \PA{i}) $  to $\tilde P(r(i)\mid \PA{i})$, when this bond has separated. Then the contribution of atom $r(i)$ to the KL-divergence from the joint distribution $P_R$ to $\tilde P_R$ is 
\begin{align}
    \mathcal{D}(\TPX\parallel  P(r(i)\mid \PA{i}).
\end{align}
\end{d1}
We can sort these quantities in descending order of magnitude to know the level of impact each constituent variable has on the overall distribution change. The KL-divergence is additive for the independent compositions of the joint distribution. That is, by generalizing the chain rule to more than two variables using the causal Markov condition, we get an additive decomposition of the KL-divergence from the joint distribution~\cite{cover1999elements}.  The summary framework is summarized in Algorithm~\ref{algm:RCA}.








\begin{algorithm}
\begin{algorithmic}
\REQUIRE Spatio-temporal data, $R \in \mathbb{R}^{N\times T \times D}$,   $\lambda_i >0$, Learning rate $1\geq\eta > 0$\\
Initialize: $f_{\text{Enc}, \phi}, f_{\text{Dec}, \theta}$

\WHILE{not converged}
\STATE compute the PCMs and SCMs for $3$ edge types as in Eq.~(\ref{eq:enc1}) to Eq.~(\ref{eq:encl})
\STATE Predict $\{\hat{\mathbf{r}}_{t+1}\}$ as in Eq.~(\ref{eq:dec1}) to Eq.~(\ref{eq:decl})
\STATE compute loss $\mathcal{L}$ as in Eq.~(\ref{eq:lossRCA})
\STATE Update parameters $\phi,\theta,C$
\ENDWHILE\\
\STATE compute KL-divergence of $a(i,j,e_1)$ and $a(i,j,e_2)$ for each variable as in Eq.~(\ref{eq:rca})
\STATE Aggregate sort KL-divergence outputs
\RETURN $\phi,\theta, a$ and sorted KL-divergence outputs  
 \end{algorithmic}\caption{Root cause analysis algorithm}\label{algm:RCA}
 \Description{RCA}

\end{algorithm}

\section{Results and Evaluation}
All training procedures utilize the Adam optimizer~\cite{kingma2014adam} with an initial learning rate of $\eta$, which was decayed by a factor of $0.1$ every $200$ epoch. The concrete distribution with a temperature parameter $\tau = 0.5$ was employed during training. For testing(simulations), the concrete distribution was replaced with a categorical distribution. Each experiment for testing was conducted for $300$ epochs, while that of RCA was performed for $100$ epochs. Model checkpoints were saved after every epoch where the validation performance, evaluated using path prediction mean square error (MSE), improved. The best-performing model based on validation results was subsequently loaded for testing. The hyperparameters used in the experiments are summarized in Table~\ref{tab:hyperrca}.

\begin{table}[!htb] 
\centering
\caption{Hyperparameters used in training}
 \label{tab:hyperrca}
\begin{tabular}{|| c |  c| c||} 
 \hline 
  Parameter& Value  & Value\\
  & Prediction Training & RCA training \\ \hline \hline 
$\tau $&0.5 &0.5\\  \hline
$\eta$&0.00005&0.0005\\  \hline
Prior &$[0.2, 0.4, 0.4]$&$[0.9, 0.05, 0.05]$\\  \hline
Prediction steps (k)&T &3\\ \hline
$\lambda_1$ & 1& 1\\ \hline
$\lambda_2$ &0.1&0.1\\ \hline
$\lambda_3$ &0.001&0.001\\ \hline
 \end{tabular}
\end{table}

The architecture details of the encoder and decoder are outlined in Table~\ref{tab:archi}.
\begin{table}[hbt!] 
\centering
 \caption{Encoder and decoder architecture}
 \label{tab:archi}
\begin{tabular}{|| c | c| c| c |c||} 
 \hline 
  \multirow{4}{*}{Encoder} &Node embeddings & Layers numbers &Activations \\ \hline \hline
 &f$_\text{emb}$ &$128$-$128$ &ELU-ELU+BatchNorm \\  
&$f_e^1$&$128$-$128$ &ELU-ELU+BatchNorm\\ 
&$f_v^1$ &$128$-$128$  &ELU-ELU+BatchNorm \\ 
&$f_\text{out}$& $128$& -\\
\hline
\multirow{2}{*}{Decoder} &$f^e_d$ & $64$-$64$ &ReLU-ReLU \\ 
&$f_v^d$ & $64$-$64$ &ReLU-ReLU \\ \hline
 \end{tabular}
\end{table}

\subsection{Dataset, Chemistry Background \& Preprocessing}
\begin{figure}[!htb]
\centering
\includegraphics[width=0.6\linewidth]{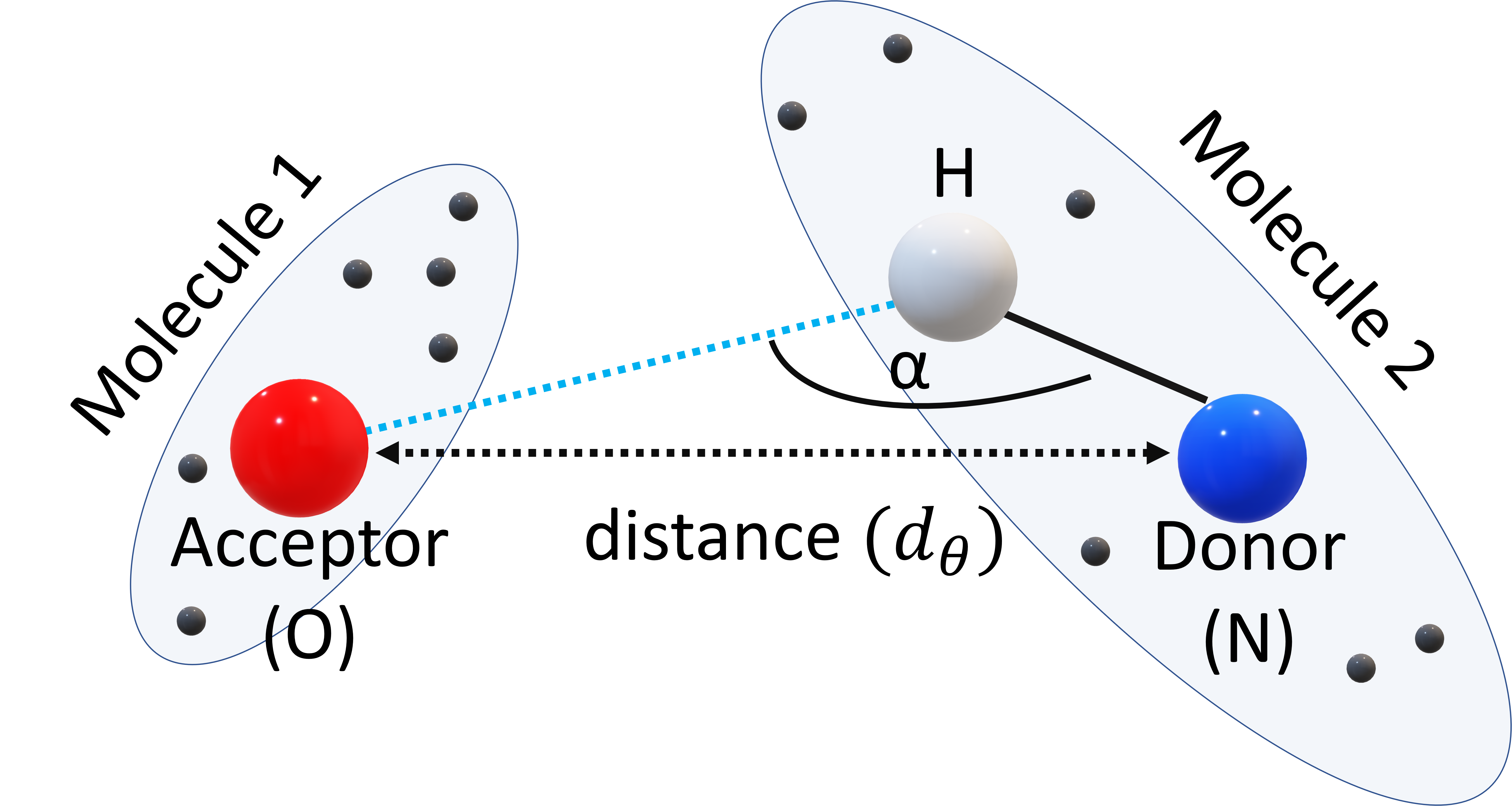}
\caption{
    HB formation~\cite{ADBIS2022}: donor atom (blue) and H atom (gray);  acceptor atom (red).}
    \Description{HB formation: donor atom (blue) and H atom (gray); acceptor atom (red).}
\label{fig:hbond}
\end{figure}

The atomic trajectories in this study were obtained from our earlier work~\cite{Wang2019, Wang2020}, which used MDS for chiral separation. Chiral separations consume almost 80\% of all energy consumption in the chemical process industry, and the application field includes separations of drug enantiomers, liquid extractions, aqueous solutions, and so on. The atomic trajectory data contains the motion and interaction between the atoms of the drug molecules and atoms from the polymer. The significance of this dataset is that it enables identifying critical chemical phenomena -- in particular, the formation of HBs and their persistence over time. HB is a type of intermolecular force that occurs when a hydrogen (H) atom covalently bonded to a highly electro-negative atom such as nitrogen (N) or oxygen (O) interacts with a second atom that has a lone pair of electrons, such as another N, O, or fluorine (F). Figure~\ref{fig:hbond} shows the process of HB formation.

Our analysis focuses on drug molecule atoms located within $12$ Angstroms (\r{A}) of HB. Two phenomena are examined: Tracking a single set of HB atoms and tracking two sets of HB atoms. The autoencoder training process consisted of three steps: (i) We first calculated $q_\phi$ based on a training MD trajectories $R$.; (ii) we then sampled $a(i,j)$ from a continuous approximation of the discrete distribution, and (iii) we reconstructed the trajectories from the sampled distributions. For prediction, the aim is to predict the next $T-1$ future movements of the atoms, represented as $\hat{\mathbf{r}}_{t+1:t+T-1}= \hat{\mathbf{r}}_{t+1}, \hat{\mathbf{r}}_{t+2}, \dots, \hat{\mathbf{r}}_{t+T-1}$, given their current states $\mathbf{r}_t$ . The objective is to minimize the distance between the simulated trajectories and the corresponding ground truth data. We validate the approach on long-duration MD simulations by predicting the states $\mathbf{r}_{2:T}$ while only observing the initial state $\mathbf{r}_1$. During training, the model is provided with collected MD trajectories, where each trajectory corresponds to a specific atom over $T$ time steps. The autoencoder is trained using Eq.~(\ref{eq:decl}) over $T-1$ time steps by observing the initial states $\mathbf{r}_1$ of all atoms and predicting their future states based on inferred causal relationships and temporal dynamics. At time $t > 2$, it uses the estimated state $\hat{\mathbf{r}}_{t-1}$ from the previous time step to predict the next state. To evaluate the robustness of the model to our data, sequence length $T$ was varied over $[5, 10, 20, 25, 50, 75, 100, 350]$. During testing, the trained model predicts the entire trajectory by observing only the initial state.  

For the prediction, three evaluation metrics are used:
\begin{enumerate}
    \item Displacement: We compute the displacement with respect to the first time step $\mathbf{r}_t$ for each trajectory and each time sampled in our simulation:
\begin{align}
\mathbf{D}_t = \sqrt{(\mathbf{r}_t(x) - \mathbf{r}_0(x))^2 + ((\mathbf{r}_t(y) - \mathbf{r}_0(y))^2 + (\mathbf{r}_t(z) - \mathbf{r}_0(z))^2},
\end{align}

where: $\mathbf{r}_0(x), \mathbf{r}_0(y),  \mathbf{r}_0(z)$ are the coordinates at the first time step; $\mathbf{r}_t(x), \mathbf{r}_t(y), \mathbf{r}_t(z)$ are the coordinates at the $t$-th time step.
\item Root mean square fluctuation (RMSF): We compute the RMSF with respect to all atoms at each time $t$ to measure the spread out of the data. The formula for RMSF at time $t$ is

\begin{align}
\text{RMSF}_t = \sqrt{\frac{1}{N} \sum_{i=1}^{N} \parallel r_t(i) - \bar{r}(i)\parallel^2},
\end{align}

where $\hat{\mathbf{r}}_i(t)$ is the mean value for the $i$-th atom.

\item Root mean square fluctuation (RMSF): We compute the RMSF with respect to all times $T$ for each atom to quantify the deviation of a particle's position from its average position over time. The formula for RMSF for a single atom $i$ is:
\begin{align}
\text{RMSF}(i) = \sqrt{\frac{1}{T} \sum_{t=1}^{T} \parallel r_t(i) - \bar{r}(i)\parallel^2}.
\end{align}
\end{enumerate}

For RCA, we assume the data is Gaussian and estimated the ground truth root causes using $3$ different distributional metrics, KL-divergence, Wasserstein distance, and the Expectation distance as defined in \cite{adesunkanmi2024expectation}. We estimated the difference in the data when the HB persists and when it breaks. 

\subsection{Phenomenon 1: Tracking a single set of HB atoms}
\begin{figure}[H]
\centering
\subfloat[$T = 350$]{\includegraphics[width=0.3\textwidth]{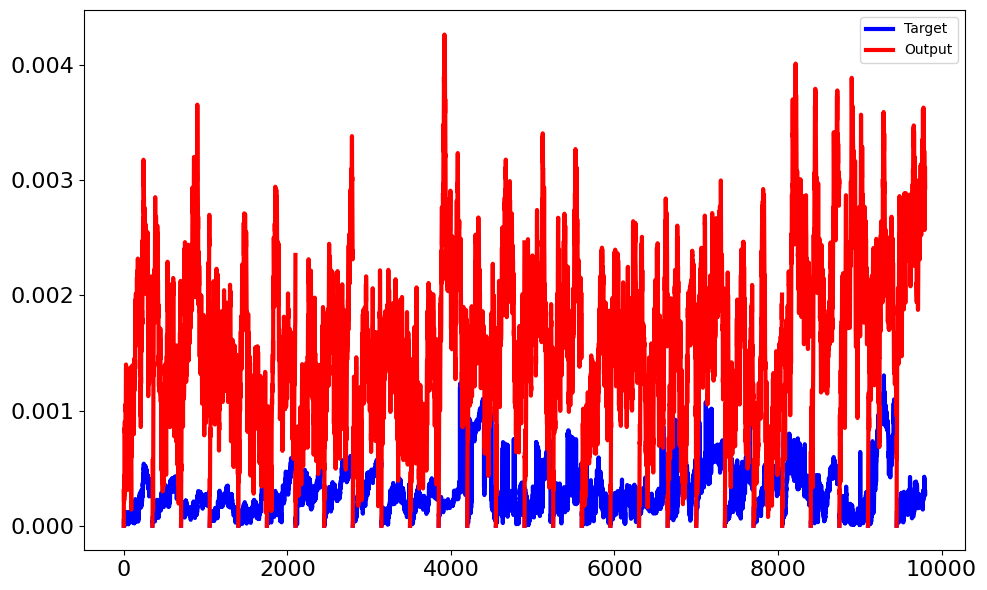}}\hfill
\subfloat[ $T = 100$]{\includegraphics[width=0.3\textwidth]{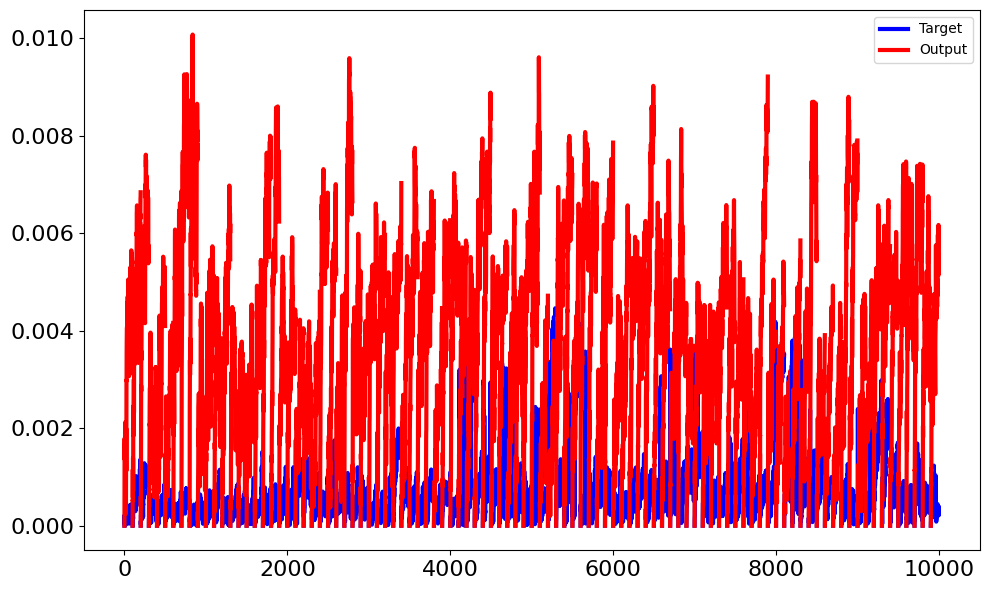}}\hfill
\subfloat[ $T = 70$]{\includegraphics[width=0.3\textwidth]{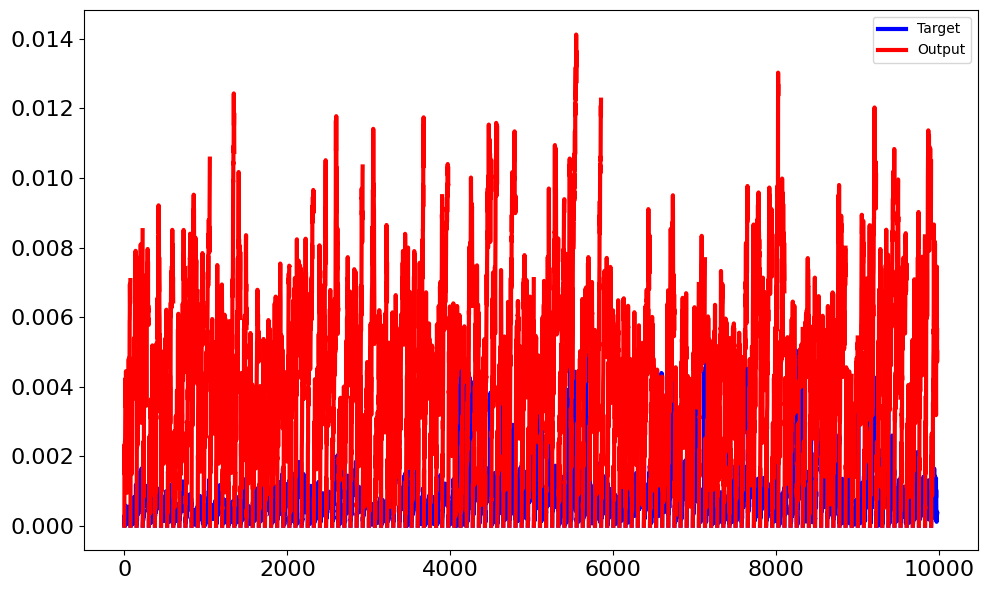}}\hfill
\subfloat[ $T = 50$]{\includegraphics[width=0.3\textwidth]{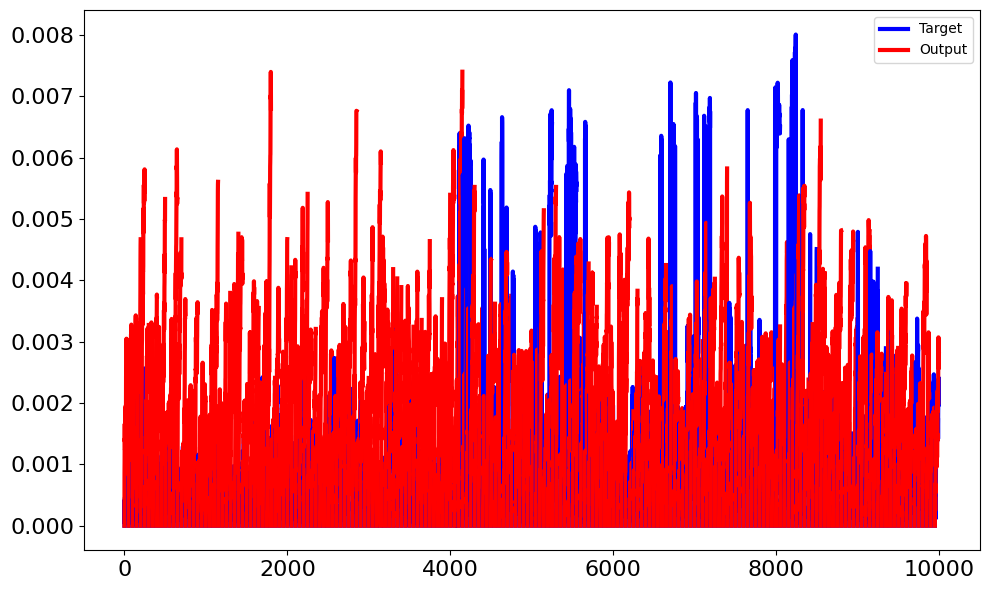}}\hfill
\subfloat[$T = 25$]{\includegraphics[width=0.3\textwidth]{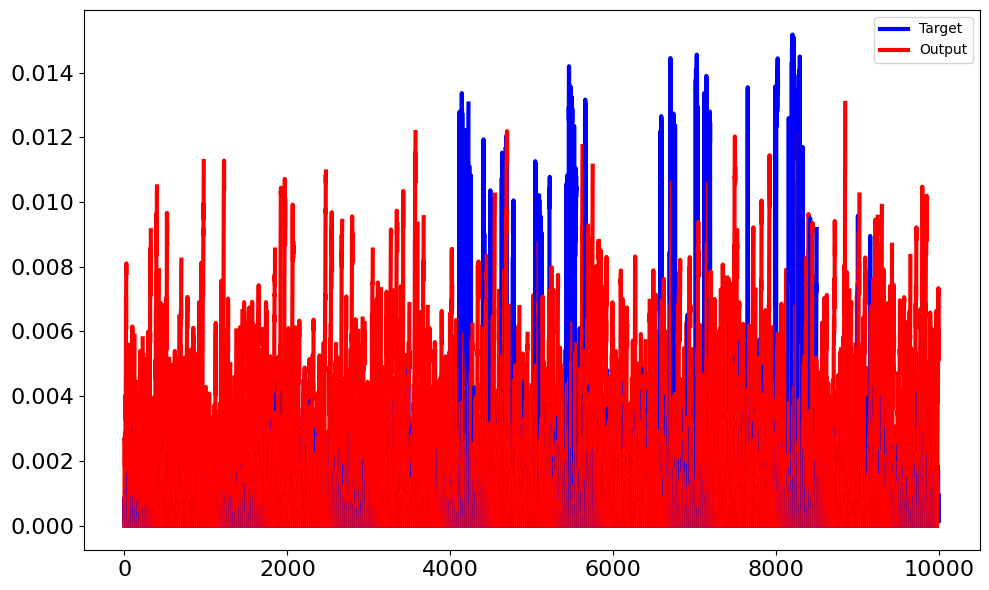}}\hfill
\subfloat[ $T = 20$]{\includegraphics[width=0.3\textwidth]{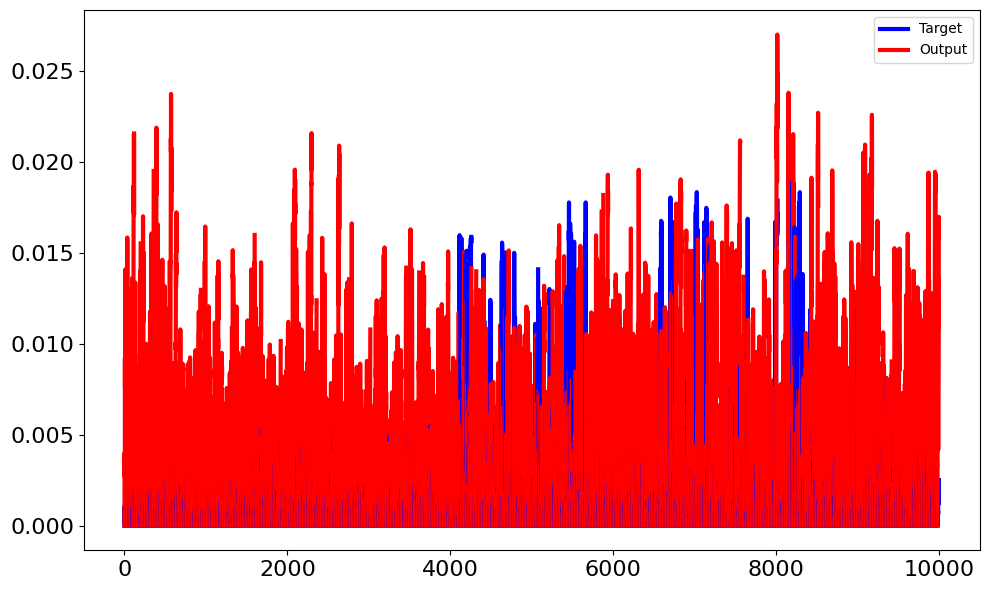}}\hfill
\subfloat[ $T = 10$]{\includegraphics[width=0.3\textwidth]{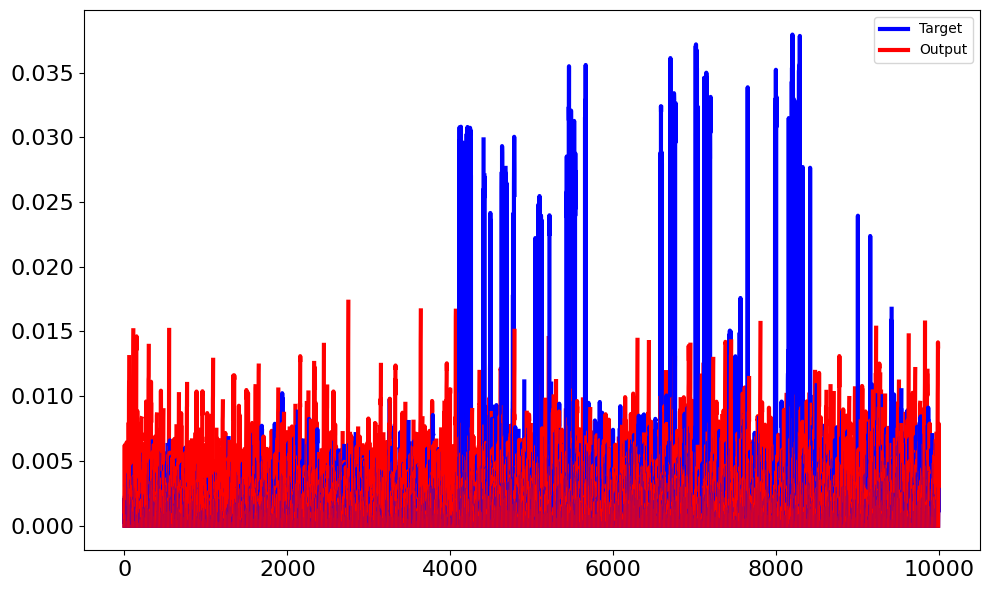}}\hfill
\subfloat[ $T = 5$]{\includegraphics[width=0.3\textwidth]{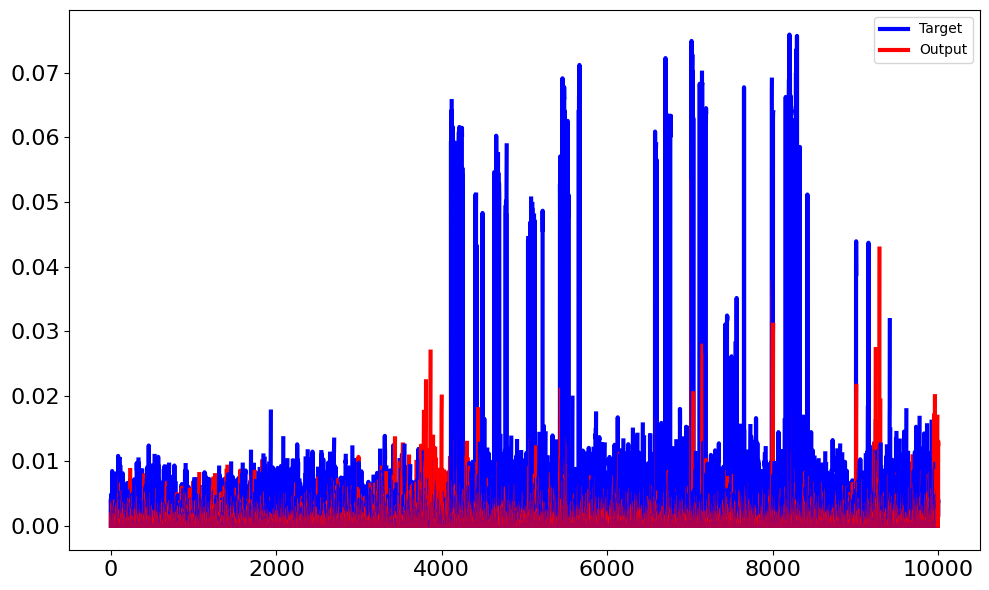}}
\caption{Plots of the displacement between the ground truth(blue) and the predicted(red) values}\label{fig:loss1}
\Description{Plots of the displacement between the ground truth (blue) and the predicted (red) values}
\end{figure}

\begin{figure}[H]
\centering
\subfloat[$T = 350$]{\includegraphics[width=0.3\textwidth]{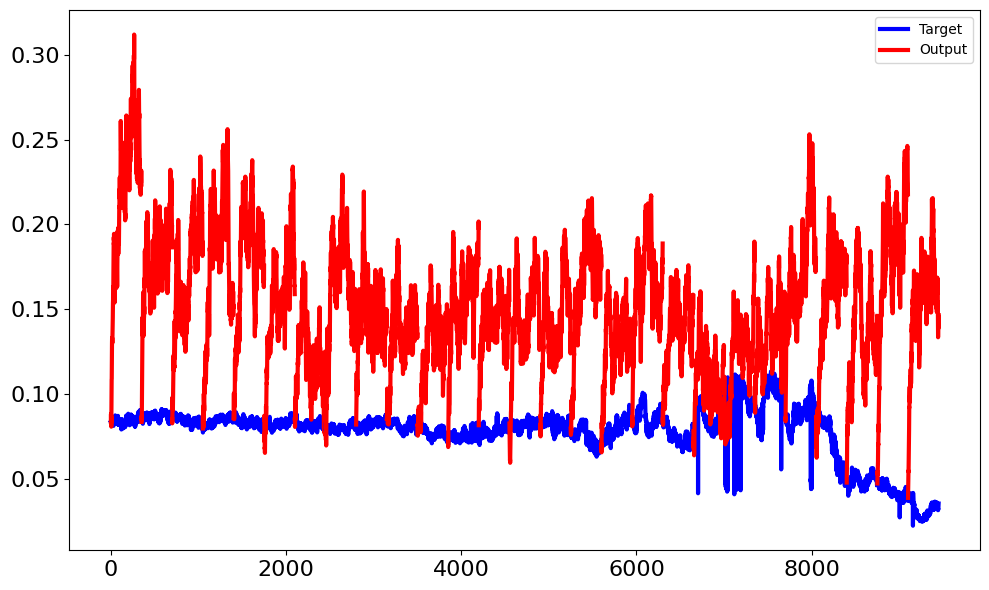}}\hfill
\subfloat[ $T = 100$]{\includegraphics[width=0.3\textwidth]{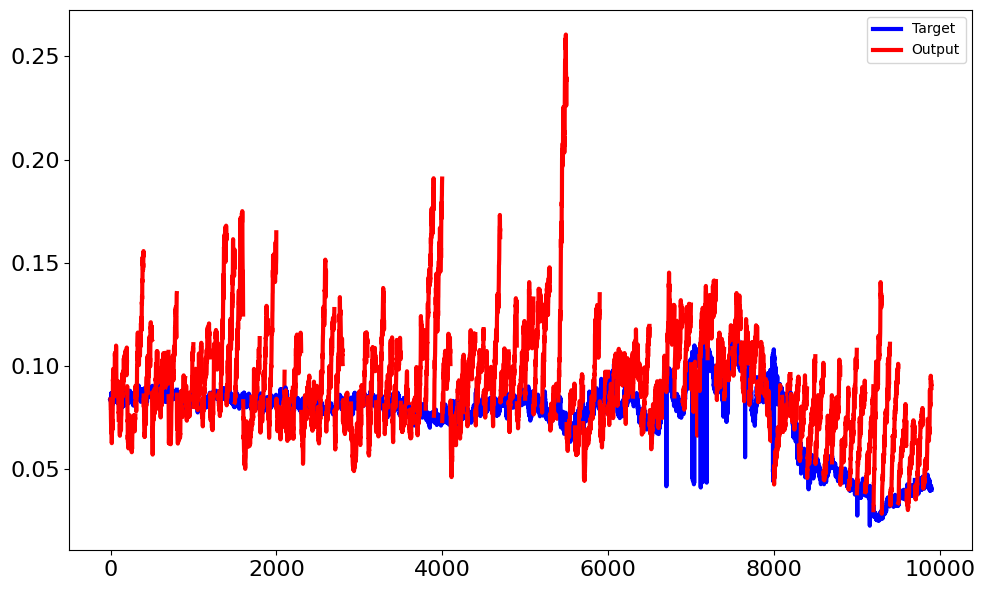}}\hfill
\subfloat[ $T = 75$]{\includegraphics[width=0.3\textwidth]{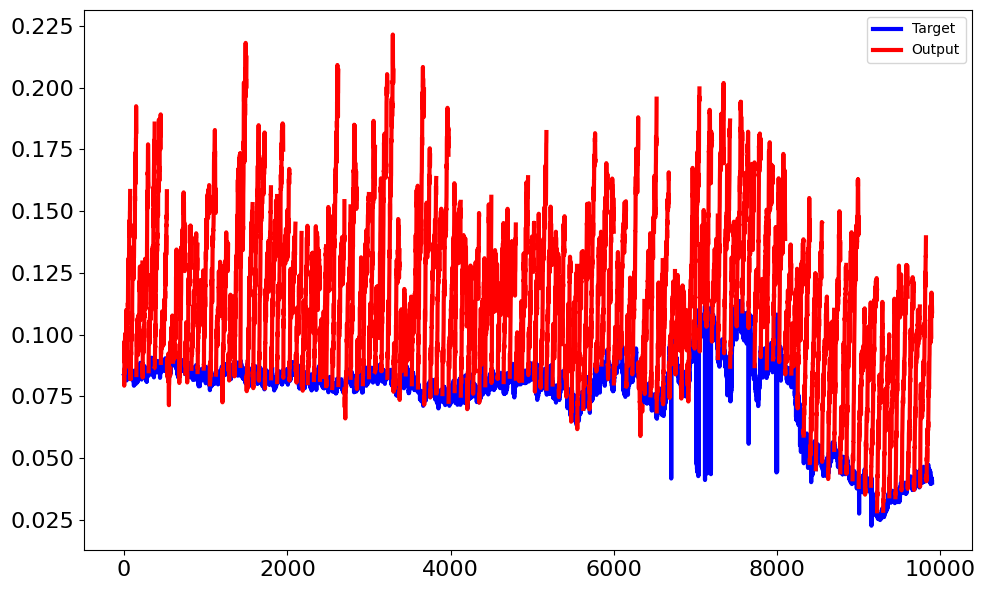}}\hfill
\subfloat[ $T = 50$]{\includegraphics[width=0.3\textwidth]{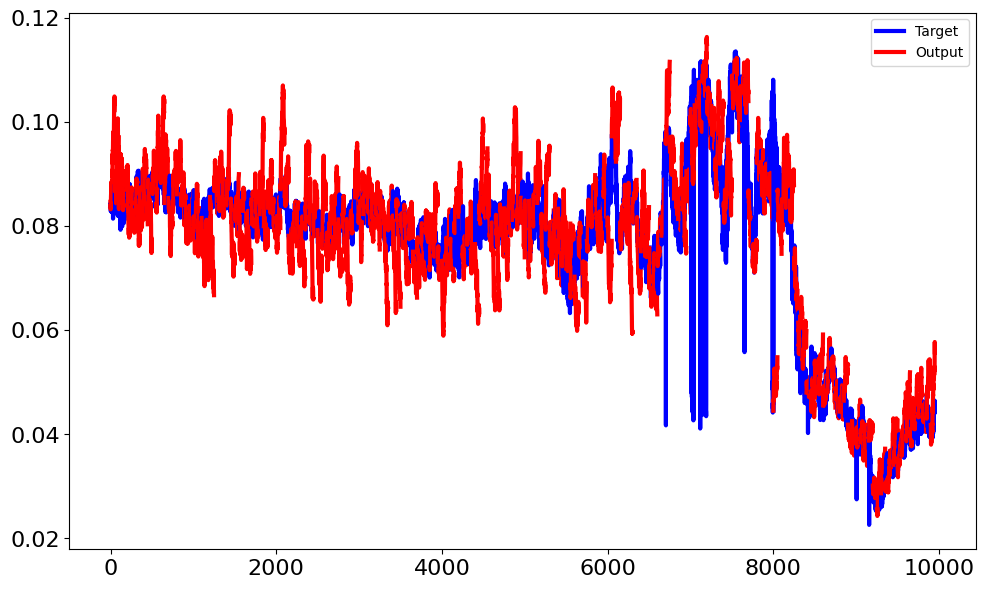}}\hfill
\subfloat[$T = 25$]{\includegraphics[width=0.3\textwidth]{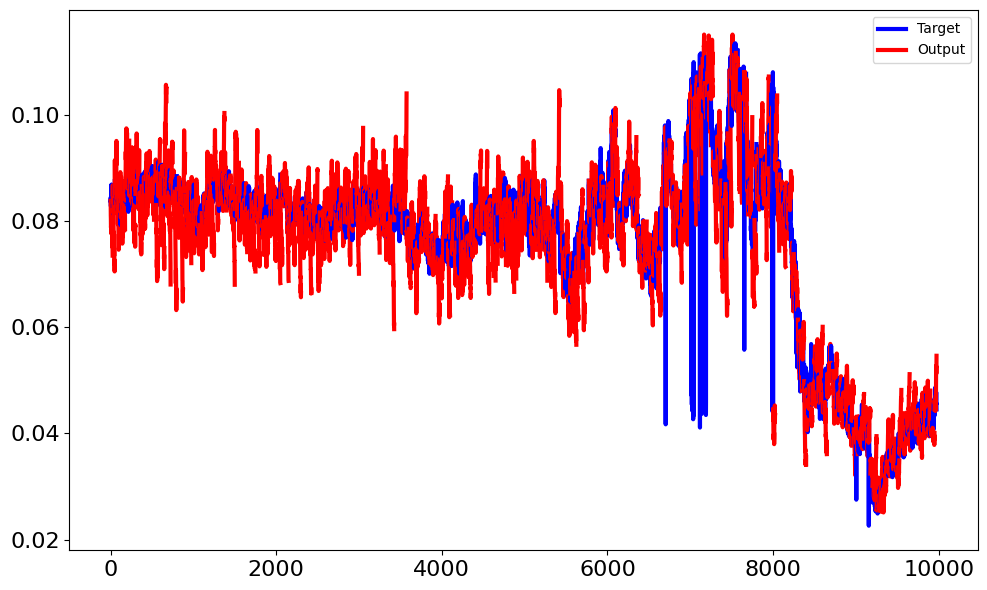}}\hfill
\subfloat[ $T = 20$]{\includegraphics[width=0.3\textwidth]{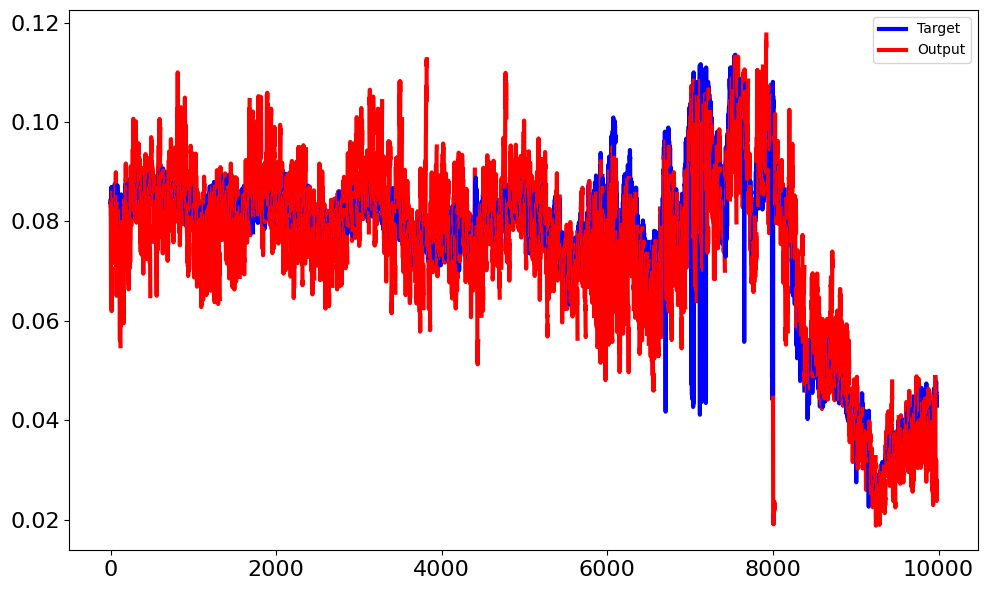}}\hfill
\subfloat[ $T = 10$]{\includegraphics[width=0.3\textwidth]{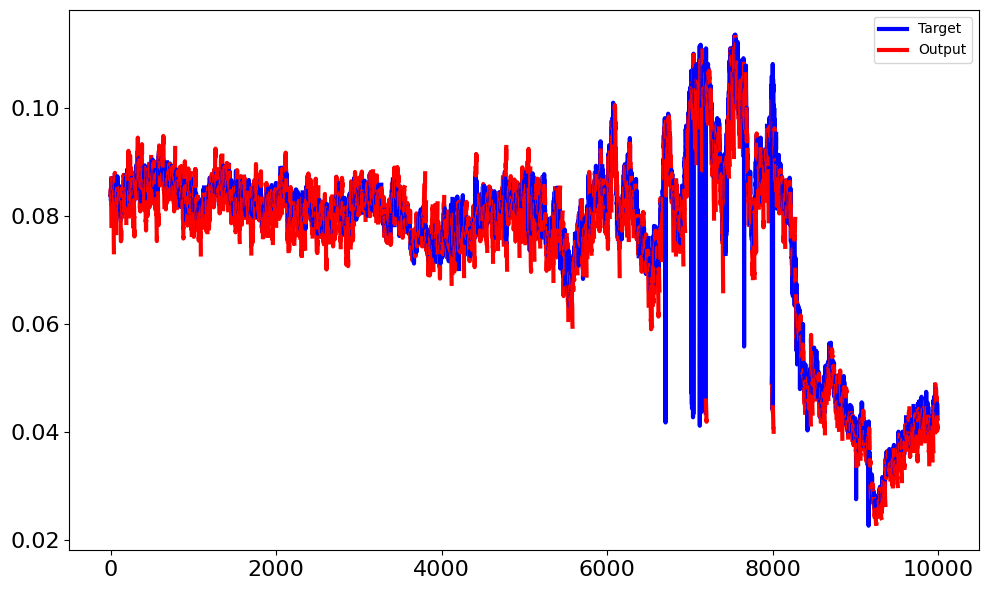}}\hfill
\subfloat[ $T = 5$]{\includegraphics[width=0.3\textwidth]{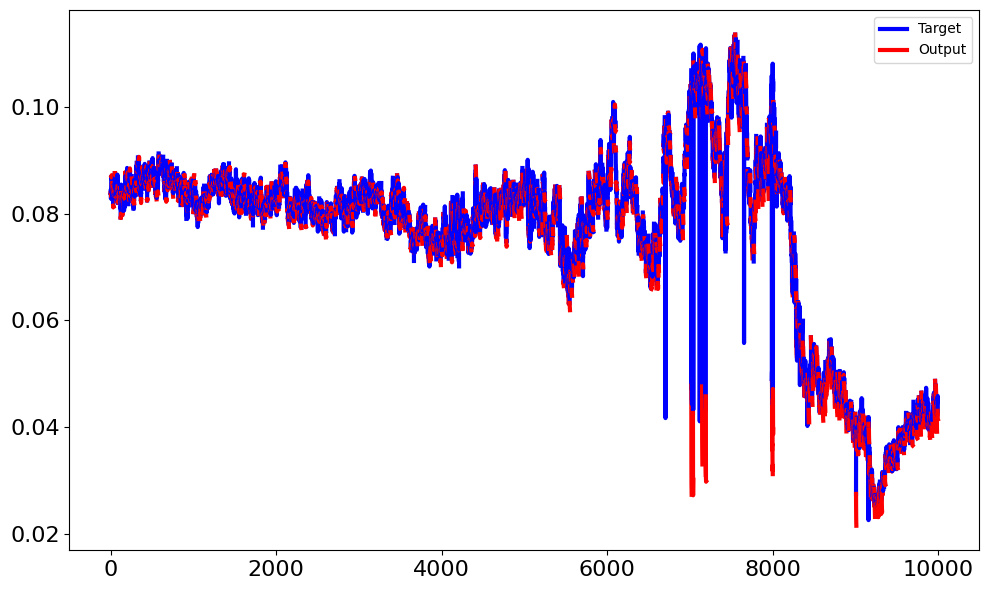}}
\caption{Plots of RMSF with respect to all atoms at each time $t$}\label{fig:R1train}
\Description{Value square deviation plots}
\end{figure}

\begin{figure}[!htb]
\centering
\subfloat[$T = 350$]{\includegraphics[width=0.3\textwidth]{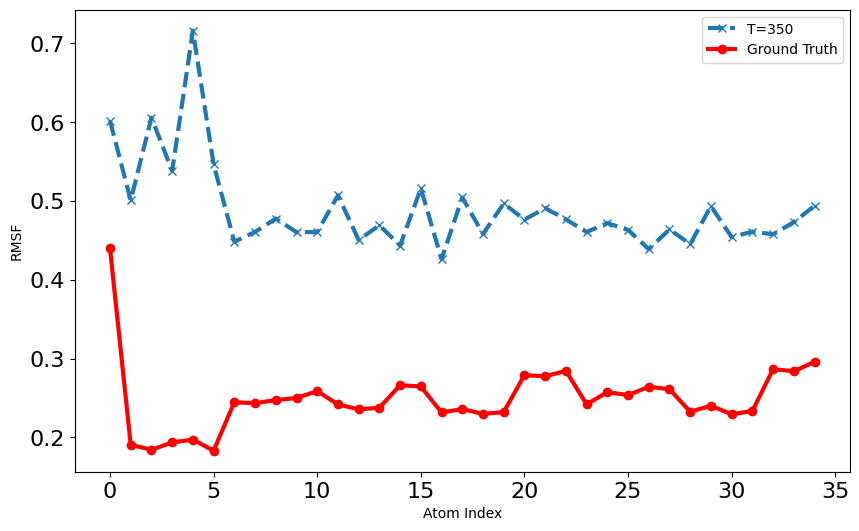}}\hfill
\subfloat[ $T = 100$]{\includegraphics[width=0.3\textwidth]{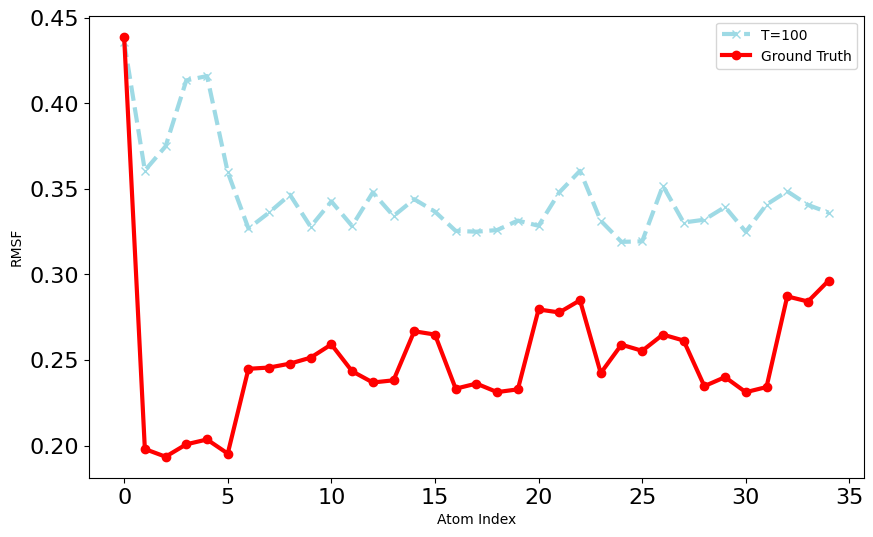}}\hfill
\subfloat[ $T = 75$]{\includegraphics[width=0.3\textwidth]{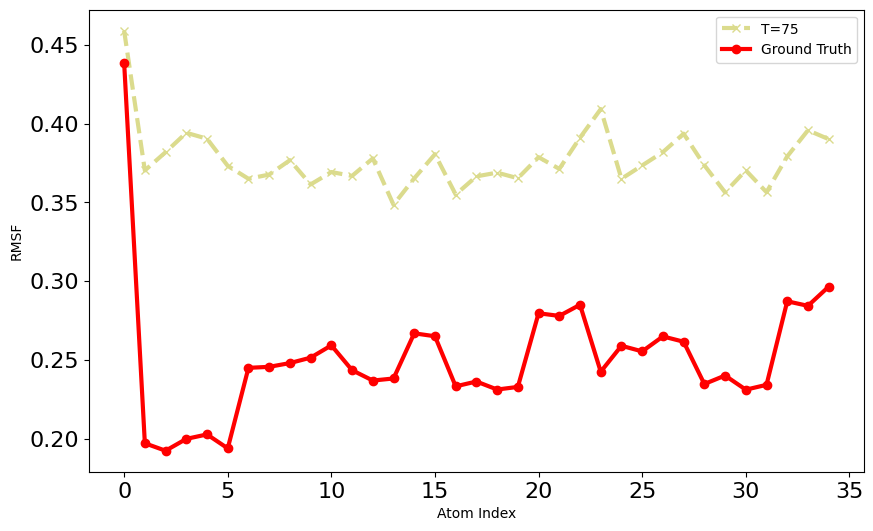}}\hfill
\subfloat[ $T = 50$]{\includegraphics[width=0.3\textwidth]{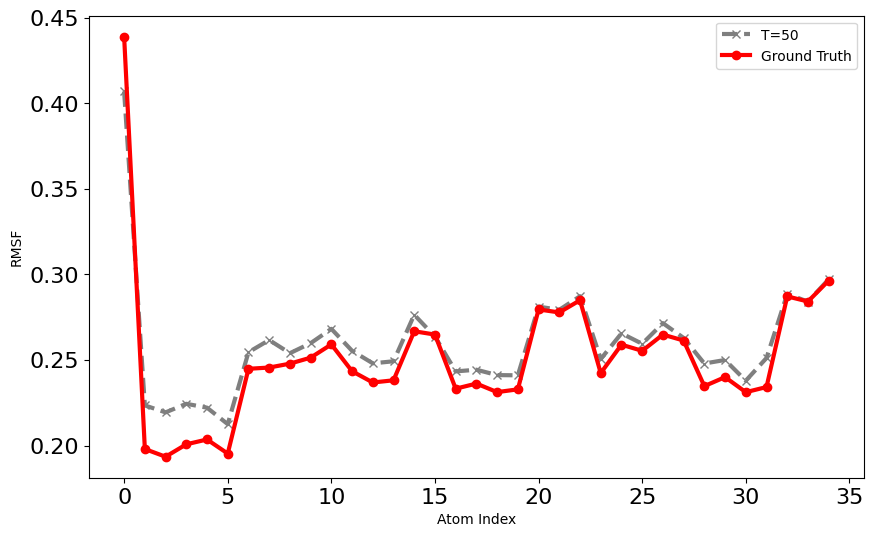}}\hfill
\subfloat[$T = 25$]{\includegraphics[width=0.3\textwidth]{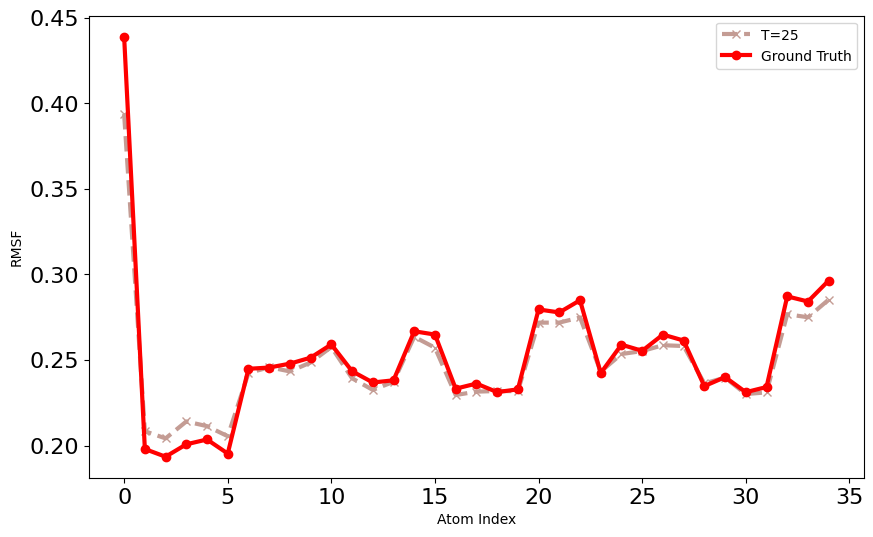}}\hfill
\subfloat[ $T = 20$]{\includegraphics[width=0.3\textwidth]{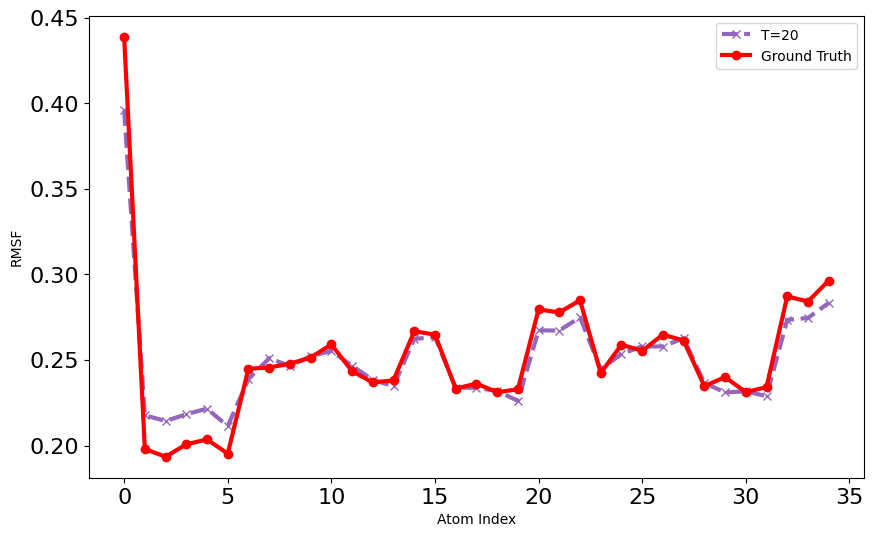}}\hfill
\subfloat[ $T = 10$]{\includegraphics[width=0.3\textwidth]{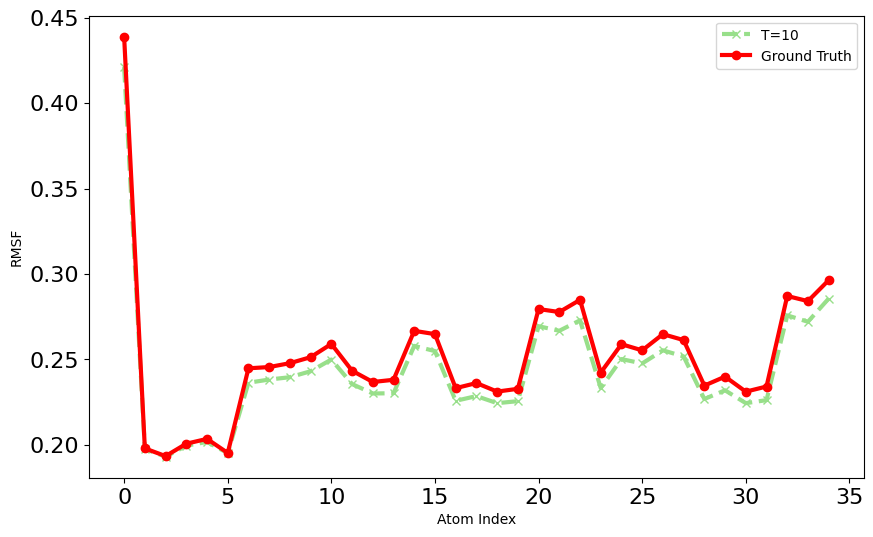}}\hfill
\subfloat[ $T = 5$]{\includegraphics[width=0.3\textwidth]{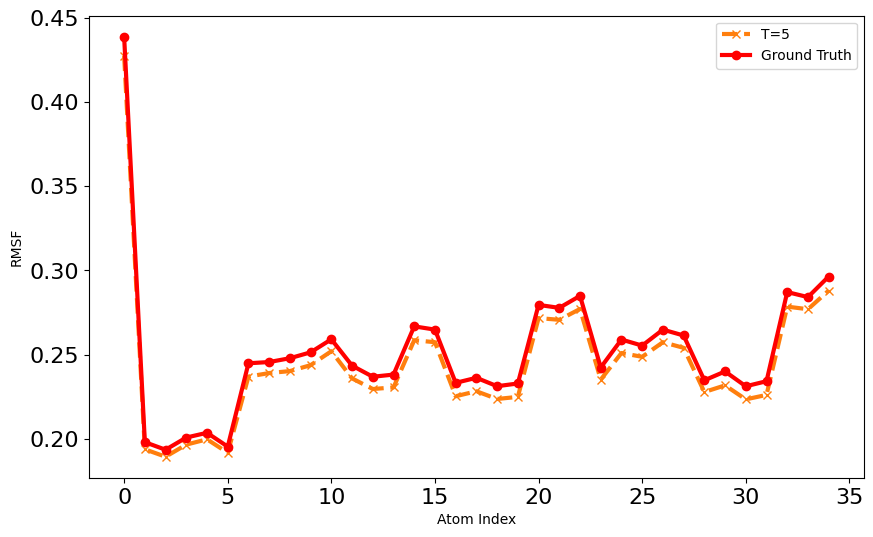}}
\caption{Plots of RMSF with respect to all times $T$ for each atom}\label{fig:R1rmsf}
 \Description{Root Mean Square Fluctuation plots}
\end{figure}

This dataset comprises 35 drug molecule atoms represented in $3$ dimensions trajectories, spanning $10,000$ time steps. We normalized the trajectories to the maximum absolute value of $1$. Our primary objective was to assess the ability of the model to predict atom movements accurately for varying time segments $T$. The mean square and the mean absolute errors are reported in Table~\ref{tab:restest1}. The displacement between the predicted and ground-truth trajectories, as well as the Root Mean Square Fluctuations (RMSF), were computed and visualized in Figures~\ref{fig:loss1},\ref{fig:R1train} and \ref{fig:R1rmsf}. These metrics offer insights into how well the model captures both short- and long-term molecular dynamics. As shown in Table~\ref{tab:restest1}, the model's prediction accuracy improves with shorter $T$ lengths. For smaller $T$, the Mean Square Error (MSE) and Mean Absolute Error (MAE) decrease significantly, indicating that the model performs better when predicting shorter trajectories. The smallest errors were observed for $T = 5$, with an MSE of $0.0019$ and an MAE of $0.0141$.

\begin{figure}[!htb]
\centering
\subfloat[MSE]{\includegraphics[width=0.45\textwidth]{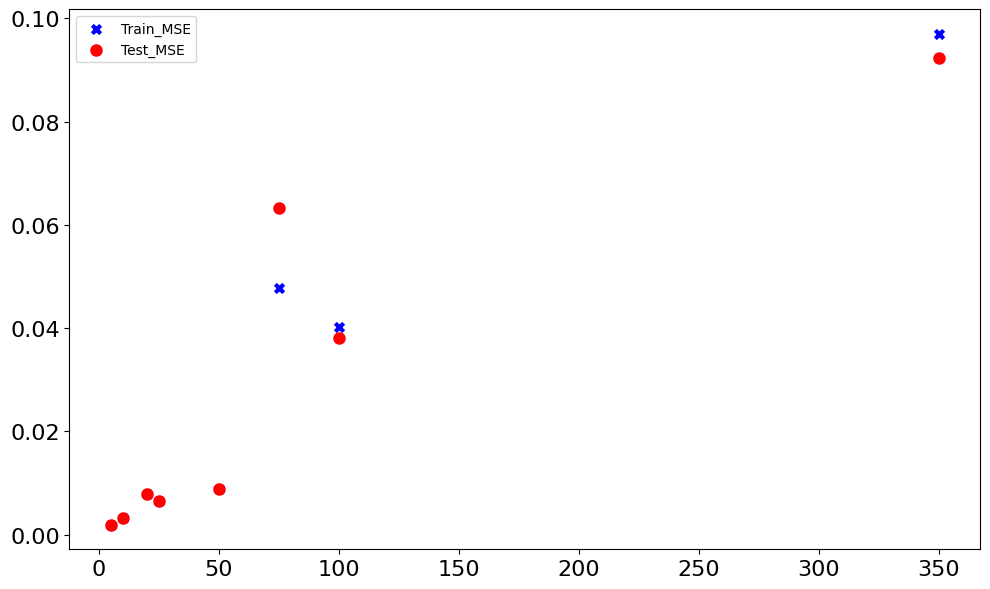}}\hfill
\subfloat[ MAE]{\includegraphics[width=0.45\textwidth]{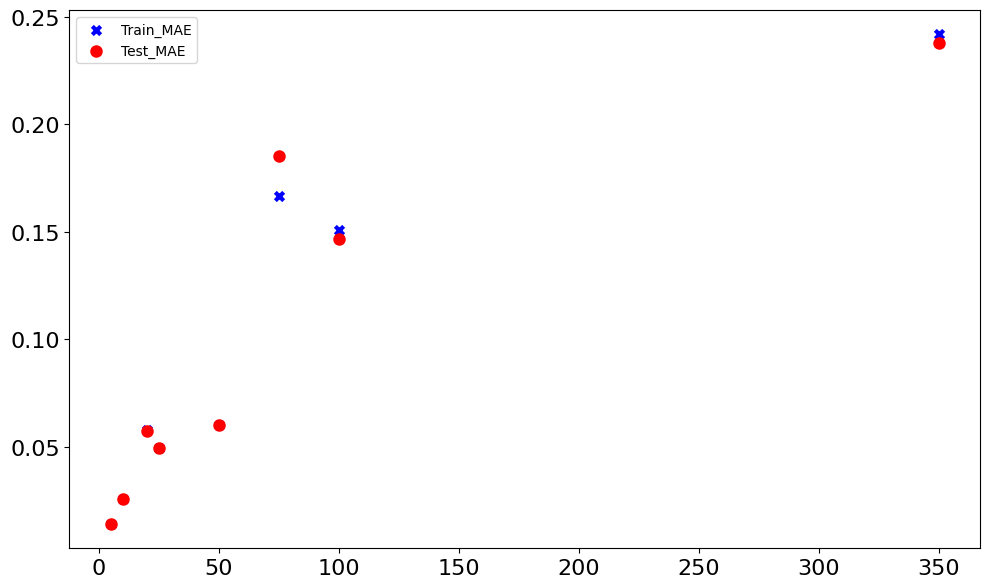}}\hfill
\caption{MAE \& MSE plots after training (blue) and testing (red)}\label{fig:maese}
 \Description{MAE & MSE plots after training (blue) and testing (red)}
\end{figure}

Figure~\ref{fig:loss1} highlights the displacement of trajectories for different $T$. The blue lines represent the ground truth, while the red lines represent the predicted trajectories. For larger $T\geq 50$, discrepancies between the predicted and true trajectories become more apparent; however, for shorter $T<50$, the predicted trajectories closely follow the ground truth.
The Root Mean Square Fluctuation (RMSF) results, depicted in Figures~\ref{fig:R1train} and~\ref{fig:R1rmsf}, provide a more quantitative measure. For larger $T (T\geq 50)$, fluctuations are less accurately captured, with higher RMSF values in predicted trajectories compared to ground truth. In contrast, for smaller $T (T<50)$, the RMSF values of the predicted trajectories align closely with the ground truth, demonstrating that the model is more robust to short-term predictions. Figure~\ref{fig:maese} shows the MSE and MAE values for the training and testing phases. The errors were consistently lower for smaller $T$, further supporting the model's strength in short-duration predictions.

\begin{table*}[!thb] 
\centering
\resizebox{\textwidth}{!}{
\begin{tabular}{|| c | c |c |c| c | c | c |c| c||} 
\hline
 $T$ & $350$& $100$&$75$& $50$&$25$& $20$& $10$& $5$\\ \hline
  Samples & $28$& $100$&$133$&$200$&$400$&$500$&$1000$&$2000$ \\  \hline
    MSE& $0.0923$& $0.0380$& $0.0633$& $0.0089$& $0.0066$& $0.0079$&$0.0033$&$0.0019$\\ \hline
    MAE &$0.2378$& $0.1466$& $ 0.1854$& $0.0603$& $0.0496$& $0.0573$&$0.0254$& $0.0141$\\ \hline
 \end{tabular}}
  \caption{The results on MAE and MSE of trajectories versus different number of samples and simulation duration}\label{tab:restest1}
\end{table*}

\begin{figure}[!htb]
\centering
\subfloat[$x-y$ axes]{\includegraphics[width=0.3\textwidth]{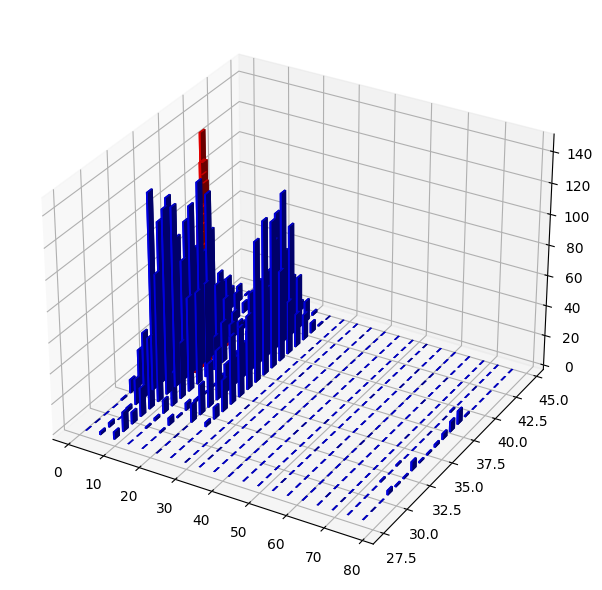}}\hfill
\subfloat[$x-z$ axes]{\includegraphics[width=0.3\textwidth]{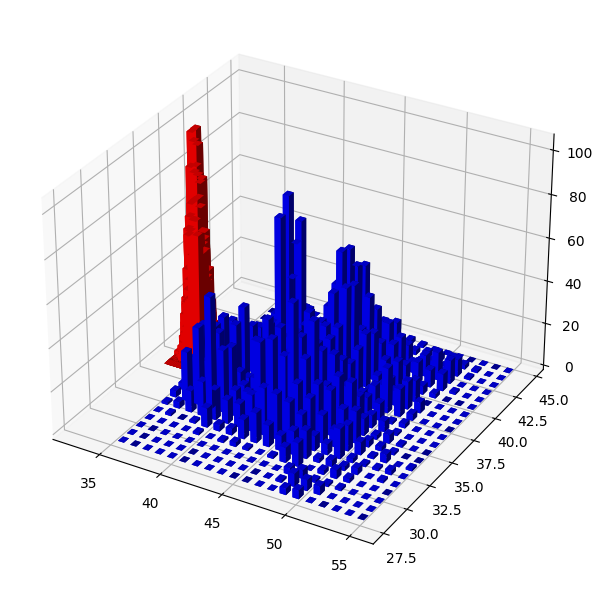}}\hfill
\subfloat[$y-z$ axes]{\includegraphics[width=0.3\textwidth]{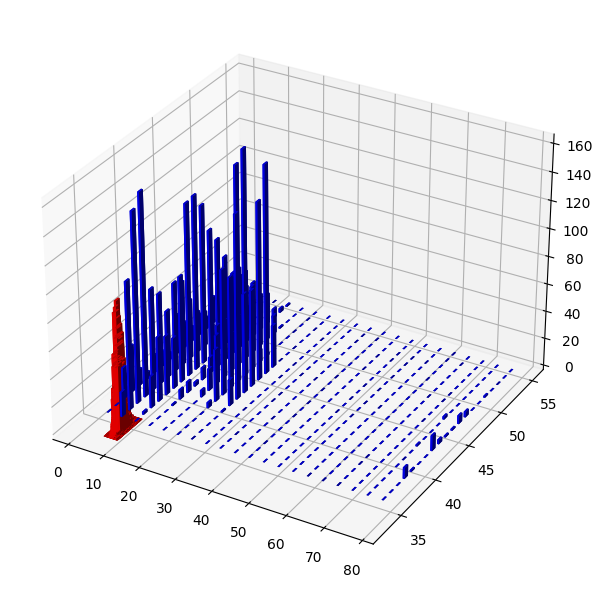}}\hfill
\caption{Probability distribution of donor atom (O) when HB is persistent (red) and when it separates (blue)}
    \Description{Probability distribution of O that formed bond}
\label{fig:r1pdf}
\end{figure}

To effectively analyze the root causes of why a hydrogen bond (HB) separates, we observe the donor atom, $O$, both before and after the separation and plot the resulting distribution as shown in Figure~\ref{fig:r1pdf}.

We use $T = 3$ as it provides the best result and employ the group lasso penalty to maintain the same graph structure through both edge types. By comparing the distribution between the time when the bond persists and when it breaks, we create a ground truth. Table~\ref{tab:RCAD1} reports the accuracy results.

\begin{table}[!htb] 
\centering
\begin{tabular}{|| c| c||} 
 \hline
 Distance & Value   \\ \hline \hline 
Wasserstein distance &$0.83\pm 0.06$\\  \hline
Expectation distance & $0.94\pm 0.06$\\ \hline
KL divergence  &$0.49\pm 0.0.40$\\ \hline
 \end{tabular}\caption{RCA Accuracy}\label{tab:RCAD1}
 \Description{RCA Accuracy}
\end{table}

In Table~\ref{KL-score1}, we list the KL divergence scores for the identified variables driving changes in the trajectories. Figure~\ref{fig:dag1} below shows the PCM matrix obtained from the RCA framework.

\begin{table}[!htb] 
    \centering
    \begin{tabular}{ccccccccccc}
    \hline 
       Atom name  &C2\_10  & H6\_10 & C2\_10 & H3\_10 & H2\_10 & C4\_10 & H10\_11 &  H12\_11 &C14\_11 &C15\_11 \\\hline 
        KL score &  $10.4038$&$ 7.2570$ & $7.2470 $&  $7.2222$&$ 7.0440 $& $6.8911$ & $4.1549$ &$ 3.9925$ & $3.9011$&$ 3.9285$ \\\hline 
    \end{tabular}
    \caption{KL-divergence score for RCA results}
    \Description{KL-divergence score for RCA results}
    \label{KL-score1}
\end{table}
The PCM matrix obtained is shown below in Figure~\ref{fig:dag1}:
\begin{figure}[!ht]
\centering
\includegraphics[width=0.8\linewidth]{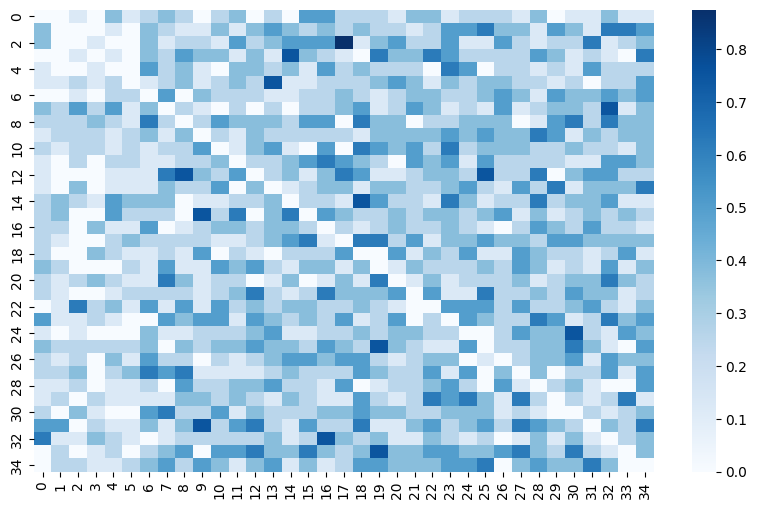}
\caption{PCM matrix obtained}
    \Description{Adjacency matrix}
\label{fig:dag1}
\end{figure}

\begin{figure}[!ht]
\centering
\subfloat[x axis ]{\includegraphics[width=0.3\textwidth]{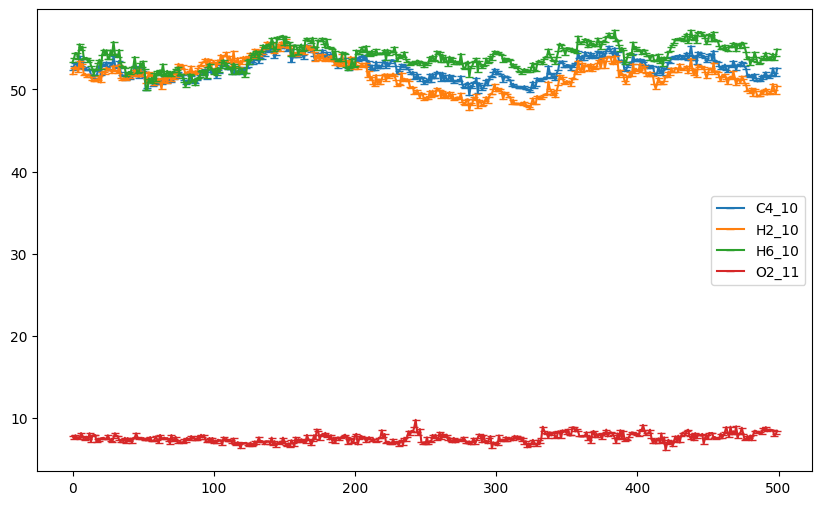}}\hfill
\subfloat[y axis]{\includegraphics[width=0.3\textwidth]{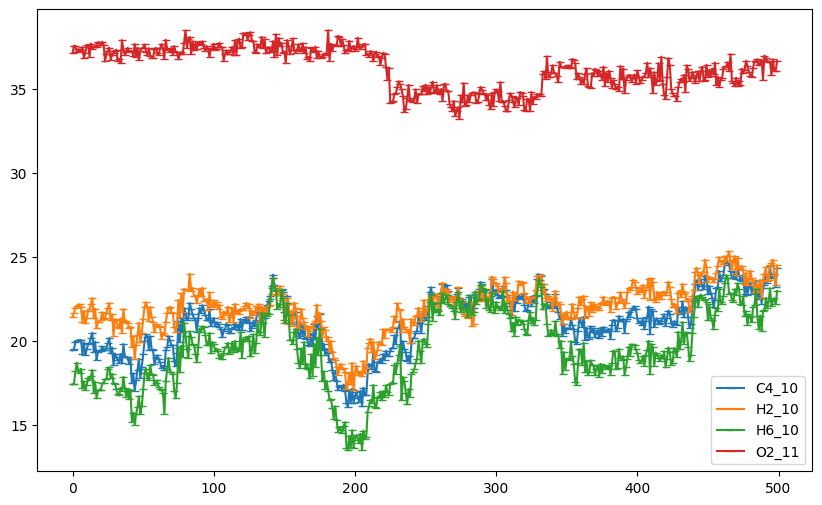}}\hfill
\subfloat[z axis]{\includegraphics[width=0.3\textwidth]{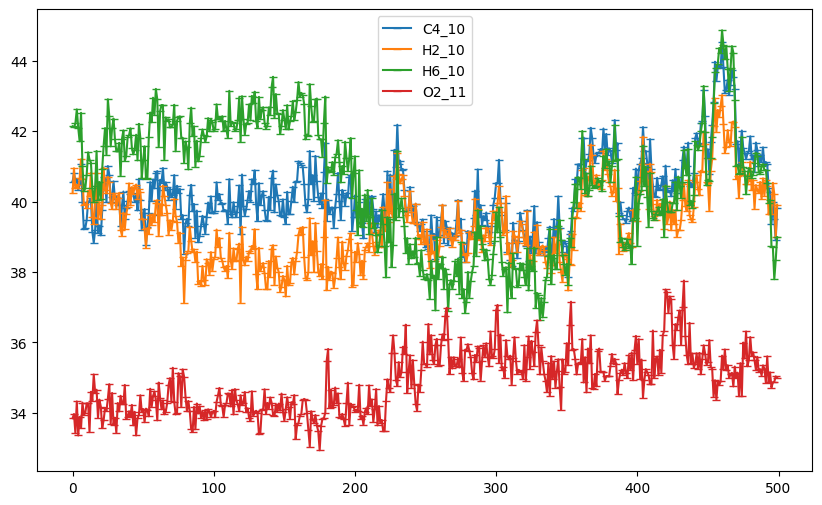}}\hfill
\caption{RCA Trajectory plots}\label{fig:R1rcatraj}
 \Description{RCA trajectory}
\end{figure}

The top $10$ variables that drive changes in the system's trajectory are identified based on their KL divergence scores, as shown in Table~\ref{KL-score1}. Figure~\ref{fig:R1rcatraj} plots the trajectories of four of those atoms right before the HB separation and after it. The $y$ and $z$ axes show that these trajectories are moving towards the donor atom, possibly disrupting the mechanism ongoing at those points. 

\subsection{Phenomenon 2: Tracking two sets of HB atoms}
\begin{figure}[!htb]
\centering
\subfloat[$T = 350$]{\includegraphics[width=0.3\textwidth]{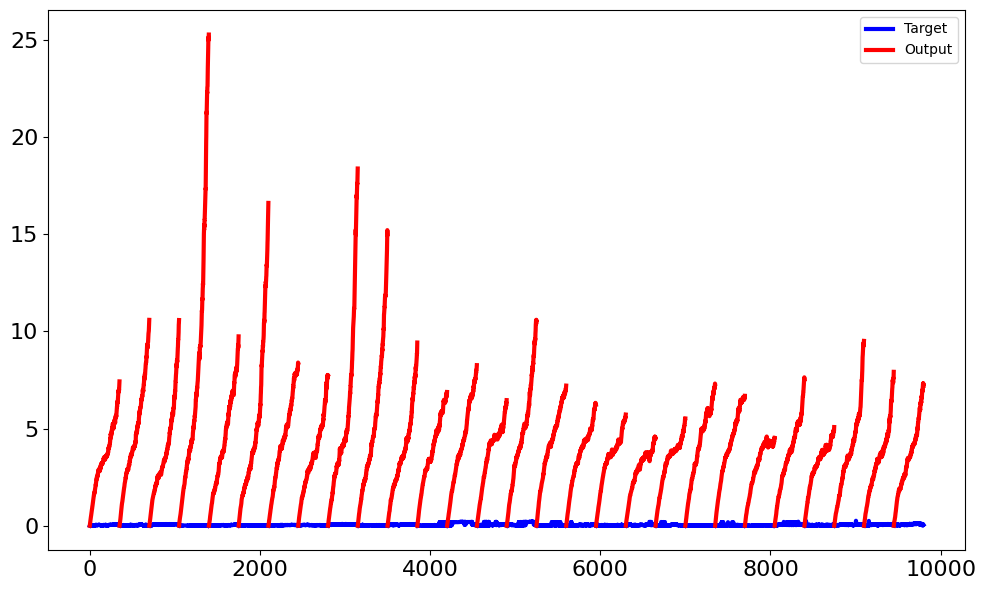}}\hfill
\subfloat[ $T = 100$]{\includegraphics[width=0.3\textwidth]{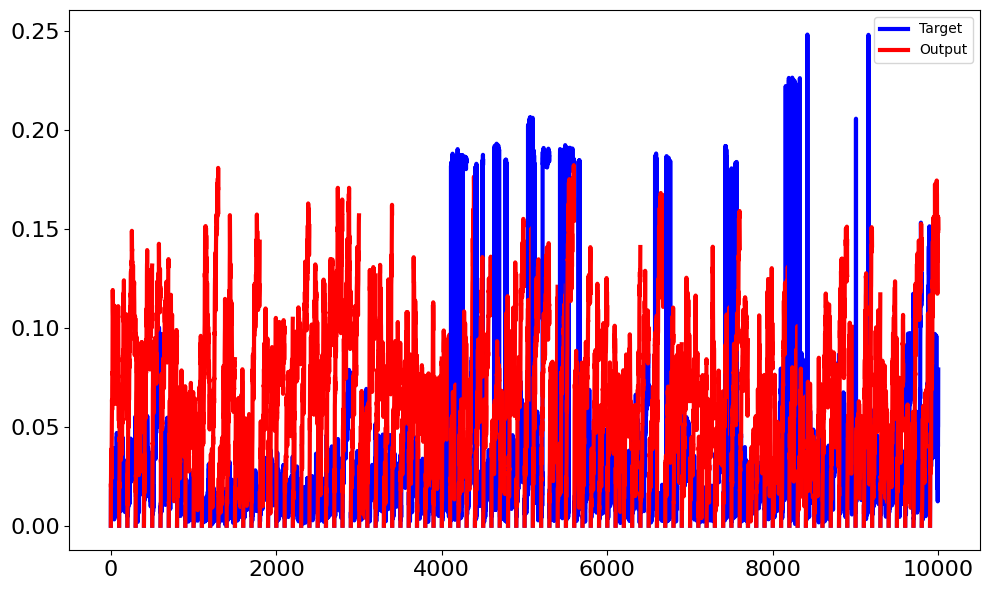}}\hfill
\subfloat[ $T = 70$]{\includegraphics[width=0.3\textwidth]{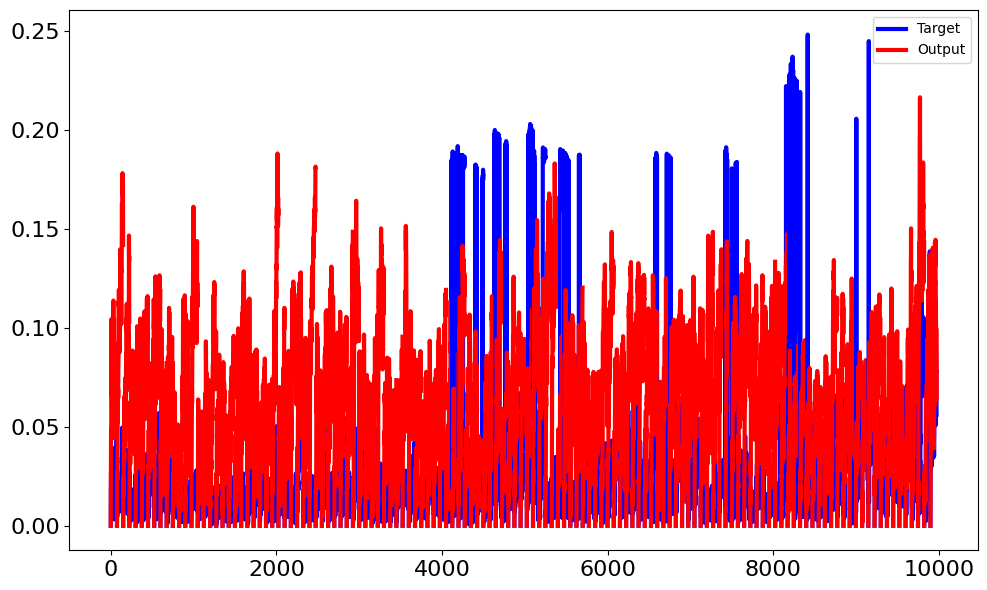}}\hfill
\subfloat[ $T = 50$]{\includegraphics[width=0.3\textwidth]{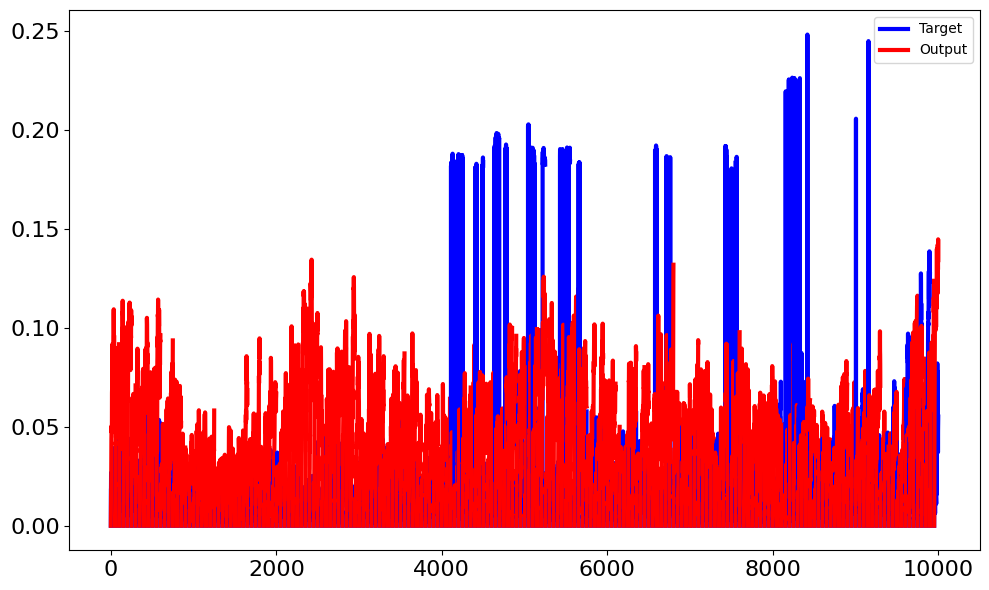}}\hfill
\subfloat[$T = 25$]{\includegraphics[width=0.3\textwidth]{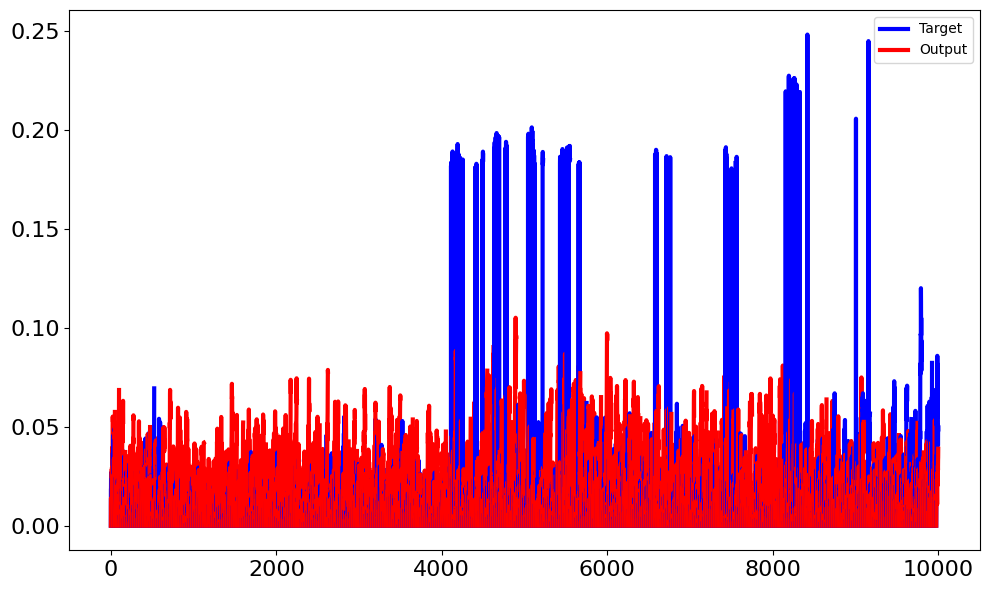}}\hfill
\subfloat[ $T = 20$]{\includegraphics[width=0.3\textwidth]{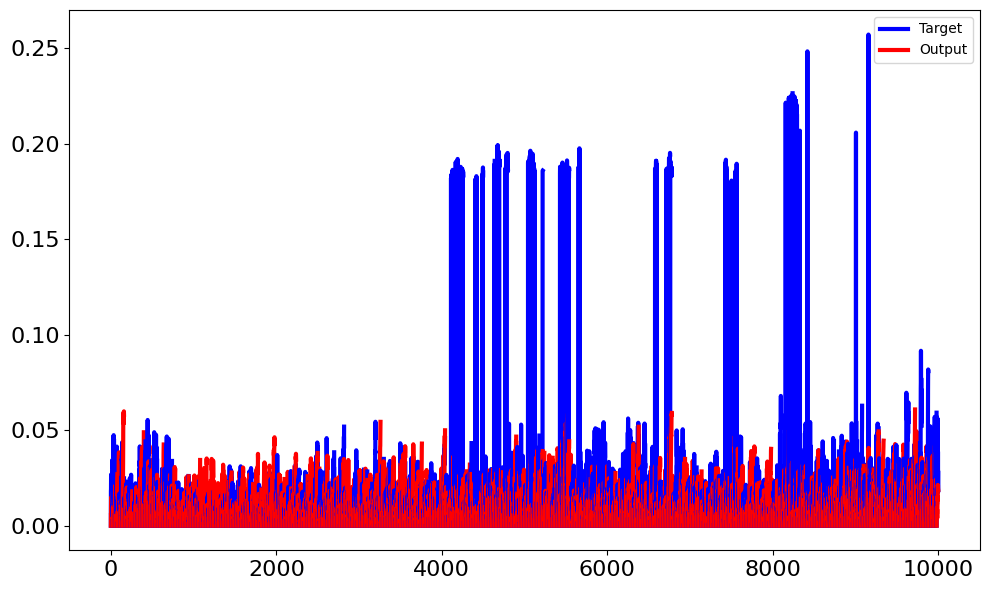}}\hfill
\subfloat[ $T = 10$]{\includegraphics[width=0.3\textwidth]{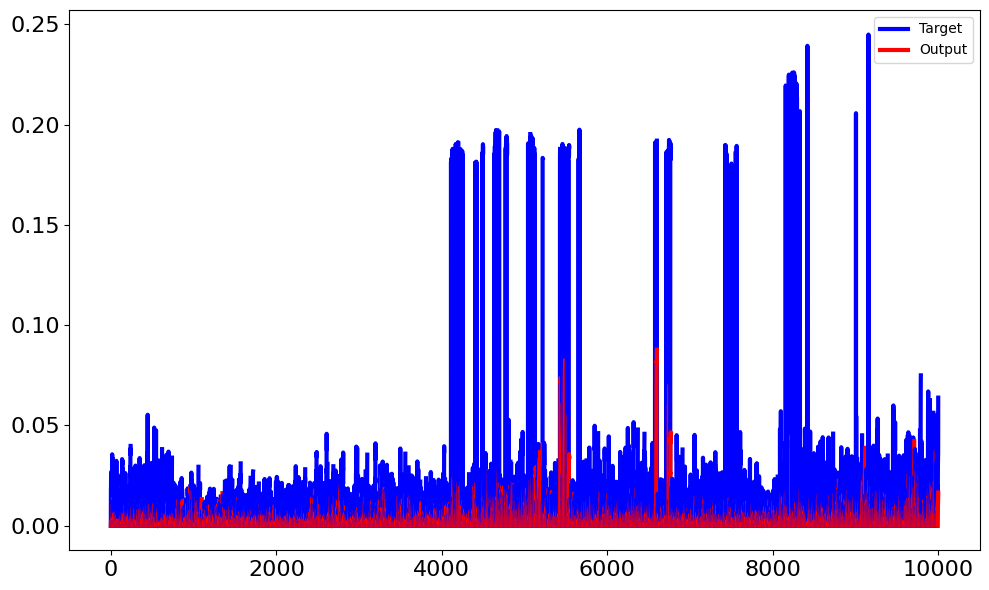}}\hfill
\subfloat[ $T = 5$]{\includegraphics[width=0.3\textwidth]{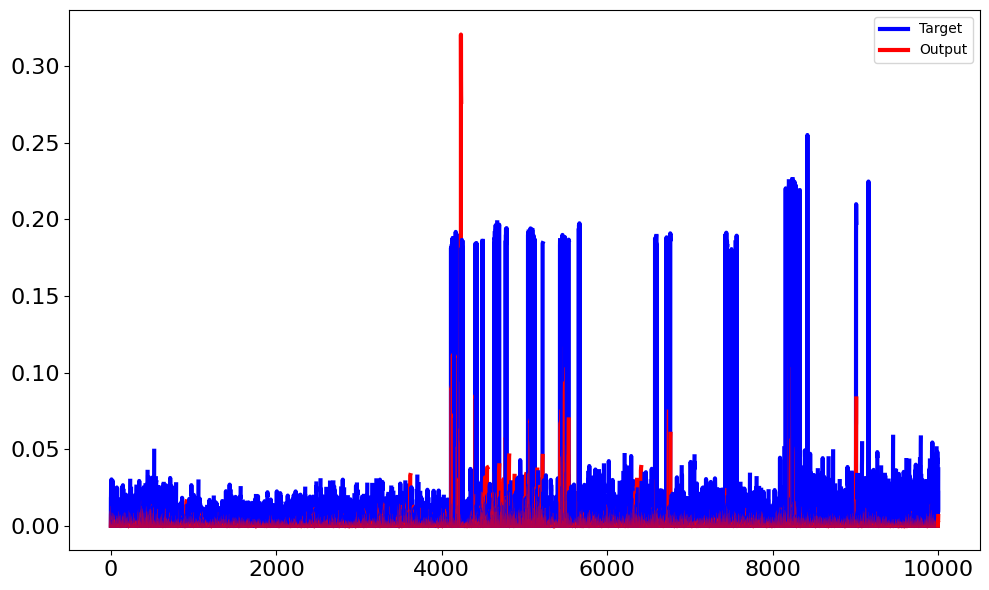}}
\caption{Plots of the displacement between the ground truth(blue) and the predicted(red) values}\label{fig:loss2}
\Description{Prediction plots of $Z$}
\end{figure}

\begin{figure}[!htb]
\centering
\subfloat[$T = 350$]{\includegraphics[width=0.3\textwidth]{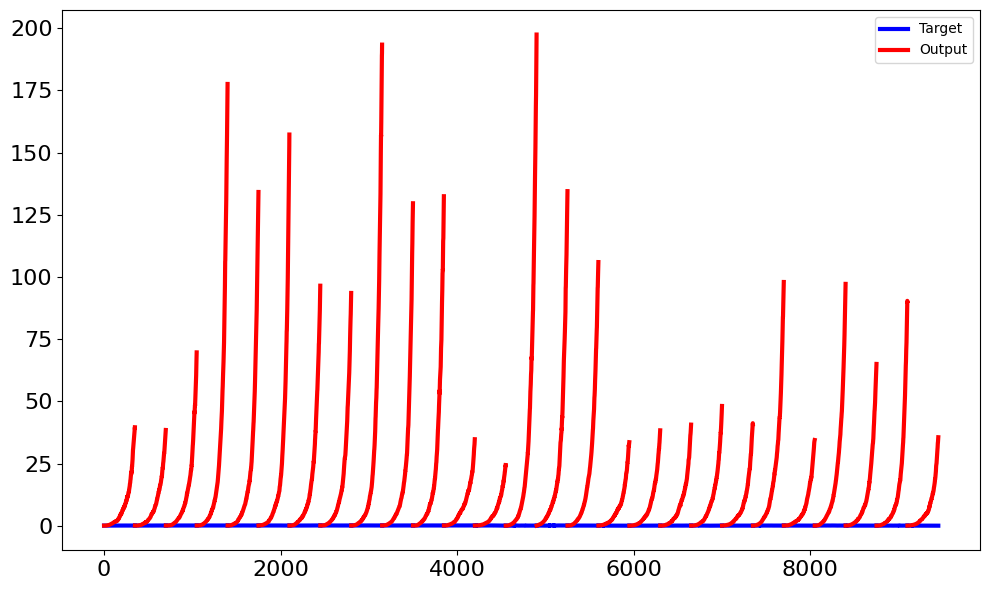}}\hfill
\subfloat[ $T = 100$]{\includegraphics[width=0.3\textwidth]{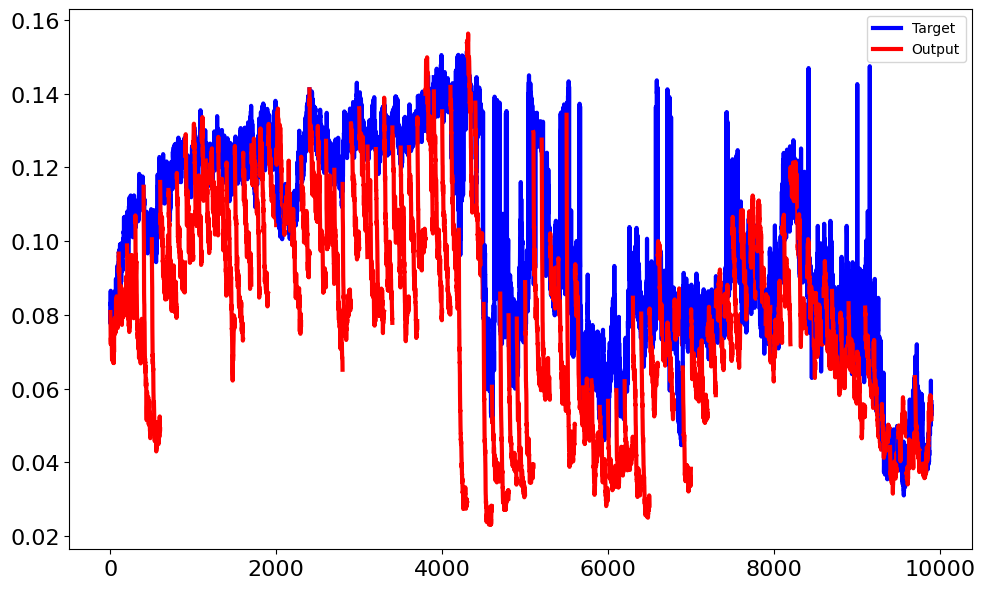}}\hfill
\subfloat[ $T = 75$]{\includegraphics[width=0.3\textwidth]{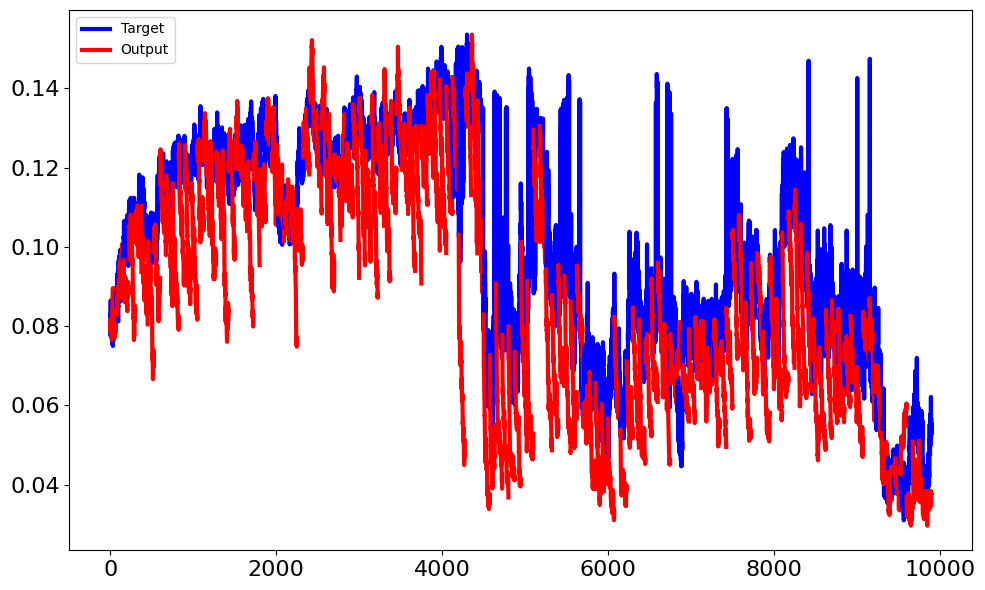}}\hfill
\subfloat[ $T = 50$]{\includegraphics[width=0.3\textwidth]{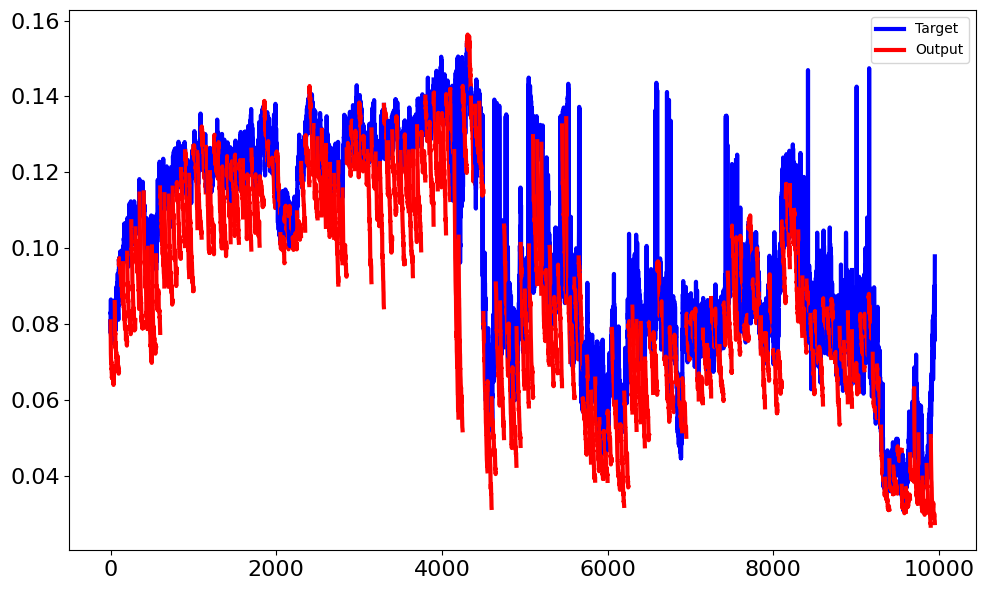}}\hfill
\subfloat[$T = 25$]{\includegraphics[width=0.3\textwidth]{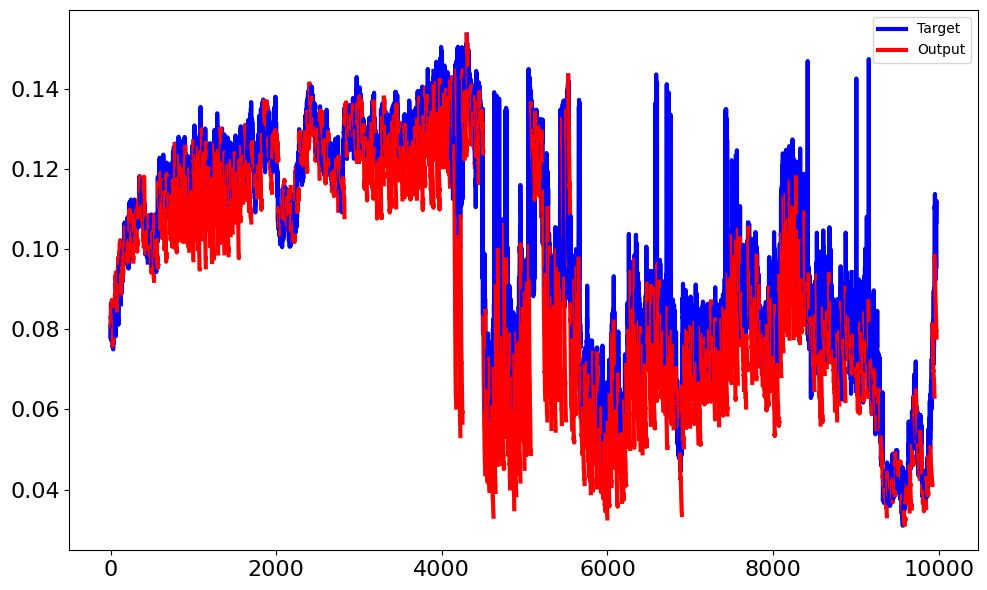}}\hfill
\subfloat[ $T = 20$]{\includegraphics[width=0.3\textwidth]{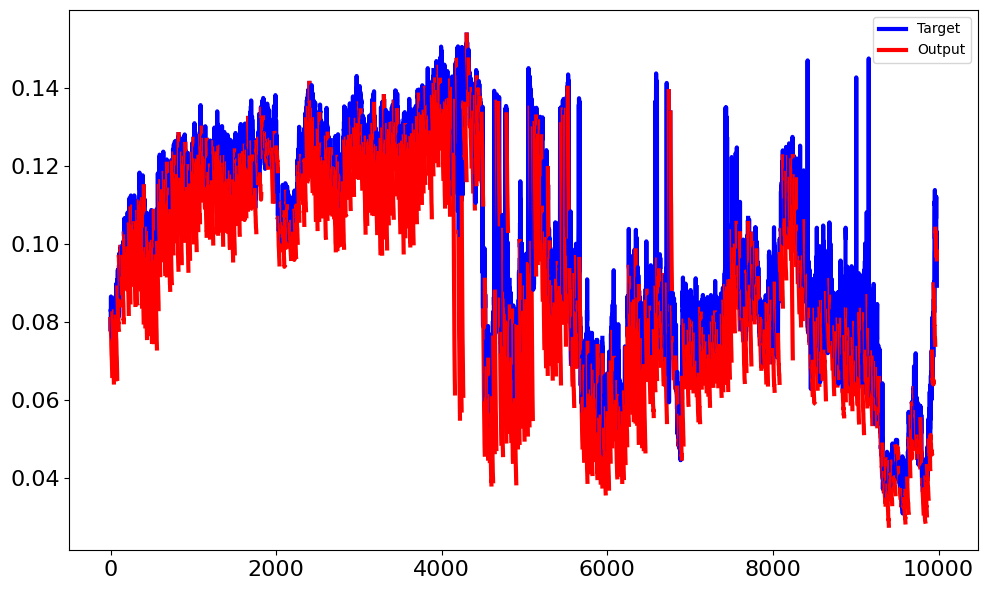}}\hfill
\subfloat[ $T = 10$]{\includegraphics[width=0.3\textwidth]{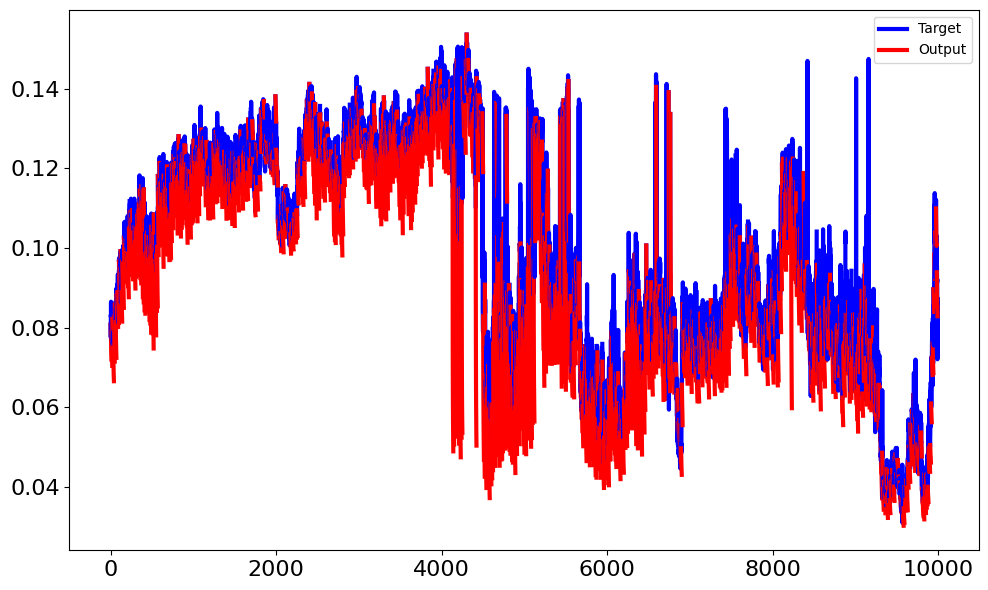}}\hfill
\subfloat[ $T = 5$]{\includegraphics[width=0.3\textwidth]{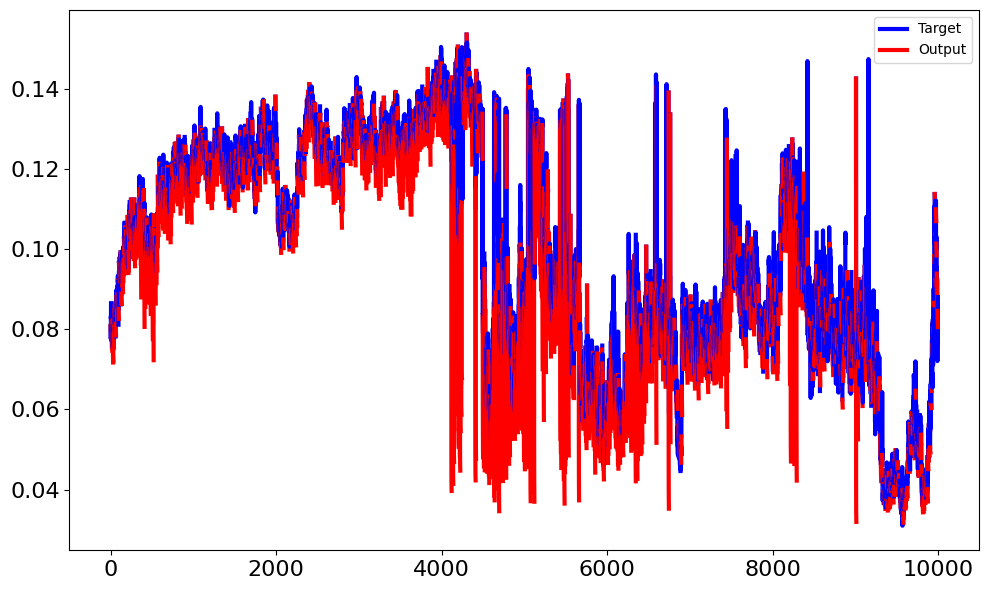}}
\caption{Plots of RMSF with respect to all atoms at each time $t$}\label{fig:R2train}
\Description{Value square deviation plots}
\end{figure}

\begin{figure}[!htb]
\centering
\subfloat[$T = 350$]{\includegraphics[width=0.3\textwidth]{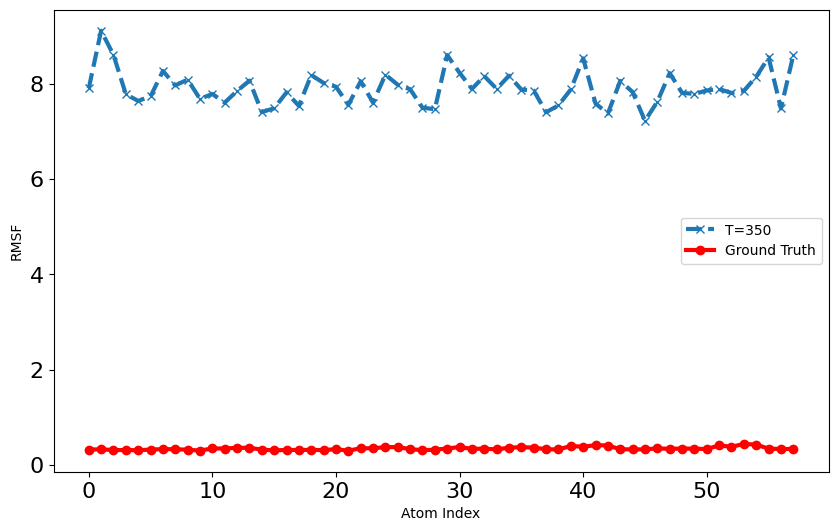}}\hfill
\subfloat[ $T = 100$]{\includegraphics[width=0.3\textwidth]{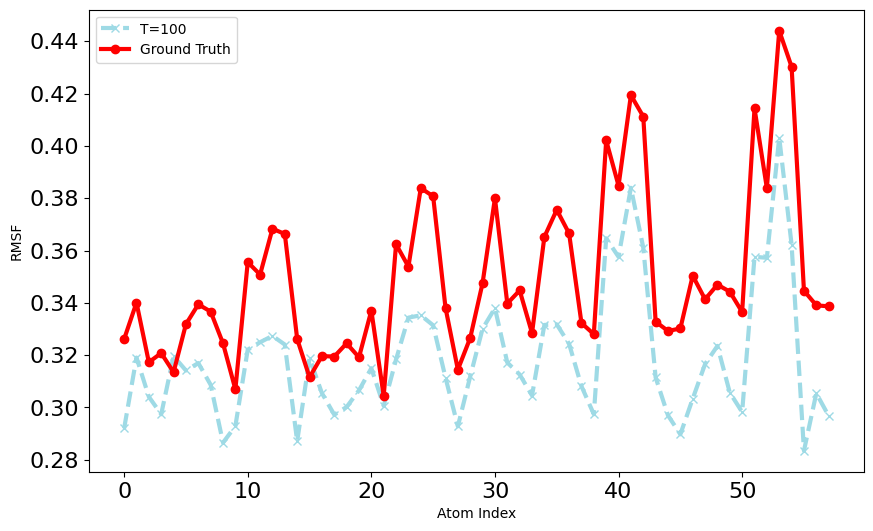}}\hfill
\subfloat[ $T = 75$]{\includegraphics[width=0.3\textwidth]{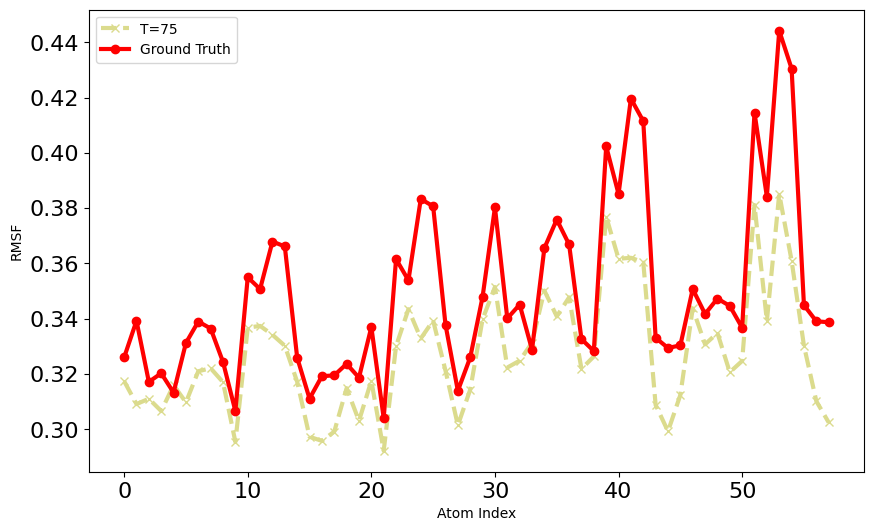}}\hfill
\subfloat[ $T = 50$]{\includegraphics[width=0.3\textwidth]{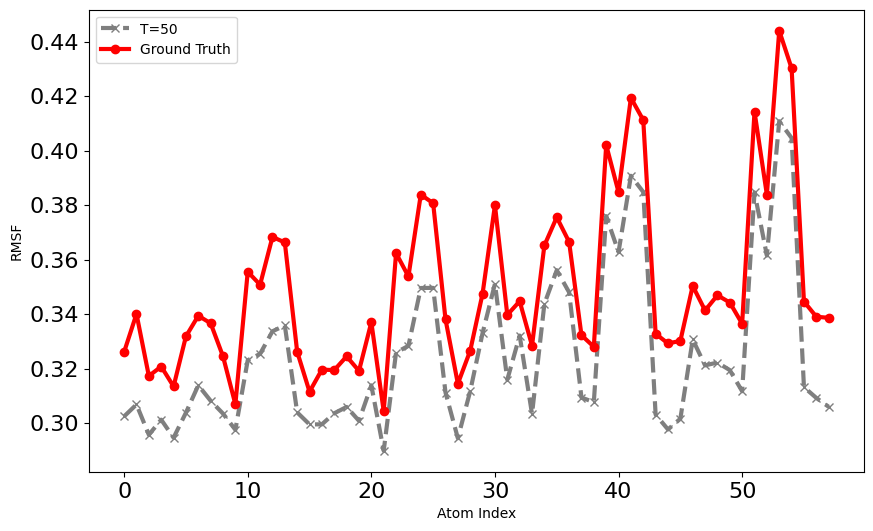}}\hfill
\subfloat[$T = 25$]{\includegraphics[width=0.3\textwidth]{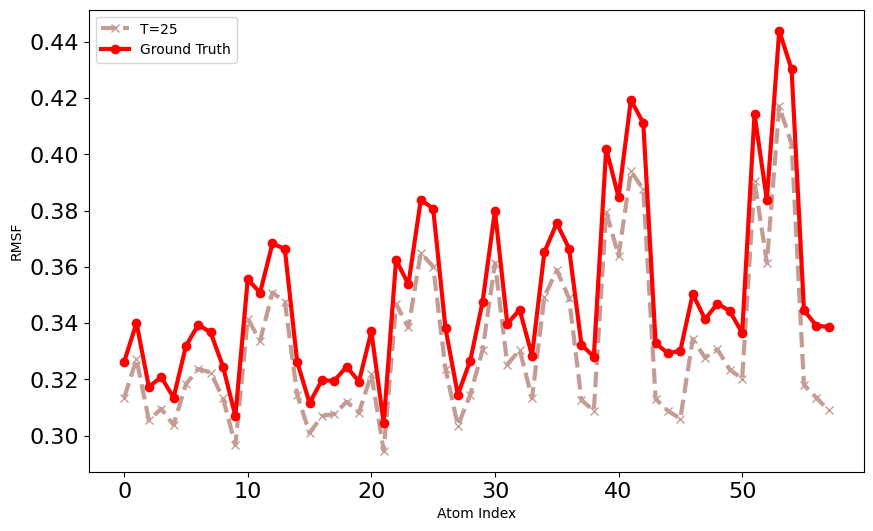}}\hfill
\subfloat[ $T = 20$]{\includegraphics[width=0.3\textwidth]{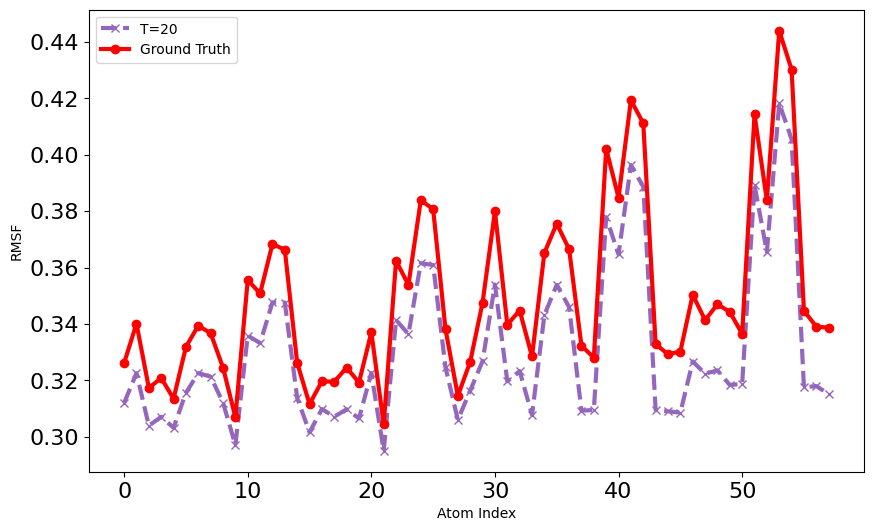}}\hfill
\subfloat[ $T = 10$]{\includegraphics[width=0.3\textwidth]{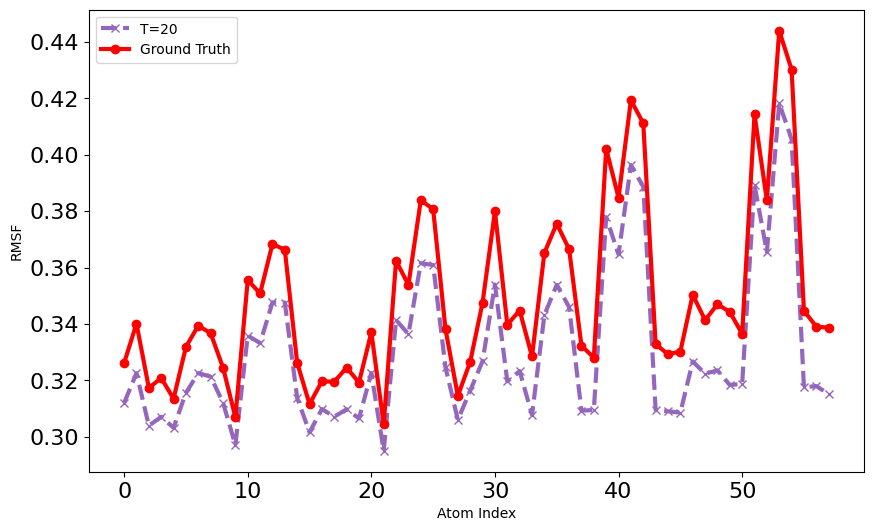}}\hfill
\subfloat[ $T = 5$]{\includegraphics[width=0.3\textwidth]{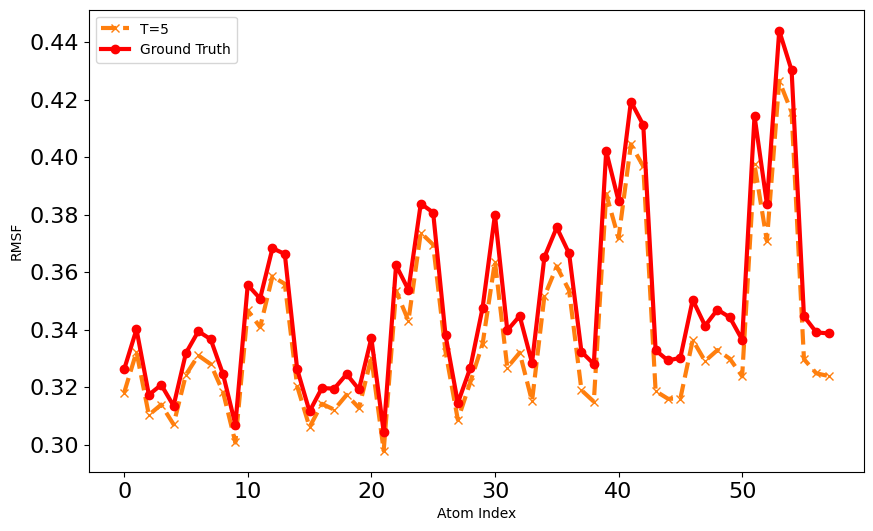}}
\caption{Plots of RMSF with respect to all times $T$ for each atom}\label{fig:R2rmsf}
 \Description{Root Mean Square Fluctuation plots}
\end{figure}

\begin{figure}[!htb]
\centering
\subfloat[MSE]{\includegraphics[width=0.45\textwidth]{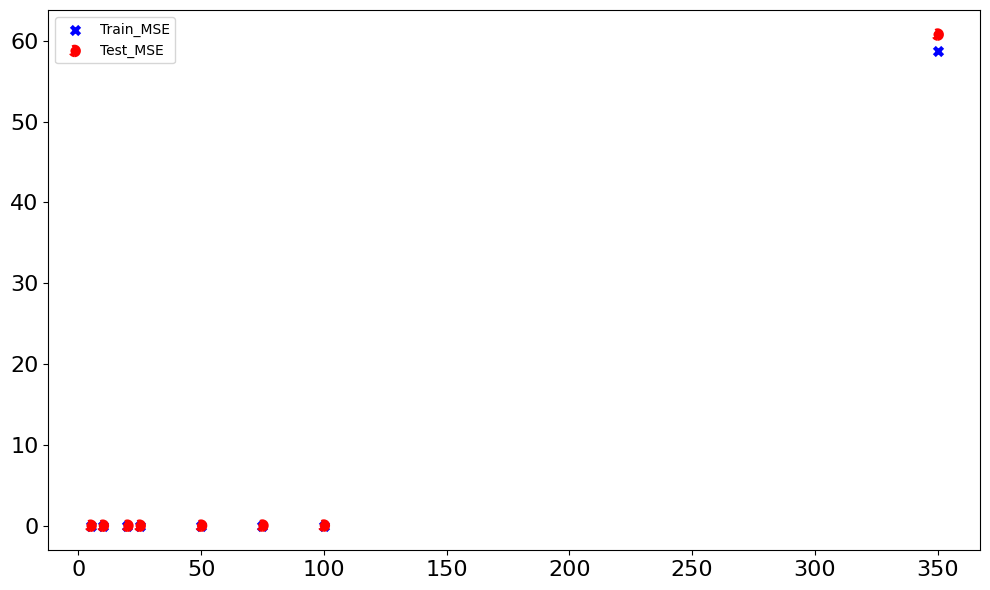}}\hfill
\subfloat[ MAE]{\includegraphics[width=0.45\textwidth]{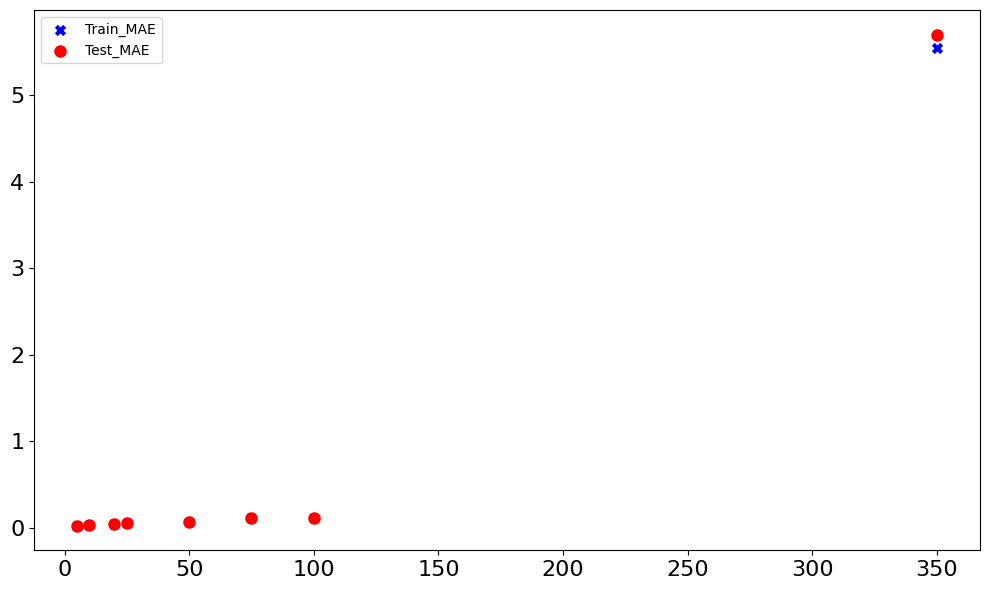}}\hfill
\caption{MAE \& MSE plots after training (blue) and testing (red)}\label{fig:maese2}
 \Description{MAE \& MSE plots after training (blue) and testing (red)}
\end{figure}

This dataset comprises 35 drug molecule atoms represented in $3$ dimensions trajectories, spanning $10,000$ time steps. We normalized the trajectories to the maximum absolute value of $1$. Our objectives remain the same as the previous results. The mean square and the mean absolute errors are reported in Table~\ref{tab:restest2}. The displacement between the predicted and ground-truth trajectories, as well as the Root Mean Square Fluctuations (RMSF), were computed and visualized in Figures~\ref{fig:loss2},\ref{fig:R2train} and \ref{fig:R2rmsf}. As shown in Table~\ref{tab:restest2}, the model's prediction accuracy improves with shorter $T$ lengths. For smaller $T$, the Mean Square Error (MSE) and Mean Absolute Error (MAE) decrease significantly, indicating that the model performs better when predicting shorter trajectories. The smallest errors were observed for $T = 5$, with an MSE of $0.0031$ and an MAE of $0.0207$. Figure~\ref{fig:loss2} highlights the displacement of trajectories for different $T$. The blue lines represent the ground truth, while the red lines represent the predicted trajectories. For larger $T\geq 25$, discrepancies between the predicted and true trajectories become more apparent; however, for shorter $T<25$, the predicted trajectories closely follow the ground truth. The Root Mean Square Fluctuation (RMSF) results, depicted in Figures~\ref{fig:R2train} and~\ref{fig:R2rmsf}, provide a more quantitative measure. For larger $T (T\geq 25)$, fluctuations are less accurately captured, with higher RMSF values in predicted trajectories compared to ground truth. In contrast, for smaller $T (T<25)$, the RMSF values of the predicted trajectories align closely with the ground truth, demonstrating that the model is more robust to short-term predictions. Figure~\ref{fig:maese2} shows the MSE and MAE values for the training and testing phases. The errors were consistently lower for smaller $T$, further supporting the model's strength in short-duration predictions.

\begin{table}[thb!] 
\centering
 \caption{The results on MAE and MSE of trajectories versus different number of samples and simulation duration (Test)}
 \label{tab:restest2}
\resizebox{\textwidth}{!}{
\begin{tabular}{|| c | c |c |c| c | c | c |c| c||} 
\hline
 $T$ & $350$& $100$&$75$& $50$&$25$& $20$& $10$& $5$\\ \hline
  Samples & $28$& $100$&$133$&$200$&$400$&$500$&$1000$&$2000$ \\  \hline
    MSE& $0.6076$ &$0.0224$&$0.0216$ &$0.0112$ &$0.0075$& $0.0061$ &$0.0043$ &$0.0031$ \\ \hline
    MAE&$5.7006$ &$0.1086$& $0.1065$& $0.06914$& $0.0492$& $0.0373$&$0.0267$& $0.0207$\\ \hline
 \end{tabular}}
\end{table}

\begin{figure}[!htb]
\centering
\subfloat[$x-y$ axes]{\includegraphics[width=0.3\textwidth]{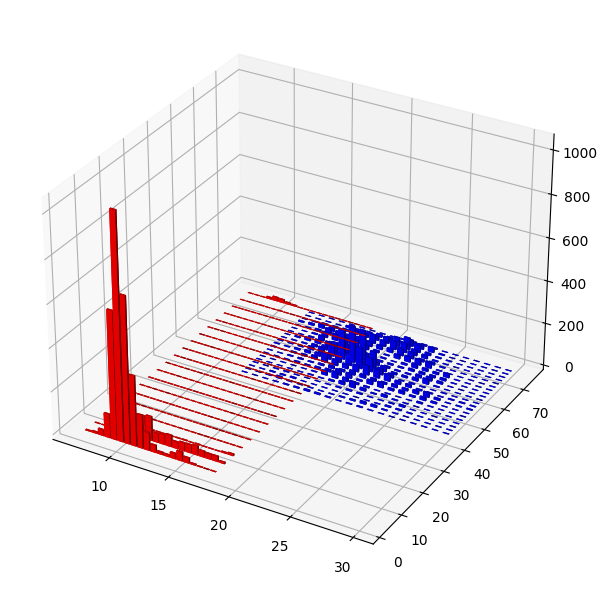}}\hfill
\subfloat[$x-z$ axes]{\includegraphics[width=0.3\textwidth]{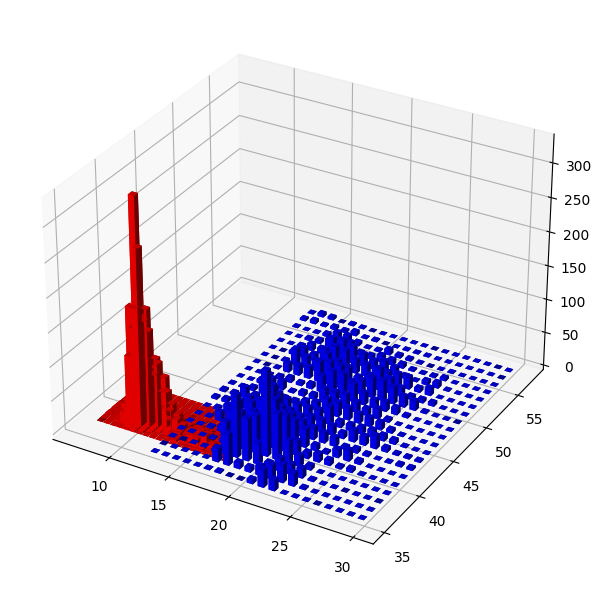}}\hfill
\subfloat[$y-z$ axes]{\includegraphics[width=0.3\textwidth]{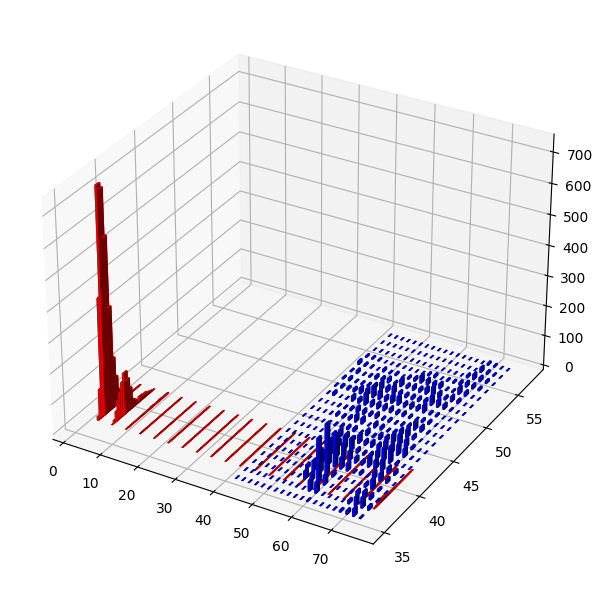}}\hfill
\caption{Probability distribution of donor atom (O) when HB is persistent (red) and when it separates (blue)}
    \Description{Probability distribution of O that formed bond}
\label{fig:r2pdf}
\end{figure}

To effectively analyze the root causes of why a hydrogen bond (HB) separates, we observe one of the two donor atoms, $O$, both before and after the separation, and plot the resulting distribution as shown in Figure~\ref{fig:r2pdf}.

$T$ remains $3$ as it provides the best result and employs the group lasso penalty to maintain the same graph structure through both edge types. By comparing the distribution between the time when the bond persists and when it breaks, we create a ground truth. Table~\ref{tab:RCAD2} reports the accuracy results.

\begin{table}[!htb] 
\centering
\begin{tabular}{|| c| c||} 
 \hline
 Distance & Value   \\ \hline \hline 
Wasserstein distance &$0.90\pm 0.07$\\  \hline
Expectation distance & $0.97\pm 0.03$\\ \hline
KL divergence  &$0.86\pm 0.04$\\ \hline
 \end{tabular}
 \caption{RCA Accuracy}\label{tab:RCAD2}
\end{table}

In Table~\ref{KL-score2}, we list the KL divergence scores for the identified variables driving changes in the trajectories. Figure~\ref{fig:dag2} shows the PCM matrix obtained from the RCA framework.

\begin{table}[!htb] 
    \centering
    \begin{tabular}{ccccccccccc}
    \hline 
       Atom name  &H9\_8&C10\_8&H7\_8 & C6\_8 & H8\_8& C12\_8& C5\_8& C11\_8& C9\_8 & H6\_8 \\\hline 
        KL score  & $129.1039$ & $118.2241$&$117.0768$& $116.9391$& $110.3178$& 
$96.7274$& $ 94.7700$ & $90.6072$& $87.8148$&$76.8555$\\\hline 
    \end{tabular}
    \caption{KL-divergence score for RCA results}
    \Description{KL-divergence score for RCA results}
    \label{KL-score2}
\end{table}

The PCM matrix obtained is shown below in Figure~\ref{fig:dag2}:
\begin{figure}[!ht]
\centering
\includegraphics[width=0.8\linewidth]{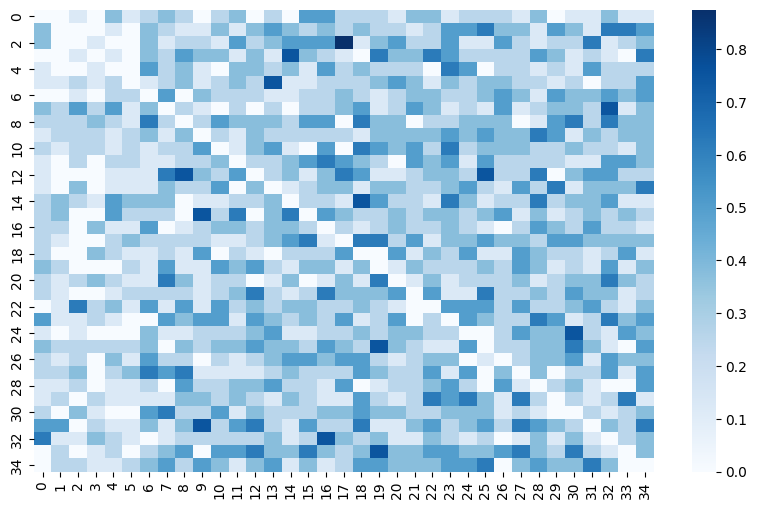}
\caption{PCM matrix obtained}
    \Description{Adjacency matrix}
\label{fig:dag2}
\end{figure}

\begin{figure}[!htb]
\centering
\subfloat[X axis ]{\includegraphics[width=0.3\textwidth]{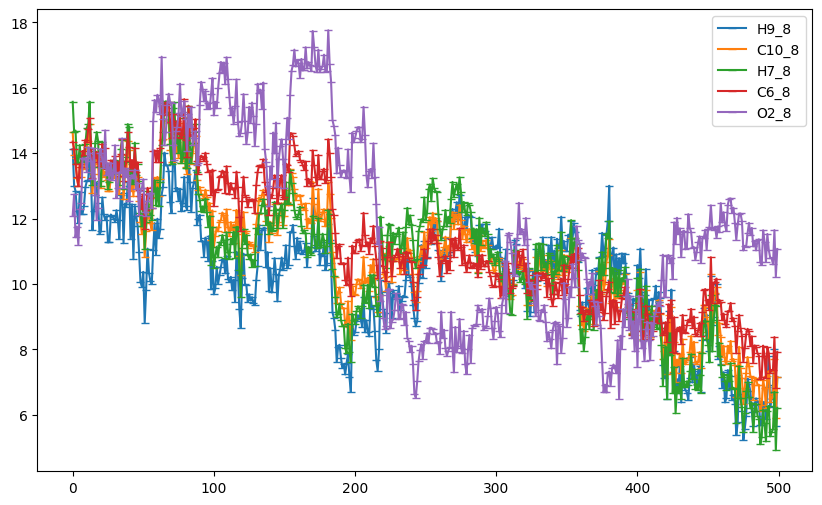}}\hfill
\subfloat[ Y axis]{\includegraphics[width=0.3\textwidth]{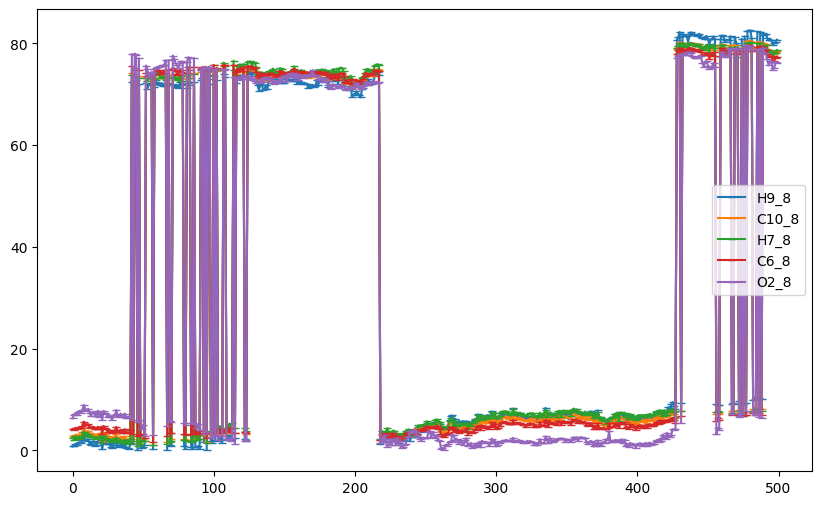}}\hfill
\subfloat[ Z axis]{\includegraphics[width=0.3\textwidth]{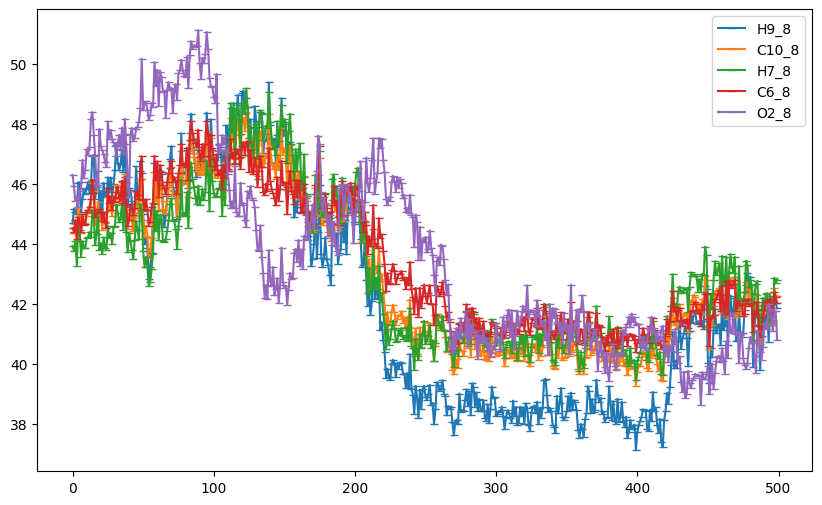}}\hfill
\caption{RCA Trajectory plots}\label{fig:R2rcatraj}
 \Description{RCA trajectory}
\end{figure}

\section{Conclusions}
In this research, we identify the root cause variables of hydrogen bond formation and separation events. Specifically, we treat the separation of hydrogen bonds as an "intervention" occurring and represent the causal structure of the bonding and separation events in the MDS as graphical causal models. These causal models are built using a variational autoencoder-inspired architecture that enables us to infer causal relationships across samples with diverse underlying causal graphs while leveraging shared dynamic information. We further include a step to infer the root causes of changes in the joint distribution of the causal models by constructing causal models that capture shifts in the conditional distributions of molecular interactions during bond formation or separation. Our results show that we are successfully able to capture these events in MDS, and to the best of our knowledge is the first time Root causes has been introduced to the field of chemistry.








  \bibliographystyle{ACM-Reference-Format}
  \bibliography{ref_rca}


\begin{thebibliography}{55}


\ifx \showCODEN    \undefined \def \showCODEN     #1{\unskip}     \fi
\ifx \showDOI      \undefined \def \showDOI       #1{#1}\fi
\ifx \showISBNx    \undefined \def \showISBNx     #1{\unskip}     \fi
\ifx \showISBNxiii \undefined \def \showISBNxiii  #1{\unskip}     \fi
\ifx \showISSN     \undefined \def \showISSN      #1{\unskip}     \fi
\ifx \showLCCN     \undefined \def \showLCCN      #1{\unskip}     \fi
\ifx \shownote     \undefined \def \shownote      #1{#1}          \fi
\ifx \showarticletitle \undefined \def \showarticletitle #1{#1}   \fi
\ifx \showURL      \undefined \def \showURL       {\relax}        \fi
\providecommand\bibfield[2]{#2}
\providecommand\bibinfo[2]{#2}
\providecommand\natexlab[1]{#1}
\providecommand\showeprint[2][]{arXiv:#2}

\bibitem[Adesunkanmi and Kumar(2024)]%
        {adesunkanmi2024expectation}
\bibfield{author}{\bibinfo{person}{Rahmat Adesunkanmi} {and} \bibinfo{person}{Ratnesh Kumar}.} \bibinfo{year}{2024}\natexlab{}.
\newblock \showarticletitle{Expectation Distance-Based Distributional Clustering for Noise-Robustness}.
\newblock \bibinfo{journal}{\emph{IEEE Transactions on Knowledge and Data Engineering}} (\bibinfo{year}{2024}).
\newblock


\bibitem[Adesunkanmi et~al\mbox{.}(2024)]%
        {adesunkanmi2024neuro}
\bibfield{author}{\bibinfo{person}{Rahmat Adesunkanmi}, \bibinfo{person}{Balaji Sesha~Srikanth Pokuri}, {and} \bibinfo{person}{Ratnesh Kumar}.} \bibinfo{year}{2024}\natexlab{}.
\newblock \bibinfo{title}{NeuroKoopman Dynamic Causal Discovery}.
\newblock
\newblock
\showeprint[arxiv]{2404.16326}~[cs.LG]
\urldef\tempurl%
\url{https://arxiv.org/abs/2404.16326}
\showURL{%
\tempurl}


\bibitem[Anowar et~al\mbox{.}(2022)]%
        {ADBIS2022}
\bibfield{author}{\bibinfo{person}{Md~Hasan Anowar}, \bibinfo{person}{Abdullah Shamail}, \bibinfo{person}{Xiaoyu Wang}, \bibinfo{person}{Goce Trajcevski}, \bibinfo{person}{Sohail Murad}, \bibinfo{person}{Cynthia~J. Jameson}, {and} \bibinfo{person}{Ashfaq Khokhar}.} \bibinfo{year}{2022}\natexlab{}.
\newblock \showarticletitle{Generalization Aware Compression of Molecular Trajectories}. In \bibinfo{booktitle}{\emph{Advances in Databases and Information Systems}}, \bibfield{editor}{\bibinfo{person}{Silvia Chiusano}, \bibinfo{person}{Tania Cerquitelli}, {and} \bibinfo{person}{Robert Wrembel}} (Eds.). \bibinfo{publisher}{Springer International Publishing}, \bibinfo{address}{Cham}, \bibinfo{pages}{270--284}.
\newblock


\bibitem[Anowar et~al\mbox{.}(2024)]%
        {Anowar_InfoSys2024}
\bibfield{author}{\bibinfo{person}{Md~Hasan Anowar}, \bibinfo{person}{Abdullah Shamail}, \bibinfo{person}{Xiaoyu Wang}, \bibinfo{person}{Goce Trajcevski}, \bibinfo{person}{Sohail Murad}, \bibinfo{person}{Cynthia~J. Jameson}, {and} \bibinfo{person}{Ashfaq Khokhar}.} \bibinfo{year}{2024}\natexlab{}.
\newblock \showarticletitle{Compressing generalized trajectories of molecular motion for efficient detection of chemical interactions}.
\newblock \bibinfo{journal}{\emph{Information Systems}}  \bibinfo{volume}{125} (\bibinfo{year}{2024}), \bibinfo{pages}{102426}.
\newblock
\showISSN{0306-4379}
\urldef\tempurl%
\url{https://doi.org/10.1016/j.is.2024.102426}
\showDOI{\tempurl}


\bibitem[Blöbaum et~al\mbox{.}(2024)]%
        {dowhy_gcm}
\bibfield{author}{\bibinfo{person}{Patrick Blöbaum}, \bibinfo{person}{Peter Götz}, \bibinfo{person}{Kailash Budhathoki}, \bibinfo{person}{Atalanti~A. Mastakouri}, {and} \bibinfo{person}{Dominik Janzing}.} \bibinfo{year}{2024}\natexlab{}.
\newblock \bibinfo{title}{DoWhy-GCM: An extension of DoWhy for causal inference in graphical causal models}.
\newblock
\newblock
\showeprint[arxiv]{2206.06821}~[stat.ME]
\urldef\tempurl%
\url{https://arxiv.org/abs/2206.06821}
\showURL{%
\tempurl}


\bibitem[Catlett et~al\mbox{.}(2019)]%
        {Catlett_2019}
\bibfield{author}{\bibinfo{person}{Charlie Catlett}, \bibinfo{person}{Eugenio Cesario}, \bibinfo{person}{Domenico Talia}, {and} \bibinfo{person}{Andrea Vinci}.} \bibinfo{year}{2019}\natexlab{}.
\newblock \showarticletitle{Spatio-temporal crime predictions in smart cities: A data-driven approach and experiments}.
\newblock \bibinfo{journal}{\emph{Pervasive and Mobile Computing}}  \bibinfo{volume}{53} (\bibinfo{year}{2019}), \bibinfo{pages}{62--74}.
\newblock


\bibitem[Chen et~al\mbox{.}(2011)]%
        {Chen_2011}
\bibfield{author}{\bibinfo{person}{Qiuwen Chen}, \bibinfo{person}{Rui Han}, \bibinfo{person}{Fei Ye}, {and} \bibinfo{person}{Weifeng Li}.} \bibinfo{year}{2011}\natexlab{}.
\newblock \showarticletitle{Spatio-temporal ecological models}.
\newblock \bibinfo{journal}{\emph{Ecological Informatics}} \bibinfo{volume}{6}, \bibinfo{number}{1} (\bibinfo{year}{2011}), \bibinfo{pages}{37--43}.
\newblock


\bibitem[Chmiela et~al\mbox{.}(2017)]%
        {chmiela2017machine}
\bibfield{author}{\bibinfo{person}{Stefan Chmiela}, \bibinfo{person}{Alexandre Tkatchenko}, \bibinfo{person}{Huziel~E Sauceda}, \bibinfo{person}{Igor Poltavsky}, \bibinfo{person}{Kristof~T Sch{\"u}tt}, {and} \bibinfo{person}{Klaus-Robert M{\"u}ller}.} \bibinfo{year}{2017}\natexlab{}.
\newblock \showarticletitle{Machine learning of accurate energy-conserving molecular force fields}.
\newblock \bibinfo{journal}{\emph{Science advances}} \bibinfo{volume}{3}, \bibinfo{number}{5} (\bibinfo{year}{2017}), \bibinfo{pages}{e1603015}.
\newblock


\bibitem[Correa and Bareinboim(2020)]%
        {correa2020calculus}
\bibfield{author}{\bibinfo{person}{Juan Correa} {and} \bibinfo{person}{Elias Bareinboim}.} \bibinfo{year}{2020}\natexlab{}.
\newblock \showarticletitle{A Calculus for Stochastic Interventions:Causal Effect Identification and Surrogate Experiments}.
\newblock \bibinfo{journal}{\emph{Proceedings of the AAAI Conference on Artificial Intelligence}} \bibinfo{volume}{34}, \bibinfo{number}{06} (\bibinfo{date}{Apr.} \bibinfo{year}{2020}), \bibinfo{pages}{10093--10100}.
\newblock
\urldef\tempurl%
\url{https://doi.org/10.1609/aaai.v34i06.6567}
\showDOI{\tempurl}


\bibitem[Cover(1999)]%
        {cover1999elements}
\bibfield{author}{\bibinfo{person}{Thomas~M Cover}.} \bibinfo{year}{1999}\natexlab{}.
\newblock \bibinfo{booktitle}{\emph{Elements of information theory}}.
\newblock \bibinfo{publisher}{John Wiley \& Sons}, \bibinfo{address}{USA}.
\newblock


\bibitem[Dean and Kanazawa(1989)]%
        {dean1989model}
\bibfield{author}{\bibinfo{person}{Thomas Dean} {and} \bibinfo{person}{Keiji Kanazawa}.} \bibinfo{year}{1989}\natexlab{}.
\newblock \showarticletitle{A model for reasoning about persistence and causation}.
\newblock \bibinfo{journal}{\emph{Computational intelligence}} \bibinfo{volume}{5}, \bibinfo{number}{2} (\bibinfo{year}{1989}), \bibinfo{pages}{142--150}.
\newblock


\bibitem[Doersch(2016)]%
        {doersch2016tutorial}
\bibfield{author}{\bibinfo{person}{Carl Doersch}.} \bibinfo{year}{2016}\natexlab{}.
\newblock \showarticletitle{Tutorial on Variational Autoencoders}.
\newblock \bibinfo{journal}{\emph{ArXiv}}  \bibinfo{volume}{abs/1606.05908} (\bibinfo{year}{2016}), \bibinfo{pages}{23}.
\newblock
\urldef\tempurl%
\url{https://api.semanticscholar.org/CorpusID:10510670}
\showURL{%
\tempurl}


\bibitem[Erwig et~al\mbox{.}(1999)]%
        {Erwig_1999}
\bibfield{author}{\bibinfo{person}{Martin Erwig}, \bibinfo{person}{Ralf~Hartmut Gu{\"{}}~ting}, \bibinfo{person}{Markus Schneider}, {and} \bibinfo{person}{Michalis Vazirgiannis}.} \bibinfo{year}{1999}\natexlab{}.
\newblock \showarticletitle{Spatio-temporal data types: An approach to modeling and querying moving objects in databases}.
\newblock \bibinfo{journal}{\emph{GeoInformatica}} \bibinfo{volume}{3}, \bibinfo{number}{3} (\bibinfo{year}{1999}), \bibinfo{pages}{269--296}.
\newblock


\bibitem[Gilmer et~al\mbox{.}(2017)]%
        {gilmer2017neural}
\bibfield{author}{\bibinfo{person}{Justin Gilmer}, \bibinfo{person}{Samuel~S. Schoenholz}, \bibinfo{person}{Patrick~F. Riley}, \bibinfo{person}{Oriol Vinyals}, {and} \bibinfo{person}{George~E. Dahl}.} \bibinfo{year}{2017}\natexlab{}.
\newblock \showarticletitle{Neural message passing for Quantum chemistry}. In \bibinfo{booktitle}{\emph{Proceedings of the 34th International Conference on Machine Learning - Volume 70}} \emph{(\bibinfo{series}{ICML'17})}. \bibinfo{publisher}{JMLR.org}, \bibinfo{address}{Sydney, NSW, Australia}, \bibinfo{pages}{1263–1272}.
\newblock


\bibitem[Hofstetter et~al\mbox{.}(2022)]%
        {Hofstetter_2022}
\bibfield{author}{\bibinfo{person}{Albert Hofstetter}, \bibinfo{person}{Lennard B{\"o}selt}, {and} \bibinfo{person}{Sereina Riniker}.} \bibinfo{year}{2022}\natexlab{}.
\newblock \showarticletitle{Graph-convolutional neural networks for (QM) ML/MM molecular dynamics simulations}.
\newblock \bibinfo{journal}{\emph{Physical Chemistry Chemical Physics}} \bibinfo{volume}{24}, \bibinfo{number}{37} (\bibinfo{year}{2022}), \bibinfo{pages}{22497--22512}.
\newblock


\bibitem[Ikram et~al\mbox{.}(2022)]%
        {ikram2022root}
\bibfield{author}{\bibinfo{person}{Azam Ikram}, \bibinfo{person}{Sarthak Chakraborty}, \bibinfo{person}{Subrata Mitra}, \bibinfo{person}{Shiv Saini}, \bibinfo{person}{Saurabh Bagchi}, {and} \bibinfo{person}{Murat Kocaoglu}.} \bibinfo{year}{2022}\natexlab{}.
\newblock \showarticletitle{Root cause analysis of failures in microservices through causal discovery}.
\newblock \bibinfo{journal}{\emph{Advances in Neural Information Processing Systems}}  \bibinfo{volume}{35} (\bibinfo{year}{2022}), \bibinfo{pages}{31158--31170}.
\newblock


\bibitem[Jain et~al\mbox{.}(2016)]%
        {Jain_2016}
\bibfield{author}{\bibinfo{person}{Ashesh Jain}, \bibinfo{person}{Amir~R. Zamir}, \bibinfo{person}{Silvio Savarese}, {and} \bibinfo{person}{Ashutosh Saxena}.} \bibinfo{year}{2016}\natexlab{}.
\newblock \showarticletitle{{ Structural-RNN: Deep Learning on Spatio-Temporal Graphs }}. In \bibinfo{booktitle}{\emph{2016 IEEE Conference on Computer Vision and Pattern Recognition (CVPR)}}. \bibinfo{publisher}{IEEE Computer Society}, \bibinfo{address}{Los Alamitos, CA, USA}, \bibinfo{pages}{5308--5317}.
\newblock
\showISSN{1063-6919}
\urldef\tempurl%
\url{https://doi.ieeecomputersociety.org/10.1109/CVPR.2016.573}
\showURL{%
\tempurl}


\bibitem[Janzing et~al\mbox{.}(2019)]%
        {janzing2019causal}
\bibfield{author}{\bibinfo{person}{Dominik Janzing}, \bibinfo{person}{Kailash Budhathoki}, \bibinfo{person}{Lenon Minorics}, {and} \bibinfo{person}{Patrick Blöbaum}.} \bibinfo{year}{2019}\natexlab{}.
\newblock \bibinfo{title}{Causal structure based root cause analysis of outliers}.
\newblock
\newblock
\showeprint[arxiv]{1912.02724}~[stat.ML]


\bibitem[Jin et~al\mbox{.}(2024)]%
        {Jin_2023}
\bibfield{author}{\bibinfo{person}{Guangyin Jin}, \bibinfo{person}{Yuxuan Liang}, \bibinfo{person}{Yuchen Fang}, \bibinfo{person}{Zezhi Shao}, \bibinfo{person}{Jincai Huang}, \bibinfo{person}{Junbo Zhang}, {and} \bibinfo{person}{Yu Zheng}.} \bibinfo{year}{2024}\natexlab{}.
\newblock \showarticletitle{{ Spatio-Temporal Graph Neural Networks for Predictive Learning in Urban Computing: A Survey }}.
\newblock \bibinfo{journal}{\emph{IEEE Transactions on Knowledge \& Data Engineering}} \bibinfo{volume}{36}, \bibinfo{number}{10} (\bibinfo{date}{Oct.} \bibinfo{year}{2024}), \bibinfo{pages}{5388--5408}.
\newblock
\showISSN{1558-2191}
\urldef\tempurl%
\url{https://doi.org/10.1109/TKDE.2023.3333824}
\showDOI{\tempurl}


\bibitem[Kingma(2014)]%
        {kingma2014adam}
\bibfield{author}{\bibinfo{person}{Diederik~P Kingma}.} \bibinfo{year}{2014}\natexlab{}.
\newblock \showarticletitle{Adam: A method for stochastic optimization}.
\newblock \bibinfo{journal}{\emph{arXiv preprint arXiv:1412.6980}} (\bibinfo{year}{2014}).
\newblock


\bibitem[Kingma and Welling(2022)]%
        {kingma2013auto}
\bibfield{author}{\bibinfo{person}{Diederik~P Kingma} {and} \bibinfo{person}{Max Welling}.} \bibinfo{year}{2022}\natexlab{}.
\newblock \bibinfo{title}{Auto-Encoding Variational Bayes}.
\newblock
\newblock
\showeprint[arxiv]{1312.6114}~[stat.ML]
\urldef\tempurl%
\url{https://arxiv.org/abs/1312.6114}
\showURL{%
\tempurl}


\bibitem[Kipf et~al\mbox{.}(2018)]%
        {kipf2018neural}
\bibfield{author}{\bibinfo{person}{Thomas Kipf}, \bibinfo{person}{Ethan Fetaya}, \bibinfo{person}{Kuan-Chieh Wang}, \bibinfo{person}{Max Welling}, {and} \bibinfo{person}{Richard Zemel}.} \bibinfo{year}{2018}\natexlab{}.
\newblock \showarticletitle{Neural Relational Inference for Interacting Systems}. In \bibinfo{booktitle}{\emph{Proceedings of the 35th International Conference on Machine Learning}} \emph{(\bibinfo{series}{Proceedings of Machine Learning Research}, Vol.~\bibinfo{volume}{80})}, \bibfield{editor}{\bibinfo{person}{Jennifer Dy} {and} \bibinfo{person}{Andreas Krause}} (Eds.). \bibinfo{publisher}{PMLR}, \bibinfo{address}{Stockholm, Sweden}, \bibinfo{pages}{2688--2697}.
\newblock
\urldef\tempurl%
\url{https://proceedings.mlr.press/v80/kipf18a.html}
\showURL{%
\tempurl}


\bibitem[Li et~al\mbox{.}(2017)]%
        {yujia2016gated}
\bibfield{author}{\bibinfo{person}{Yujia Li}, \bibinfo{person}{Daniel Tarlow}, \bibinfo{person}{Marc Brockschmidt}, {and} \bibinfo{person}{Richard Zemel}.} \bibinfo{year}{2017}\natexlab{}.
\newblock \bibinfo{title}{Gated Graph Sequence Neural Networks}.
\newblock
\newblock
\showeprint[arxiv]{1511.05493}~[cs.LG]
\urldef\tempurl%
\url{https://arxiv.org/abs/1511.05493}
\showURL{%
\tempurl}


\bibitem[Li et~al\mbox{.}(2022)]%
        {Li_2022}
\bibfield{author}{\bibinfo{person}{Zijie Li}, \bibinfo{person}{Kazem Meidani}, \bibinfo{person}{Prakarsh Yadav}, {and} \bibinfo{person}{Amir Barati~Farimani}.} \bibinfo{year}{2022}\natexlab{}.
\newblock \showarticletitle{Graph neural networks accelerated molecular dynamics}.
\newblock \bibinfo{journal}{\emph{The Journal of Chemical Physics}} \bibinfo{volume}{156}, \bibinfo{number}{14} (\bibinfo{date}{April} \bibinfo{year}{2022}), \bibinfo{pages}{144103--144103}.
\newblock
\showISSN{1089-7690}
\urldef\tempurl%
\url{https://doi.org/10.1063/5.0083060}
\showDOI{\tempurl}


\bibitem[Liu et~al\mbox{.}(2022)]%
        {liu2022structural}
\bibfield{author}{\bibinfo{person}{Qi Liu}, \bibinfo{person}{Yuanqi Du}, \bibinfo{person}{Fan Feng}, \bibinfo{person}{Qiwei Ye}, {and} \bibinfo{person}{Jie Fu}.} \bibinfo{year}{2022}\natexlab{}.
\newblock \showarticletitle{Structural Causal Model for Molecular Dynamics Simulation}. In \bibinfo{booktitle}{\emph{NeurIPS 2022 AI for Science: Progress and Promises}}. \bibinfo{publisher}{NeurIPS}, \bibinfo{address}{New Orleans, USA}.
\newblock


\bibitem[L{\'o}pez-Qu{\i}lez and Munoz(2009)]%
        {López‐Quílez_2009}
\bibfield{author}{\bibinfo{person}{Antonio L{\'o}pez-Qu{\i}lez} {and} \bibinfo{person}{Facundo Munoz}.} \bibinfo{year}{2009}\natexlab{}.
\newblock \showarticletitle{Review of spatio-temporal models for disease mapping}.
\newblock \bibinfo{journal}{\emph{Final Report for the EUROHEIS}}  \bibinfo{volume}{2} (\bibinfo{year}{2009}).
\newblock


\bibitem[L{\"o}we et~al\mbox{.}(2022)]%
        {lowe2022amortized}
\bibfield{author}{\bibinfo{person}{Sindy L{\"o}we}, \bibinfo{person}{David Madras}, \bibinfo{person}{Richard Zemel}, {and} \bibinfo{person}{Max Welling}.} \bibinfo{year}{2022}\natexlab{}.
\newblock \showarticletitle{Amortized causal discovery: Learning to infer causal graphs from time-series data}. In \bibinfo{booktitle}{\emph{Conference on Causal Learning and Reasoning}}. PMLR, \bibinfo{publisher}{PMLR}, \bibinfo{address}{CA, USA}, \bibinfo{pages}{509--525}.
\newblock


\bibitem[Maddison et~al\mbox{.}(2016)]%
        {maddison2016concrete}
\bibfield{author}{\bibinfo{person}{Chris~J Maddison}, \bibinfo{person}{Andriy Mnih}, {and} \bibinfo{person}{Yee~Whye Teh}.} \bibinfo{year}{2016}\natexlab{}.
\newblock \showarticletitle{The concrete distribution: A continuous relaxation of discrete random variables}.
\newblock \bibinfo{journal}{\emph{arXiv preprint arXiv:1611.00712}} (\bibinfo{year}{2016}).
\newblock


\bibitem[Mao et~al\mbox{.}(2019)]%
        {Mao_2019}
\bibfield{author}{\bibinfo{person}{Zhenyu Mao}, \bibinfo{person}{Yi Su}, \bibinfo{person}{Guangquan Xu}, \bibinfo{person}{Xueping Wang}, \bibinfo{person}{Yu Huang}, \bibinfo{person}{Weihua Yue}, \bibinfo{person}{Li Sun}, {and} \bibinfo{person}{Naixue Xiong}.} \bibinfo{year}{2019}\natexlab{}.
\newblock \showarticletitle{Spatio-temporal deep learning method for adhd fmri classification}.
\newblock \bibinfo{journal}{\emph{Information Sciences}}  \bibinfo{volume}{499} (\bibinfo{year}{2019}), \bibinfo{pages}{1--11}.
\newblock


\bibitem[Murtagh(1991)]%
        {murtagh1991multilayer}
\bibfield{author}{\bibinfo{person}{Fionn Murtagh}.} \bibinfo{year}{1991}\natexlab{}.
\newblock \showarticletitle{Multilayer perceptrons for classification and regression}.
\newblock \bibinfo{journal}{\emph{Neurocomputing}} \bibinfo{volume}{2}, \bibinfo{number}{5-6} (\bibinfo{year}{1991}), \bibinfo{pages}{183--197}.
\newblock


\bibitem[Pearl(2009)]%
        {pearl2009causality}
\bibfield{author}{\bibinfo{person}{J. Pearl}.} \bibinfo{year}{2009}\natexlab{}.
\newblock \bibinfo{booktitle}{\emph{Causality}}.
\newblock \bibinfo{publisher}{Cambridge University Press}, \bibinfo{address}{NY,USA}.
\newblock
\showISBNx{9780521895606}
\showLCCN{99042108}
\urldef\tempurl%
\url{https://books.google.com/books?id=f4nuexsNVZIC}
\showURL{%
\tempurl}


\bibitem[Peerally et~al\mbox{.}(2017)]%
        {peerally2017problem}
\bibfield{author}{\bibinfo{person}{Mohammad~Farhad Peerally}, \bibinfo{person}{Susan Carr}, \bibinfo{person}{Justin Waring}, {and} \bibinfo{person}{Mary Dixon-Woods}.} \bibinfo{year}{2017}\natexlab{}.
\newblock \showarticletitle{The problem with root cause analysis}.
\newblock \bibinfo{journal}{\emph{BMJ quality \& safety}} \bibinfo{volume}{26}, \bibinfo{number}{5} (\bibinfo{year}{2017}), \bibinfo{pages}{417--422}.
\newblock


\bibitem[Peters et~al\mbox{.}(2017)]%
        {peters2017elements}
\bibfield{author}{\bibinfo{person}{Jonas Peters}, \bibinfo{person}{Dominik Janzing}, {and} \bibinfo{person}{Bernhard Sch{\"o}lkopf}.} \bibinfo{year}{2017}\natexlab{}.
\newblock \bibinfo{booktitle}{\emph{Elements of causal inference: foundations and learning algorithms}}.
\newblock \bibinfo{publisher}{The MIT Press}, \bibinfo{address}{USA}.
\newblock


\bibitem[Qiu et~al\mbox{.}(2020)]%
        {qiu2020causality}
\bibfield{author}{\bibinfo{person}{Juan Qiu}, \bibinfo{person}{Qingfeng Du}, \bibinfo{person}{Kanglin Yin}, \bibinfo{person}{Shuang-Li Zhang}, {and} \bibinfo{person}{Chongshu Qian}.} \bibinfo{year}{2020}\natexlab{}.
\newblock \showarticletitle{A causality mining and knowledge graph based method of root cause diagnosis for performance anomaly in cloud applications}.
\newblock \bibinfo{journal}{\emph{Applied Sciences}} \bibinfo{volume}{10}, \bibinfo{number}{6} (\bibinfo{year}{2020}), \bibinfo{pages}{2166}.
\newblock


\bibitem[Rezende et~al\mbox{.}(2014)]%
        {rezende2014stochastic}
\bibfield{author}{\bibinfo{person}{Danilo~Jimenez Rezende}, \bibinfo{person}{Shakir Mohamed}, {and} \bibinfo{person}{Daan Wierstra}.} \bibinfo{year}{2014}\natexlab{}.
\newblock \showarticletitle{Stochastic backpropagation and approximate inference in deep generative models}. In \bibinfo{booktitle}{\emph{International conference on machine learning}}. \bibinfo{publisher}{PMLR}, \bibinfo{address}{China}, \bibinfo{pages}{1278--1286}.
\newblock


\bibitem[Rubin(2005)]%
        {rubin2005causal}
\bibfield{author}{\bibinfo{person}{Donald~B Rubin}.} \bibinfo{year}{2005}\natexlab{}.
\newblock \showarticletitle{Causal inference using potential outcomes: Design, modeling, decisions}.
\newblock \bibinfo{journal}{\emph{J. Amer. Statist. Assoc.}} \bibinfo{volume}{100}, \bibinfo{number}{469} (\bibinfo{year}{2005}), \bibinfo{pages}{322--331}.
\newblock


\bibitem[Sahili and Awad(2023)]%
        {Sahili_2023}
\bibfield{author}{\bibinfo{person}{Zahraa~Al Sahili} {and} \bibinfo{person}{Mariette Awad}.} \bibinfo{year}{2023}\natexlab{}.
\newblock \showarticletitle{Spatio-temporal graph neural networks: A survey}.
\newblock \bibinfo{journal}{\emph{arXiv preprint arXiv:2301.10569}}  \bibinfo{volume}{abs/2301.10569} (\bibinfo{year}{2023}), \bibinfo{pages}{arXiv--2301}.
\newblock


\bibitem[Scarselli et~al\mbox{.}(2008)]%
        {scarselli2008graph}
\bibfield{author}{\bibinfo{person}{Franco Scarselli}, \bibinfo{person}{Marco Gori}, \bibinfo{person}{Ah~Chung Tsoi}, \bibinfo{person}{Markus Hagenbuchner}, {and} \bibinfo{person}{Gabriele Monfardini}.} \bibinfo{year}{2008}\natexlab{}.
\newblock \showarticletitle{The graph neural network model}.
\newblock \bibinfo{journal}{\emph{IEEE transactions on neural networks}} \bibinfo{volume}{20}, \bibinfo{number}{1} (\bibinfo{year}{2008}), \bibinfo{pages}{61--80}.
\newblock


\bibitem[Sch{\"o}lkopf et~al\mbox{.}(2021)]%
        {scholkopf2021toward}
\bibfield{author}{\bibinfo{person}{Bernhard Sch{\"o}lkopf}, \bibinfo{person}{Francesco Locatello}, \bibinfo{person}{Stefan Bauer}, \bibinfo{person}{Nan~Rosemary Ke}, \bibinfo{person}{Nal Kalchbrenner}, \bibinfo{person}{Anirudh Goyal}, {and} \bibinfo{person}{Yoshua Bengio}.} \bibinfo{year}{2021}\natexlab{}.
\newblock \showarticletitle{Toward causal representation learning}.
\newblock \bibinfo{journal}{\emph{Proc. IEEE}} \bibinfo{volume}{109}, \bibinfo{number}{5} (\bibinfo{year}{2021}), \bibinfo{pages}{612--634}.
\newblock


\bibitem[Sharma and Kiciman(2020)]%
        {dowhy}
\bibfield{author}{\bibinfo{person}{Amit Sharma} {and} \bibinfo{person}{Emre Kiciman}.} \bibinfo{year}{2020}\natexlab{}.
\newblock \bibinfo{title}{DoWhy: An End-to-End Library for Causal Inference}.
\newblock
\newblock
\showeprint[arxiv]{2011.04216}~[stat.ME]
\urldef\tempurl%
\url{https://arxiv.org/abs/2011.04216}
\showURL{%
\tempurl}


\bibitem[Shi and Yeung(2018)]%
        {Shi_2018}
\bibfield{author}{\bibinfo{person}{Xingjian Shi} {and} \bibinfo{person}{Dit-Yan Yeung}.} \bibinfo{year}{2018}\natexlab{}.
\newblock \showarticletitle{Machine learning for spatiotemporal sequence forecasting: A survey}.
\newblock \bibinfo{journal}{\emph{arXiv preprint arXiv:1808.06865}}  \bibinfo{volume}{abs/1808.06865} (\bibinfo{year}{2018}), \bibinfo{pages}{arXiv--1808}.
\newblock


\bibitem[Spirtes et~al\mbox{.}(2001)]%
        {spirtes2001causation}
\bibfield{author}{\bibinfo{person}{Peter Spirtes}, \bibinfo{person}{Clark Glymour}, {and} \bibinfo{person}{Richard Scheines}.} \bibinfo{year}{2001}\natexlab{}.
\newblock \bibinfo{booktitle}{\emph{Causation, prediction, and search}}.
\newblock \bibinfo{publisher}{MIT press}, \bibinfo{address}{USA}.
\newblock


\bibitem[Stocker et~al\mbox{.}(2022)]%
        {Stocker_2022}
\bibfield{author}{\bibinfo{person}{Sina Stocker}, \bibinfo{person}{Johannes Gasteiger}, \bibinfo{person}{Florian Becker}, \bibinfo{person}{Stephan G{\"u}nnemann}, {and} \bibinfo{person}{Johannes~T Margraf}.} \bibinfo{year}{2022}\natexlab{}.
\newblock \showarticletitle{How robust are modern graph neural network potentials in long and hot molecular dynamics simulations?}
\newblock \bibinfo{journal}{\emph{Machine Learning: Science and Technology}} \bibinfo{volume}{3}, \bibinfo{number}{4} (\bibinfo{year}{2022}), \bibinfo{pages}{045010}.
\newblock


\bibitem[Theodoridis et~al\mbox{.}(1996)]%
        {Theoderidis_1996}
\bibfield{author}{\bibinfo{person}{Yannis Theodoridis}, \bibinfo{person}{Michael Vazirgiannis}, {and} \bibinfo{person}{Timos Sellis}.} \bibinfo{year}{1996}\natexlab{}.
\newblock \showarticletitle{Spatio-temporal indexing for large multimedia applications}. In \bibinfo{booktitle}{\emph{Proceedings of the Third IEEE International Conference on Multimedia Computing and Systems}}. IEEE, \bibinfo{publisher}{IEEE}, \bibinfo{address}{Hiroshima, Japan}, \bibinfo{pages}{441--448}.
\newblock


\bibitem[Unke et~al\mbox{.}(2021)]%
        {unke2021machine}
\bibfield{author}{\bibinfo{person}{Oliver~T Unke}, \bibinfo{person}{Stefan Chmiela}, \bibinfo{person}{Huziel~E Sauceda}, \bibinfo{person}{Michael Gastegger}, \bibinfo{person}{Igor Poltavsky}, \bibinfo{person}{Kristof~T Sch{\"u}tt}, \bibinfo{person}{Alexandre Tkatchenko}, {and} \bibinfo{person}{Klaus-Robert M{\"u}ller}.} \bibinfo{year}{2021}\natexlab{}.
\newblock \showarticletitle{Machine learning force fields}.
\newblock \bibinfo{journal}{\emph{Chemical Reviews}} \bibinfo{volume}{121}, \bibinfo{number}{16} (\bibinfo{year}{2021}), \bibinfo{pages}{10142--10186}.
\newblock


\bibitem[Unke and Meuwly(2019)]%
        {unke2019physnet}
\bibfield{author}{\bibinfo{person}{Oliver~T Unke} {and} \bibinfo{person}{Markus Meuwly}.} \bibinfo{year}{2019}\natexlab{}.
\newblock \showarticletitle{PhysNet: A neural network for predicting energies, forces, dipole moments, and partial charges}.
\newblock \bibinfo{journal}{\emph{Journal of chemical theory and computation}} \bibinfo{volume}{15}, \bibinfo{number}{6} (\bibinfo{year}{2019}), \bibinfo{pages}{3678--3693}.
\newblock


\bibitem[Urbano et~al\mbox{.}(2010)]%
        {Urbano_2010}
\bibfield{author}{\bibinfo{person}{Ferdinando Urbano}, \bibinfo{person}{Francesca Cagnacci}, \bibinfo{person}{Cl{\'e}ment Calenge}, \bibinfo{person}{Holger Dettki}, \bibinfo{person}{Alison Cameron}, {and} \bibinfo{person}{Markus Neteler}.} \bibinfo{year}{2010}\natexlab{}.
\newblock \showarticletitle{Wildlife tracking data management: a new vision}.
\newblock \bibinfo{journal}{\emph{Philosophical Transactions of the Royal Society B: Biological Sciences}} \bibinfo{volume}{365}, \bibinfo{number}{1550} (\bibinfo{year}{2010}), \bibinfo{pages}{2177--2185}.
\newblock


\bibitem[Wang et~al\mbox{.}(2023)]%
        {wang2023incremental}
\bibfield{author}{\bibinfo{person}{Dongjie Wang}, \bibinfo{person}{Zhengzhang Chen}, \bibinfo{person}{Yanjie Fu}, \bibinfo{person}{Yanchi Liu}, {and} \bibinfo{person}{Haifeng Chen}.} \bibinfo{year}{2023}\natexlab{}.
\newblock \showarticletitle{Incremental Causal Graph Learning for Online Root Cause Analysis}. In \bibinfo{booktitle}{\emph{Proceedings of the 29th ACM SIGKDD Conference on Knowledge Discovery and Data Mining}} (Long Beach, CA, USA) \emph{(\bibinfo{series}{KDD '23})}. \bibinfo{publisher}{Association for Computing Machinery}, \bibinfo{address}{New York, NY, USA}, \bibinfo{pages}{2269–2278}.
\newblock
\showISBNx{9798400701030}
\urldef\tempurl%
\url{https://doi.org/10.1145/3580305.3599392}
\showDOI{\tempurl}


\bibitem[Wang et~al\mbox{.}(2018)]%
        {wang2018cloudranger}
\bibfield{author}{\bibinfo{person}{Ping Wang}, \bibinfo{person}{Jingmin Xu}, \bibinfo{person}{Meng Ma}, \bibinfo{person}{Weilan Lin}, \bibinfo{person}{Disheng Pan}, \bibinfo{person}{Yuan Wang}, {and} \bibinfo{person}{Pengfei Chen}.} \bibinfo{year}{2018}\natexlab{}.
\newblock \showarticletitle{Cloudranger: Root cause identification for cloud native systems}. In \bibinfo{booktitle}{\emph{2018 18th IEEE/ACM International Symposium on Cluster, Cloud and Grid Computing (CCGRID)}}. \bibinfo{publisher}{IEEE}, \bibinfo{address}{Washington, DC, USA}, \bibinfo{pages}{492--502}.
\newblock


\bibitem[Wang et~al\mbox{.}(2020a)]%
        {Wang_2020}
\bibfield{author}{\bibinfo{person}{Senzhang Wang}, \bibinfo{person}{Jiannong Cao}, {and} \bibinfo{person}{S~Yu Philip}.} \bibinfo{year}{2020}\natexlab{a}.
\newblock \showarticletitle{Deep learning for spatio-temporal data mining: A survey}.
\newblock \bibinfo{journal}{\emph{IEEE transactions on knowledge and data engineering}} \bibinfo{volume}{34}, \bibinfo{number}{8} (\bibinfo{year}{2020}), \bibinfo{pages}{3681--3700}.
\newblock


\bibitem[Wang et~al\mbox{.}(2019)]%
        {Wang2019}
\bibfield{author}{\bibinfo{person}{Xiaoyu Wang}, \bibinfo{person}{David~W. House}, \bibinfo{person}{Priyanka~A. Oroskar}, \bibinfo{person}{Anil Oroskar}, \bibinfo{person}{Asha Oroskar}, \bibinfo{person}{Cynthia~J. Jameson}, {and} \bibinfo{person}{Sohail Murad}.} \bibinfo{year}{2019}\natexlab{}.
\newblock \showarticletitle{MD simulations of the chiral recognition mechanism for a polysaccharide chiral stationary phase in enantiomeric chromatographic separations}.
\newblock \bibinfo{journal}{\emph{An International Journal at the Interface Between Chemistry and Physics}}  \bibinfo{volume}{117} (\bibinfo{date}{12} \bibinfo{year}{2019}).
\newblock
Issue 23-24.
\showISSN{13623028}


\bibitem[Wang et~al\mbox{.}(2020b)]%
        {Wang2020}
\bibfield{author}{\bibinfo{person}{Xiaoyu Wang}, \bibinfo{person}{Cynthia~J. Jameson}, {and} \bibinfo{person}{Sohail Murad}.} \bibinfo{year}{2020}\natexlab{b}.
\newblock \showarticletitle{Modeling Enantiomeric Separations as an Interfacial Process Using Amylose Tris(3,5-dimethylphenyl carbamate) (ADMPC) Polymers Coated on Amorphous Silica}.
\newblock \bibinfo{journal}{\emph{Langmuir}}  \bibinfo{volume}{36} (\bibinfo{date}{2} \bibinfo{year}{2020}), \bibinfo{pages}{1113--1124}.
\newblock
Issue 5.
\showISSN{15205827}


\bibitem[Zhang et~al\mbox{.}(2021)]%
        {zhang2020influence}
\bibfield{author}{\bibinfo{person}{Keli Zhang}, \bibinfo{person}{Marcus Kalander}, \bibinfo{person}{Min Zhou}, \bibinfo{person}{Xi Zhang}, {and} \bibinfo{person}{Junjian Ye}.} \bibinfo{year}{2021}\natexlab{}.
\newblock \showarticletitle{An Influence-Based Approach for Root Cause Alarm Discovery in Telecom Networks}. In \bibinfo{booktitle}{\emph{Service-Oriented Computing -- ICSOC 2020 Workshops}}, \bibfield{editor}{\bibinfo{person}{Hakim Hacid}, \bibinfo{person}{Fatma Outay}, \bibinfo{person}{Hye-young Paik}, \bibinfo{person}{Amira Alloum}, \bibinfo{person}{Marinella Petrocchi}, \bibinfo{person}{Mohamed~Reda Bouadjenek}, \bibinfo{person}{Amin Beheshti}, \bibinfo{person}{Xumin Liu}, {and} \bibinfo{person}{Abderrahmane Maaradji}} (Eds.). \bibinfo{publisher}{Springer International Publishing}, \bibinfo{address}{Cham}, \bibinfo{pages}{124--136}.
\newblock
\showISBNx{978-3-030-76352-7}


\bibitem[Zhang et~al\mbox{.}(2023)]%
        {Zhang_2023}
\bibfield{author}{\bibinfo{person}{Zijian Zhang}, \bibinfo{person}{Ze Huang}, \bibinfo{person}{Zhiwei Hu}, \bibinfo{person}{Xiangyu Zhao}, \bibinfo{person}{Wanyu Wang}, \bibinfo{person}{Zitao Liu}, \bibinfo{person}{Junbo Zhang}, \bibinfo{person}{S.~Joe Qin}, {and} \bibinfo{person}{Hongwei Zhao}.} \bibinfo{year}{2023}\natexlab{}.
\newblock \showarticletitle{MLPST: MLP is All You Need for Spatio-Temporal Prediction}. In \bibinfo{booktitle}{\emph{Proceedings of the 32nd ACM International Conference on Information and Knowledge Management}} (Birmingham, United Kingdom) \emph{(\bibinfo{series}{CIKM '23})}. \bibinfo{publisher}{Association for Computing Machinery}, \bibinfo{address}{New York, NY, USA}, \bibinfo{pages}{3381–3390}.
\newblock
\showISBNx{9798400701245}
\urldef\tempurl%
\url{https://doi.org/10.1145/3583780.3614969}
\showDOI{\tempurl}


\bibitem[Zhu et~al\mbox{.}(2022)]%
        {zhu2022neural}
\bibfield{author}{\bibinfo{person}{Jingxuan Zhu}, \bibinfo{person}{Juexin Wang}, \bibinfo{person}{Weiwei Han}, {and} \bibinfo{person}{Dong Xu}.} \bibinfo{year}{2022}\natexlab{}.
\newblock \showarticletitle{Neural relational inference to learn long-range allosteric interactions in proteins from molecular dynamics simulations}.
\newblock \bibinfo{journal}{\emph{Nature communications}} \bibinfo{volume}{13}, \bibinfo{number}{1} (\bibinfo{year}{2022}), \bibinfo{pages}{1661}.
\newblock


\end{thebibliography}

\end{document}